\documentclass{article}

\usepackage{natbib}
\usepackage{fullpage}

\usepackage[utf8]{inputenc} 
\usepackage[T1]{fontenc}    
\usepackage{xcolor}
\usepackage{hyperref}       
\definecolor{myblue}{rgb}{0,0.2,0.8}
\hypersetup{ %
pdftitle={},
pdfauthor={},
pdfsubject={},
pdfkeywords={},
pdfborder=0 0 0,
pdfpagemode=UseNone,
colorlinks=true,
linkcolor=myblue,
citecolor=myblue,
filecolor=myblue,
urlcolor=myblue,
pdfview=FitH}
\usepackage{authblk}
\usepackage{url}            
\usepackage{booktabs}       
\usepackage{amsfonts}       
\usepackage{nicefrac}       
\usepackage{microtype}      
\RequirePackage{snapshot}
\usepackage{graphicx,color}
\usepackage{xspace}
\usepackage[varqu,varl,var0,scaled=0.97]{inconsolata}
\usepackage{caption}
\usepackage{overpic}
\usepackage{wrapfig}
\usepackage{tcolorbox}
\usepackage{enumitem}
\usepackage{threeparttable}

\usepackage{algorithm}
\usepackage{algpseudocode}
\usepackage{listings}
\usepackage{color}
\usepackage{lipsum}
\usepackage{adjustbox}

\usepackage{subfigure}

\definecolor{dkgreen}{rgb}{0,0.6,0}
\definecolor{gray}{rgb}{0.5,0.5,0.5}
\definecolor{mauve}{rgb}{0.58,0,0.82}

\newcommand{\tabref}[1]{Table~\ref{#1}}

\newcommand{\eg}{\emph{e.g.},\xspace}
\newcommand{\ie}{\emph{i.e.},\xspace}

\usepackage{amsmath,amsfonts,bm}

\def\Figref#1{Figure~\ref{#1}}

\def\secref#1{section~\ref{#1}}
\def\Secref#1{Section~\ref{#1}}

\def\eqref#1{equation~\ref{#1}}

\def\1{\bm{1}}

\DeclareMathAlphabet{\mathsfit}{\encodingdefault}{\sfdefault}{m}{sl}
\SetMathAlphabet{\mathsfit}{bold}{\encodingdefault}{\sfdefault}{bx}{n}

\newcommand\fakeparagraph[1]{\par\noindent\textbf{{#1}}.\xspace}

\title{The Role of Pre-training Data in Transfer Learning}

\author[1]{Rahim Entezari\footnote{Work done while interning at the University of Washington.}}
\author[2]{Mitchell Wortsman}
\author[1]{Olga Saukh}
\author[3]{\\M.Moein Shariatnia}
\author[4]{Hanie Sedghi}
\author[2]{Ludwig Schmidt}

\affil[1]{TU Graz / CSH Vienna}
\affil[2]{University of Washington}
\affil[3]{Tehran University of Medical Sciences}
\affil[4]{Google Research, Brain Team}

\begin{document}
\date{}
\maketitle
\begin{abstract}
The transfer learning paradigm of model pre-training and subsequent fine-tuning produces high-accuracy models. While most studies recommend scaling the pre-training size to benefit most from transfer learning, a question remains: what data and method should be used for pre-training? 
We investigate the impact of pre-training data distribution on the few-shot and full fine-tuning performance using 3 pre-training methods (supervised, contrastive language-image and image-image), 7 pre-training datasets, and 9 downstream datasets. Through extensive controlled experiments, we find that the choice of the pre-training data source is essential for the few-shot transfer, but its role decreases as more data is made available for fine-tuning. Additionally, we explore the role of data curation and examine the trade-offs between label noise and the size of the pre-training dataset. We find that using 2000$\times$ more pre-training data from LAION can match the performance of supervised ImageNet pre-training. Furthermore, we investigate the effect of pre-training methods, comparing language-image contrastive vs. image-image contrastive, and find that the latter leads to better downstream accuracy\footnote{ Our comparisons include a smaller scale pre-training compared to SOTA ImageNet results using CLIP pre-training. SOTA CLIP pre-training lacks a comprehensive comparison to image-only pre-training on few-shot and full-finetune.}\footnote{Code is available at~\url{https://github.com/rahimentezari/DataDistributionTransferLearning}}.

\end{abstract}

\section{Introduction}
\label{sec:intro}
The best-performing computer vision models are produced by the transfer learning paradigm. 
While transfer learning is not new, the substantial improvement in the quality of the pre-trained models in recent years has brought transfer learning to the spotlight~(e.g., CLIP~\citep{radford2021learning}, BASIC~\citep{pham2021combined}, and Flamingo~\citep{alayrac2022flamingo}). These improvements are driven by new datasets for pre-training as well as better pre-training algorithms. This naturally leads to a question:

\begin{center}
\emph{How do the dataset and the algorithm used for pre-training affect downstream performance?}
\end{center}

\begin{figure}[t]
    \centering
    \subfigure
    {\includegraphics[width=0.69\textwidth]
    {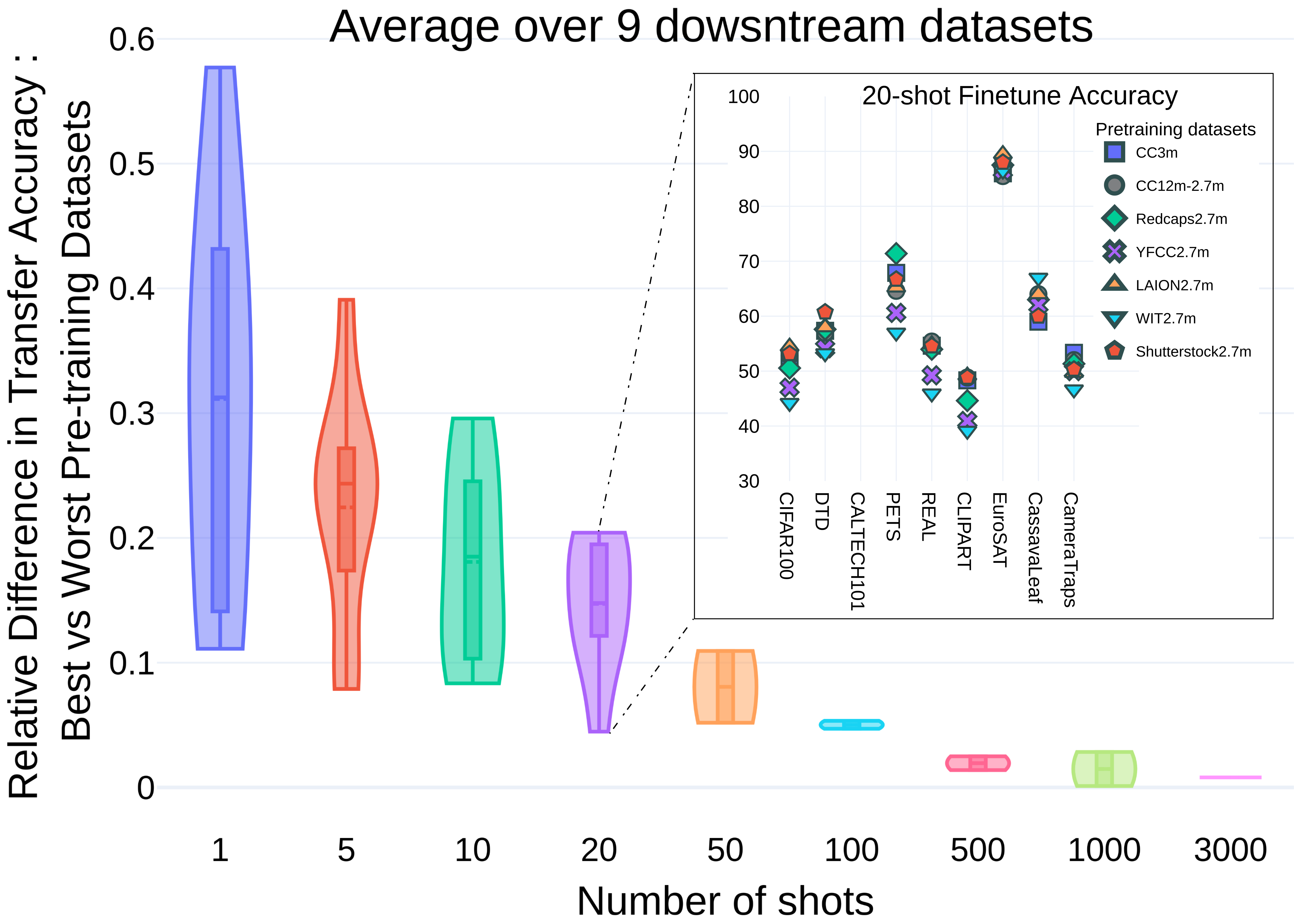}}

    \caption{\textbf{Differences between various pre-training sources diminish as more data is available for the downstream tasks.}
    In the few-shot setting, different pre-training datasets lead to noticeable differences in downstream performance. However, if many samples are available for fine-tuning, the difference in absolute accuracy between models pre-trained on different sources largely evaporates. 
    (see~\Figref{fig:shots_vs_acc_datadist_extend} for a detailed view).}
    \label{fig:shots_vs_acc_violon}
\end{figure}

In contrast to prior works~\citep{kornblith2019better, abnar2021exploring, you2021logme, nguyen2020leep, deshpande2021linearized, bolya2021scalable}, our main focus is on the role of the pre-training data distribution in downstream performance. We set up systematic experiments to explore our research questions and contributions as follow:

\paragraph{Do different pre-training distributions lead to different transfer learning performances?} 
In practice, one has many options to download pre-training checkpoints and fine-tune the model on the target dataset. 
Should we expect different pre-training datasets to perform differently in the transfer setting?
When controlling for the size of the pre-train model and the downstream dataset, but changing the pre-train dataset, we observe noticeable differences in downstream  accuracy in the few-shot setting (only a few examples per class are available for fine-tuning). However as more samples are available for fine-tuning, the difference in absolute accuracy when varying the pre-training dataset largely evaporates. 
In the few-shot regime, we observe that certain pre-training datasets (\eg Shutterstock) consistently lead to a better transfer accuracy than the other (\eg WiT) across many downstream tasks. However, the ranking of the other pre-training datasets in our selection appears mixed. Moreover, even the pre-training dataset which leads to the worst transfer accuracy (WiT) still outperforms training from scratch (see~\Secref{sec:pretraingdata}, ~\Figref{fig:shots_vs_acc_violon} and~\Figref{fig:shots_vs_acc_datadist_extend}). 

\paragraph{How much is expensive labeling worth compared to noisier but larger pre-training data?} We compare different pre-training strategies: supervised pre-training on the carefully labeled ImageNet dataset and semi-supervised pre-training on language-image pairs from larger but noisier datasets. We find that pre-training on a well-curated dataset leads to better transfer accuracy than pre-training on a noisy dataset of a similar size. Our investigations also show that pre-training on a 15x-2000x larger but noisier dataset (LAION) can close the gap for some downstream tasks (see~\secref{sec:ImageNet_sup},~\secref{sec:in_worth},~\Figref{fig:shots_vs_acc_datacuration} and \Figref{fig:INworth}). 

\paragraph{How does increasing the pre-training dataset size impact the transfer learning performance?}
Can we expect that more pre-training data implies a better performance, or can the pre-training effectiveness saturate at some point?
Our findings indicate that, while larger pre-training datasets tend to improve transfer performance, such improvement depends on the dataset and the number of fine-tuning samples.
We observe that some downstream tasks show little improvement even with large datasets, while others benefit significantly.
(see~\secref{sec:datasetsize} and ~\Figref{fig:shots_vs_acc_datasetsize}).


\paragraph{What is the impact of the pre-training method on downstream performance?} We examine the difference between supervised pre-training with the popular CLIP and SimCLR semi-supervised algorithms. Overall the diminishing effect for pre-training distributions shown in~\Figref{fig:shots_vs_acc_violon} is also observed when the pre-training method is switched from CLIP to SimcLR in~\Figref{fig:shots_vs_acc_simclr},  meaning differences in downstream  accuracy in largely observed in the few-shot setting. 
In addition, we find that the SimCLR pre-training leads to better transfer than CLIP pre-training in the few-shot regime. However, the  difference is small if many images are used for fine-tuning (see~\secref{sec:pretrainloss} and ~\Figref{fig:shots_vs_acc_simclrvsclip}). 

We conduct an extensive empirical investigation in the context of transfer learning for computer vision tasks (See~\secref{sec:trainingdetails} for details on 4000 experiments). Our study covers seven pre-training datasets including YFCC, LAION, Redcaps, Conceptual Captions-3m, Conceptual Captions-12m, WiT, Shutterstock, and ImageNet~\citep{thomee2016yfcc100m, schuhmann2021laion, desai2021redcaps, sharma2018conceptual, changpinyo2021conceptual, srinivasan2021wit, deng2009imagenet},
nine fine-tuning datasets including
CIFAR100, DTD, Caltech-101, Oxford-PETS, REAL and CLIPART from DomainNet, EuroSAT, Cassava Leaf Disease, and Caltech Camera Traps~\citep{cifar100,cimpoi2014describing,fei2004learning,parkhi2012cats,peng2019moment,helber2019eurosat,Cassavaleafdisease2021,beery2018recognition}, and three pre-training methods: supervised, CLIP~\citep{radford2021learning} and SimCLR~\citep{chen2020simple}. To evaluate downstream performance, we examine both few-shot fine-tuning and full fine-tuning.

The paper is structured as follows: we review closely related works in Section~\ref{sec:related}, followed by our experimental setup presented in Section~\ref{sec:setup}. Section~\ref{sec:exp} details our observations relating to our research questions by measuring the downstream transfer accuracy of models pre-trained on various data sources, dataset sizes, and with different pre-training losses. We discuss our findings and conclude with future research directions in Section~\ref{sec:discussion}.

\section{Related Work}
\label{sec:related}
This work is inspired by and closely related to~\citet{kim2022broad} and~\citet{ abnar2021exploring}.~\citet{kim2022broad} conducted an in-depth study of the effect of the network architecture, pre-training dataset, supervised vs self-supervised learning objectives, and different domain transfer methods on the transferability of representations to new domains. They found that the transferability of the pre-trained representations depends on factors such as the target benchmark, adaptation method, and network depth.  However, they do not study few-shot transfer (where we see the most impact of the pre-training distribution). They also did not provide a set of controlled experiments for some of the studied impacting factors because they are limited to available pre-trained models. For example, when comparing the role of data distribution (their Figure 2, ImageNet-22K vs. JFT-300m), they change the dataset size and also architecture, and the reader is left wondering if JFT has a better distribution for transfer or if the observed effects come from more data or a better architecture? 

\citet{abnar2021exploring} also explored how different upstream training settings affect transfer accuracy for two upstream datasets and more than 20 downstream tasks. They showed that as the upstream accuracy increases, the transfer learning performance on downstream datasets saturates. 
However, the authors study only upstream models that are pre-trained with a supervised loss function on ImageNet-21K~\citep{deng2009imagenet} or JFT-300M~\citep{sun2017revisiting} (different size and distributions). 
In this work, we extend these results to more pre-training datasets and methods, with a special focus on data distribution and curation. \citet{abnar2021exploring} also lacks controlled comparison between different distributions in the pre-training datasets~\eg they compare JFT and ImageNet with very different sample sizes.
We consider full fine-tuning in addition to few-shot transfer. Moreover,~\cite {you2021logme,nguyen2020leep,deshpande2021linearized} develop metrics for predicting the transferability of a model. Their main focus is to develop a measure to predict the full fine-tune accuracy without actually fine-tuning on the downstream task. While we also cover full fine-tune accuracy, our main research question lies in studying the extent to which pre-training data affect transfer accuracy. Looking at few-shot and full-shot also gives us the ability to study the effect of transfer learning as more target data become available. Moreover, predictability of the transfer performance is mostly limited to supervised ImageNet-1K pretraining, while we scale both pre-training distributions, size, and pre-training loss functions. 
Transferability line of research also mainly focuses on Internet-crawled datasets, while we extended our results to domain-specific datasets (Camera Traps, Cassava Leaf Diseases, and EuroSAT),
~\Secref{sec:extend_related} extends related works.

\begin{table}[t]
\centering
\begin{tabular}{c c c c c c c c c c |c}
 & \rotatebox{90}{CIFAR100} & \rotatebox{90}{DTD} & \rotatebox{90}{CALTECH} & \rotatebox{90}{PETS} & \rotatebox{90}{DomainNet} \rotatebox{90}{REAL} & \rotatebox{90}{DomainNet} \rotatebox{90}{CLIPART} & \rotatebox{90}{EuroSAT} & \rotatebox{90}{Cassava Leaf} \rotatebox{90}{Disease} & \rotatebox{90}{Camera Traps} & \rotatebox{90}{average}\\ 
 \hline
Train from scratch \\(no pre-training) & 72.82 & 44.62 & 55.32 & 67.96 & 77.76 & 55.20 & 58.50 & 70.30 & 89.10 & 65.73\\
 \hline
Worst performing \\pre-training & 81.62 & 61.06 & 82.38 & 83.15 & 81.87 & 68.98 & 70.72 & 87.05 & 98.85 & 79.52\\
 \hline
Best performing \\pre-training & 83.71 & 68.56 & 86.88 & 87.84 & 82.92 & 72.38 & 72.79 & 87.57 & 98.91 & 82.39\\
 \hline
\end{tabular}
\caption{Full fine-tune accuracy from six different pre-training datasets across all downstream datasets. Transfer learning from even the worst pre-training dataset outperforms training from scratch largely.}
\label{table:1}\label{tab:scratch_vs_worst}
\end{table}

\section{Experimental setup} 
\label{sec:setup}
\paragraph{Model}
The main focus of this study is the CLIP model~\citep{radford2021learning}. This model has demonstrated unprecedented robustness to natural distribution shifts~\citep{taori2020measuring, miller2021accuracy}, and transfers well to many downstream tasks~\citep{radford2021learning, wortsman2021robust}.
Given an image-text pair, CLIP learns a joint embedding space for both images and their captions and tries to maximize the cosine similarity between the text and image embedding for an image relative to the cosine similarity of unaligned pairs. We use the CLIP implementation from the OpenCLIP GitHub repository~\citep{ilharco_gabriel_2021_5143773}.

\paragraph{Pre-training} We mainly use ResNet-50~\citep{he2016deep} as the image encoder unless stated otherwise. We vary the pre-training data distribution in ~\secref{sec:pretraingdata}, curation method in \secref{sec:ImageNet_sup}, and pre-training dataset size in \secref{sec:datasetsize} to obtain different pre-trained models. We also change the contrastive loss function to SimCLR in \secref{sec:pretrainloss} to test the effect of the pre-training method on downstream transfer accuracy. Further training details are in Appendix~\ref{sec:trainingdetails}.

\paragraph{Fine-tuning} For most of the experiments we fine-tune the pre-trained model end-to-end on the target transfer dataset unless stated otherwise. For each pre-trained model and downstream transfer dataset, we used a large grid search over various fine-tuning hyperparameters including learning rate, batch size, and the number of epochs. We report the best-performing accuracy in the plots. Further training details are in Appendix~\ref{sec:trainingdetails}. 

\paragraph{Datasets}
Our large-scale experiments yield more than 4000 trained networks. Our pre-training datasets consist of  million-size image and language pairs from multiple recent multi-modal datasets including YFCC, LAION, RedCaps, Shutterstock, Conceptual Captions, WiT~\citep{thomee2016yfcc100m, schuhmann2021laion,desai2021redcaps, sharma2018conceptual, changpinyo2021conceptual, srinivasan2021wit}. Our pre-training datasets are crawled from different sources covering different data distributions (See Appendix~\secref{sec:datasets} for details of pre-training sources and their samples).
For downstream tasks, we use nine different datasets CIFAR100, DTD, Caltech-101, Oxford-PETS, REAL, and CLIPART from DomainNet, EuroSAT, Cassava Leaf Disease, and Caltech Camera Traps~\citep{cifar100,cimpoi2014describing,fei2004learning,parkhi2012cats,peng2019moment,helber2019eurosat, Cassavaleafdisease2021,beery2018recognition}. While the first six datasets
are internet-crawled datasets (similar to pre-training datasets) and are more common in transfer learning in computer vision benchmarks, we include three
new downstream datasets that are domain-specific, i.e. the dataset is created after a specific challenge is defined in a specific
domain. See Appendix~\secref{sec:datasets} for more details on downstream datasets.

\section{Experiments and Results} 
\label{sec:exp}
In this section, we ask a set of research questions to investigate the role of the pre-training data, the choice of the pre-training method (supervised vs. CLIP vs. SimCLR), and the impact of the ImageNet distribution on the downstream performance. To answer these questions, we carefully design experiments for each section while ablating other impacting factors. 

\subsection{What is the impact of different pre-training data sources on transfer learning?}
\label{sec:pretraingdata}
Do we expect different distributions to perform differently in the transfer setting?
\Figref{fig:shots_vs_acc_violon} aggregates transfer performance from different pre-training datasets across all downstream datasets. To get each point, we (1) pre-train CLIP models using a set of seven large sources, (2) fine-tune each pre-trained model on all downstream datasets across different shots (a sweep over multiple hyperparameters, see Appendix~\ref{sec:trainingdetails}), and (3) for each downstream dataset, calculate the difference between the best and worst fine-tune performance among used pre-training sources, normalized by the maximum fine-tune performance. \Figref{fig:shots_vs_acc_violon} aggregates over all downstream datasets for each number of shots, highlighting as an example different pre-training models fine-tuned using 20 samples/class on all downstream datasets. We observe that changing the pre-training dataset leads to noticeable differences in the downstream performance in a few-shot setting. However, as more images are available for fine-tuning, the difference in absolute accuracy between different pre-training models is largely diminished. 
\Figref{fig:shots_vs_acc_datadist_extend} in Appendix shows this diminishing effect in detail for different downstream datasets. The full fine-tuned models have very similar downstream performances despite different pre-training datasets (see the top-right point of CIFAR100 and REAL in \Figref{fig:shots_vs_acc_datadist_extend}, and also the top-right point for CameraTraps, Cassava Leaf, and EuroSAT in \Figref{fig:shots_vs_acc_datadist_extend}). However, this is not true for DTD, CALTECH101, PETS, and CLIPART, where they have far fewer images per class for fine-tuning on the full dataset.~\tabref{tab:scratch_vs_worst} compares fine-tune accuracy for different pre-training choices along all downstream datasets. Transfer learning from even the worst pre-training dataset outperforms training from scratch largely. The gap between best  and worst-performing pre-training datasets is small. Appendix~\ref{sec:extend_archs} extends our results to Vision Transformers~\citep{dosovitskiy2021an} instead of ResNet-50.

\begin{figure*}[t]
    \centering
    \subfigure{\includegraphics[width=.32\textwidth]{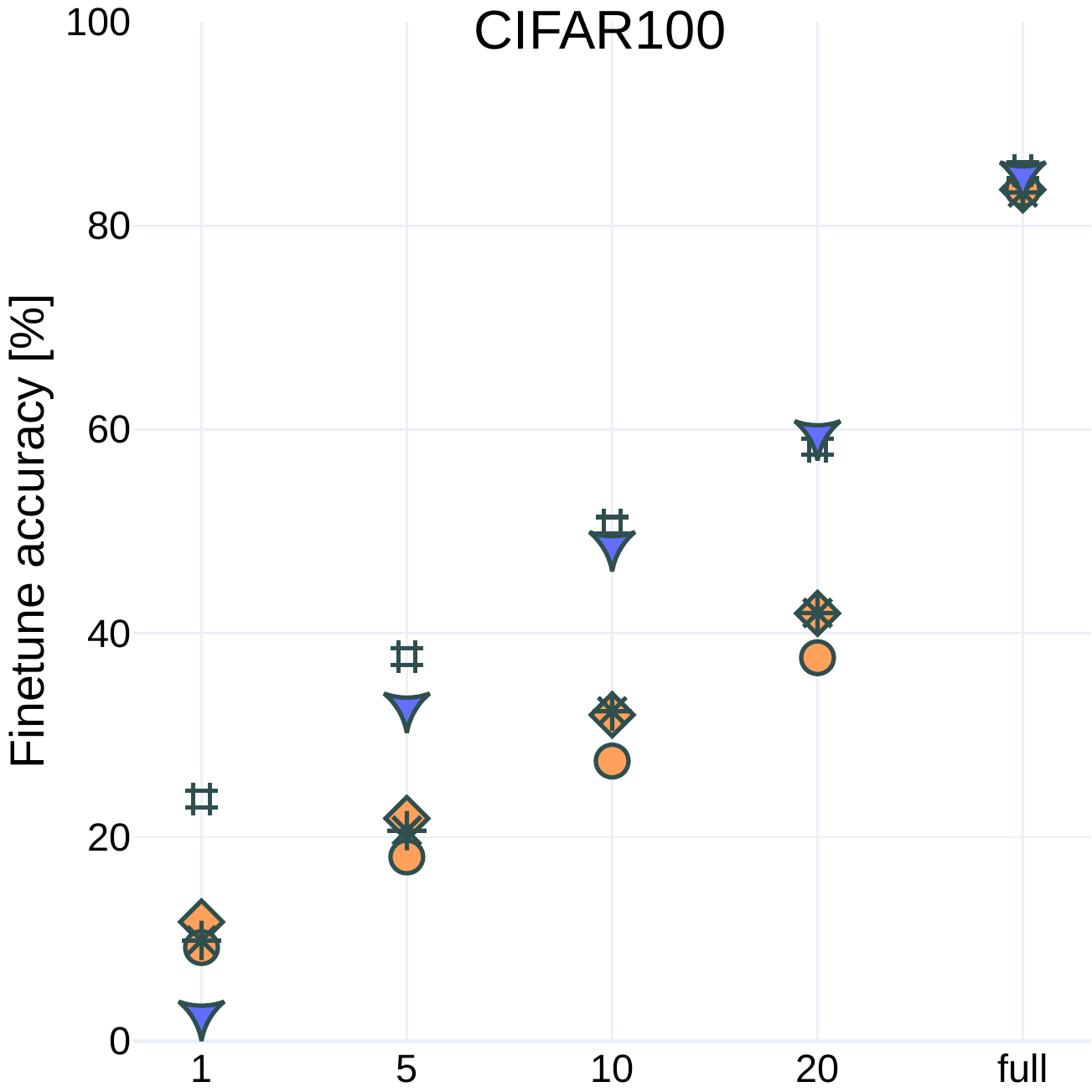}}
    \subfigure{\includegraphics[width=.32\textwidth]{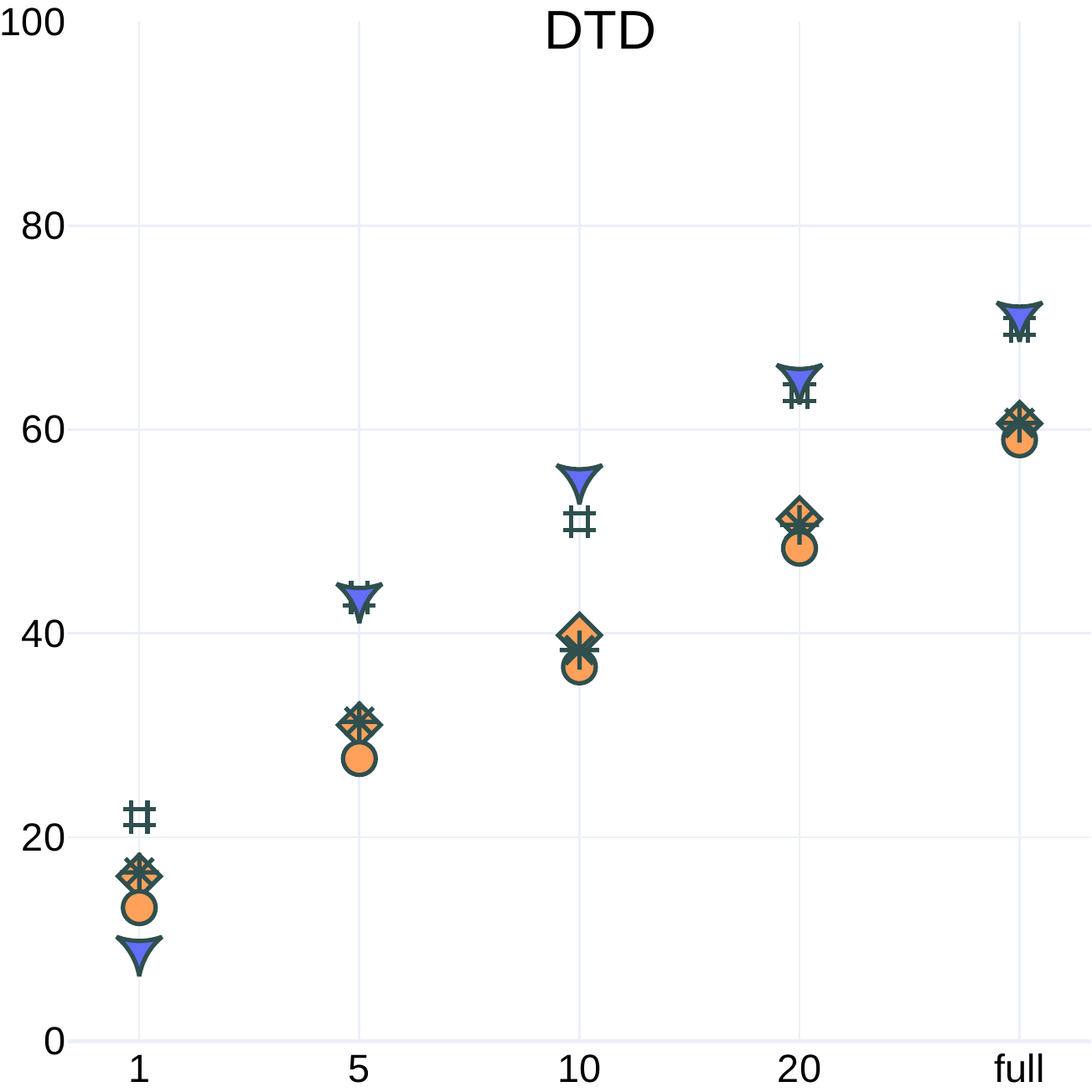}}
    \subfigure{\includegraphics[width=.32\textwidth]{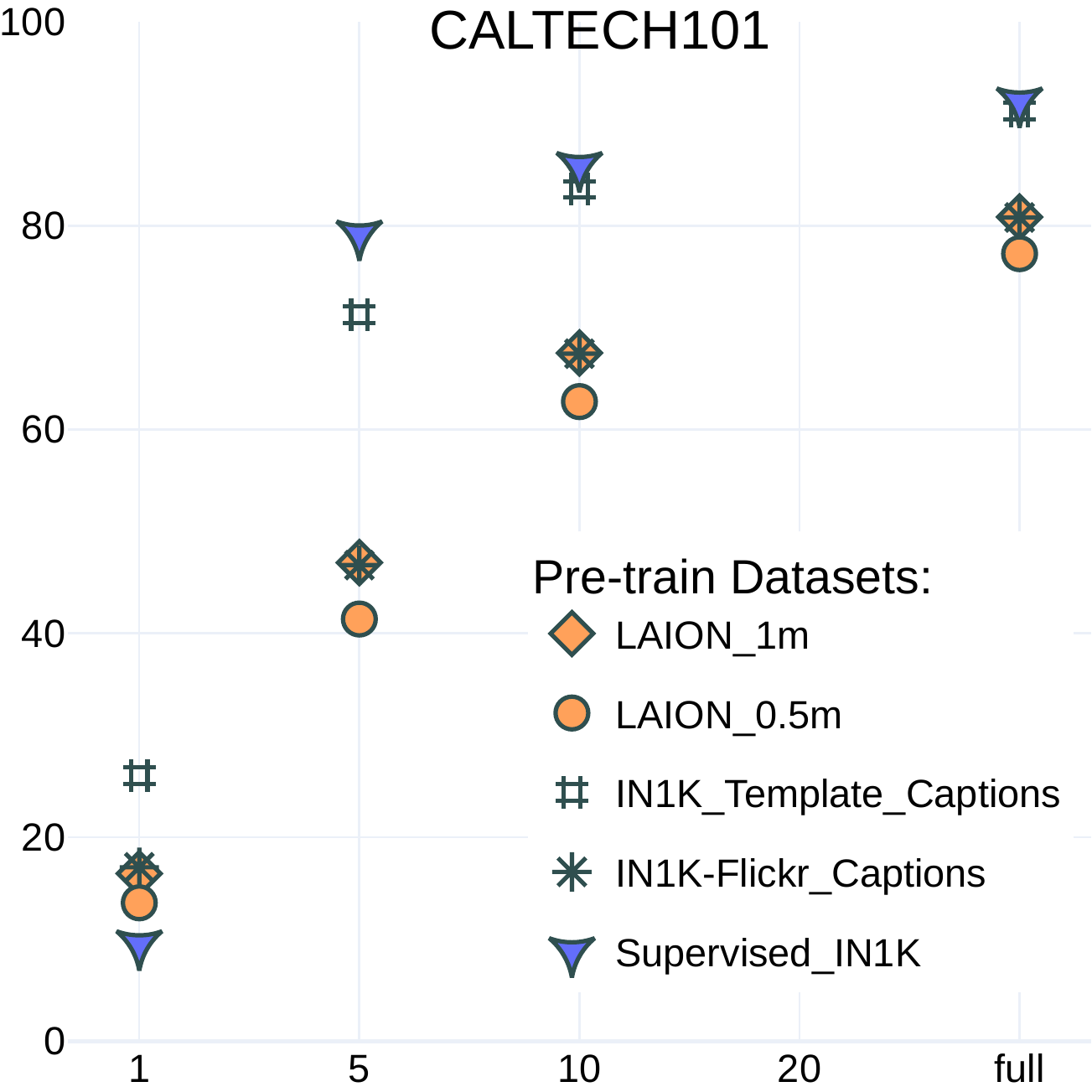}}
    
    \subfigure{\includegraphics[width=.32\textwidth]{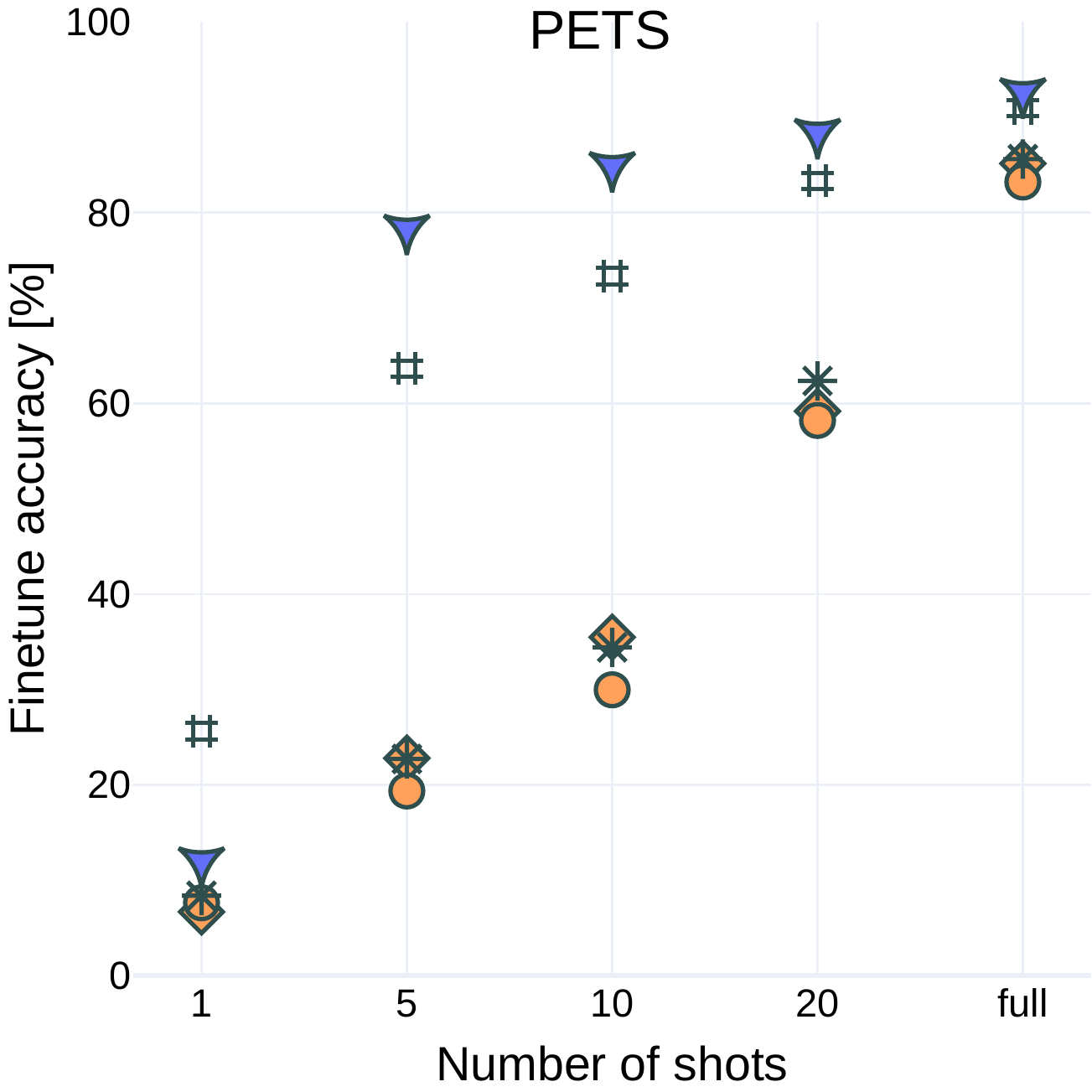}}
    \subfigure{\includegraphics[width=.32\textwidth]{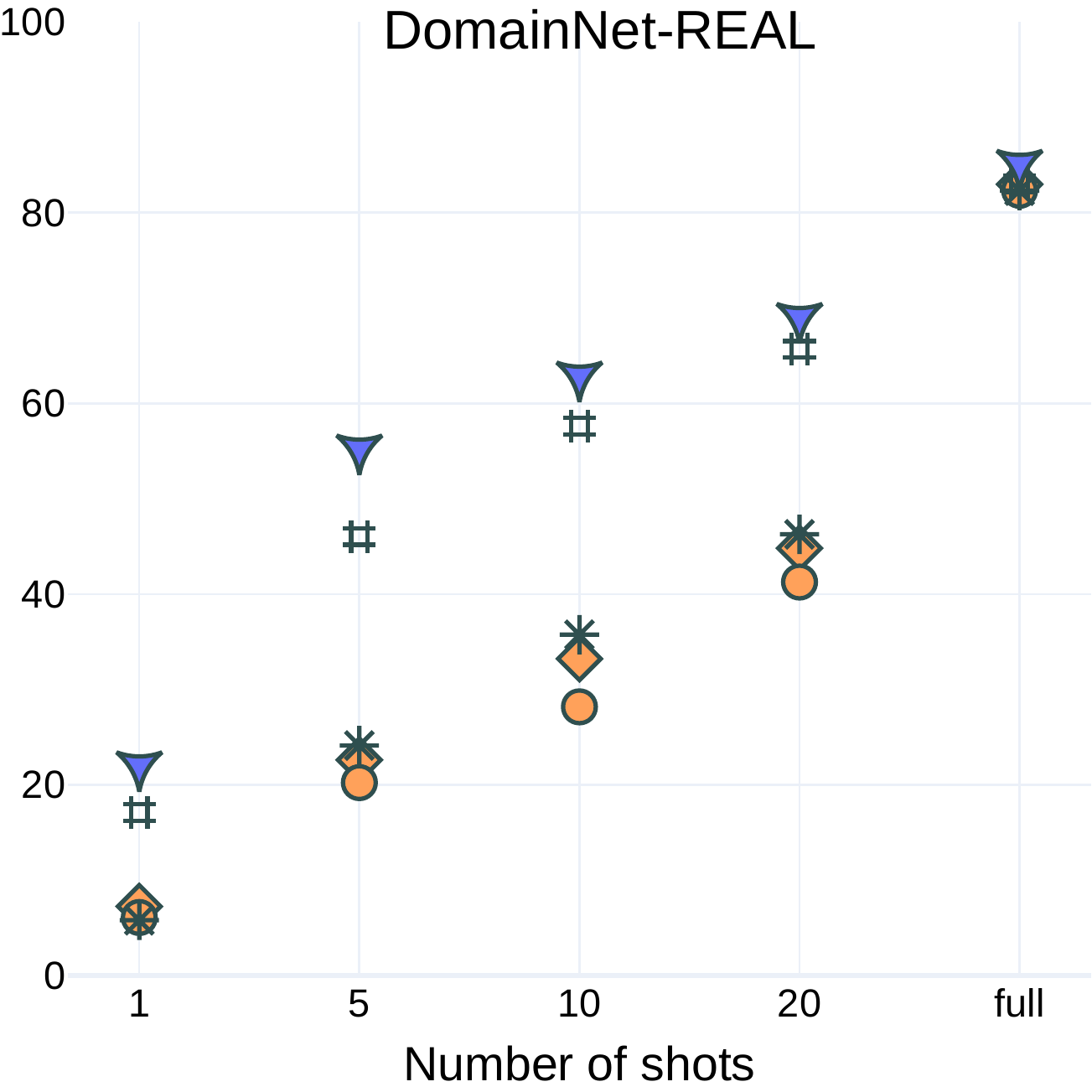}}
    \subfigure{\includegraphics[width=.32\textwidth]{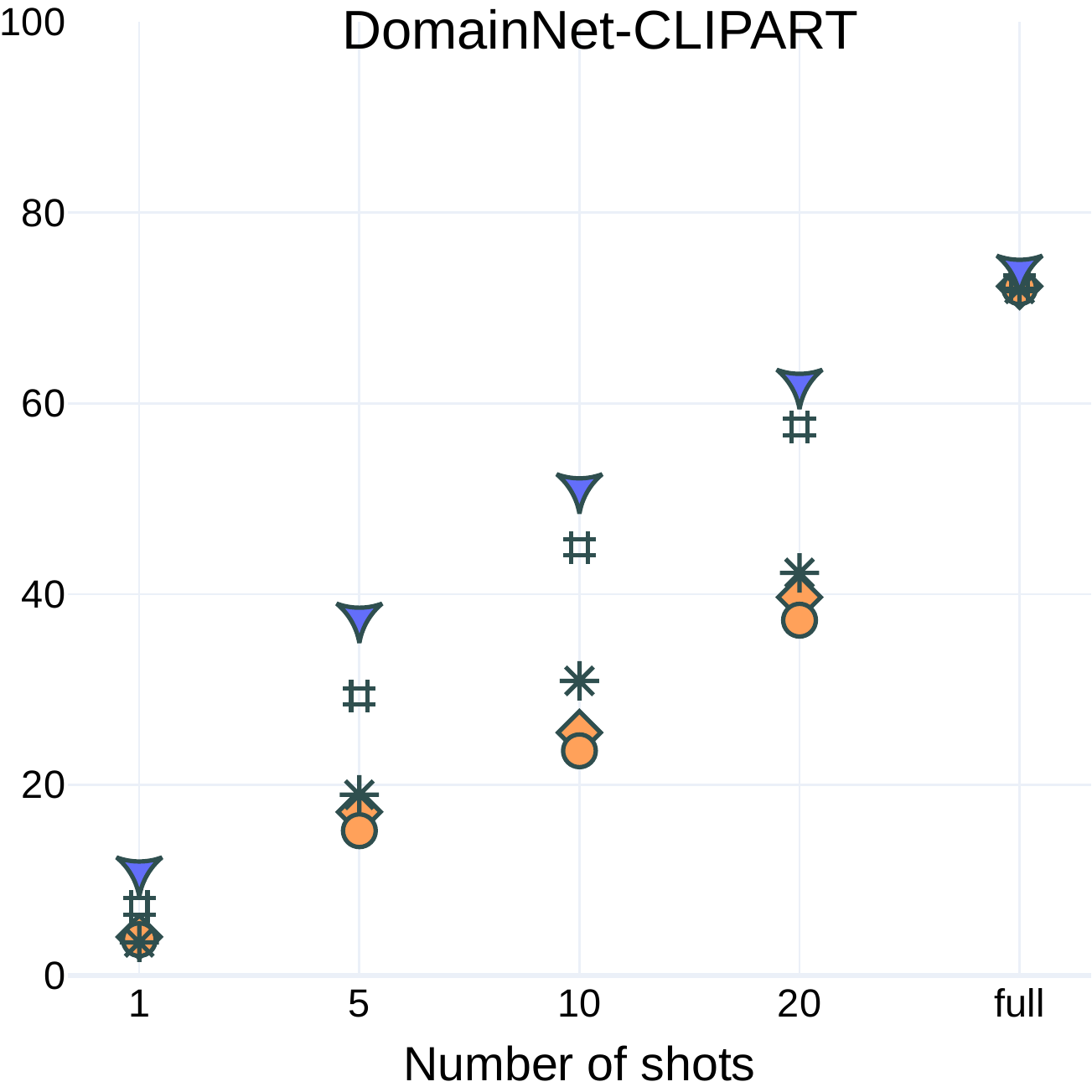}}
    \caption{\textbf{Effect of data curation and labeling.}
    We compare supervised pre-training on ImageNet-1K to (1) contrastive pre-training on original captions from Flickr with 0.5m samples, and (2) contrastive pre-training on Templated (clean) captions using ImageNet labels with 1.2m samples. 
    Supervised pre-training on ImageNet leads to better transfer accuracy than contrastive pre-training. Improving captions quality from Flickr to Template leads to huge improvements in downstream transfer accuracy, highlighting the importance of captions quality. On a different comparison between ImageNet and LAION distributions, pre-training CLIP on Flickr captions performs better than pre-training on LAION distribution with the same size (0.5m).}
    \label{fig:shots_vs_acc_datacuration}
\end{figure*}

\begin{figure*}[t]
    \centering
    \subfigure{\includegraphics[width=.32\textwidth]{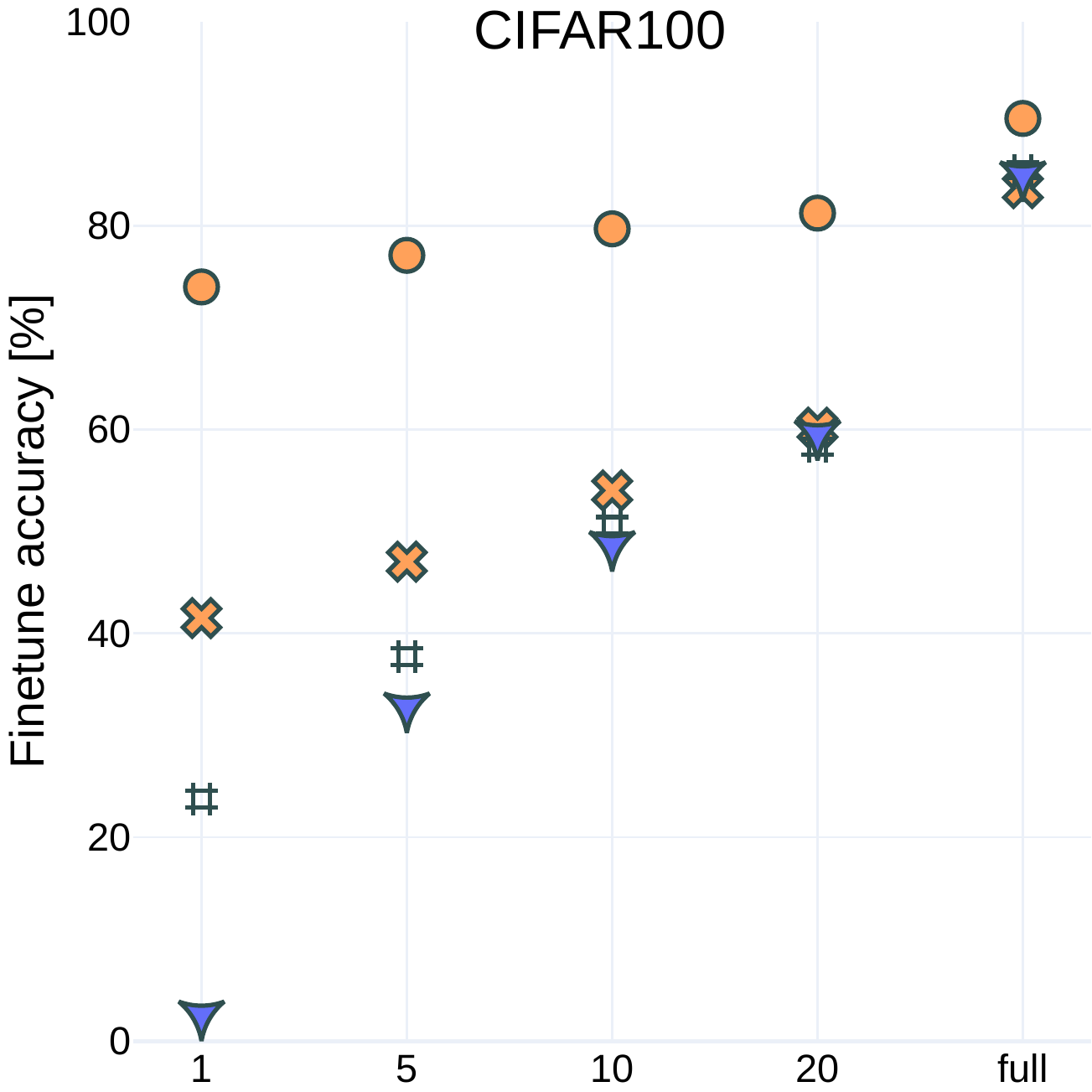}}
    \subfigure{\includegraphics[width=.32\textwidth]{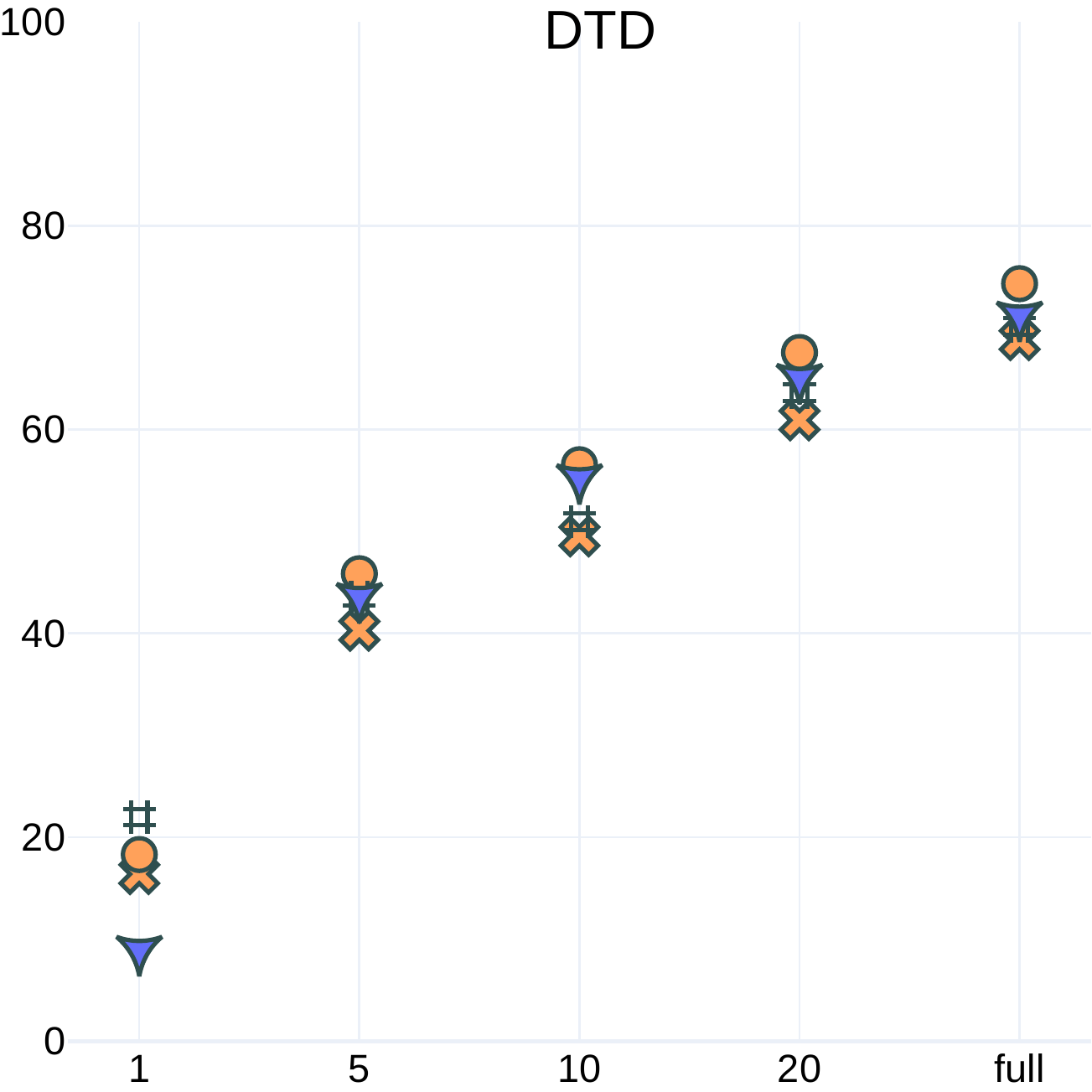}}
    \subfigure{\includegraphics[width=.32\textwidth]{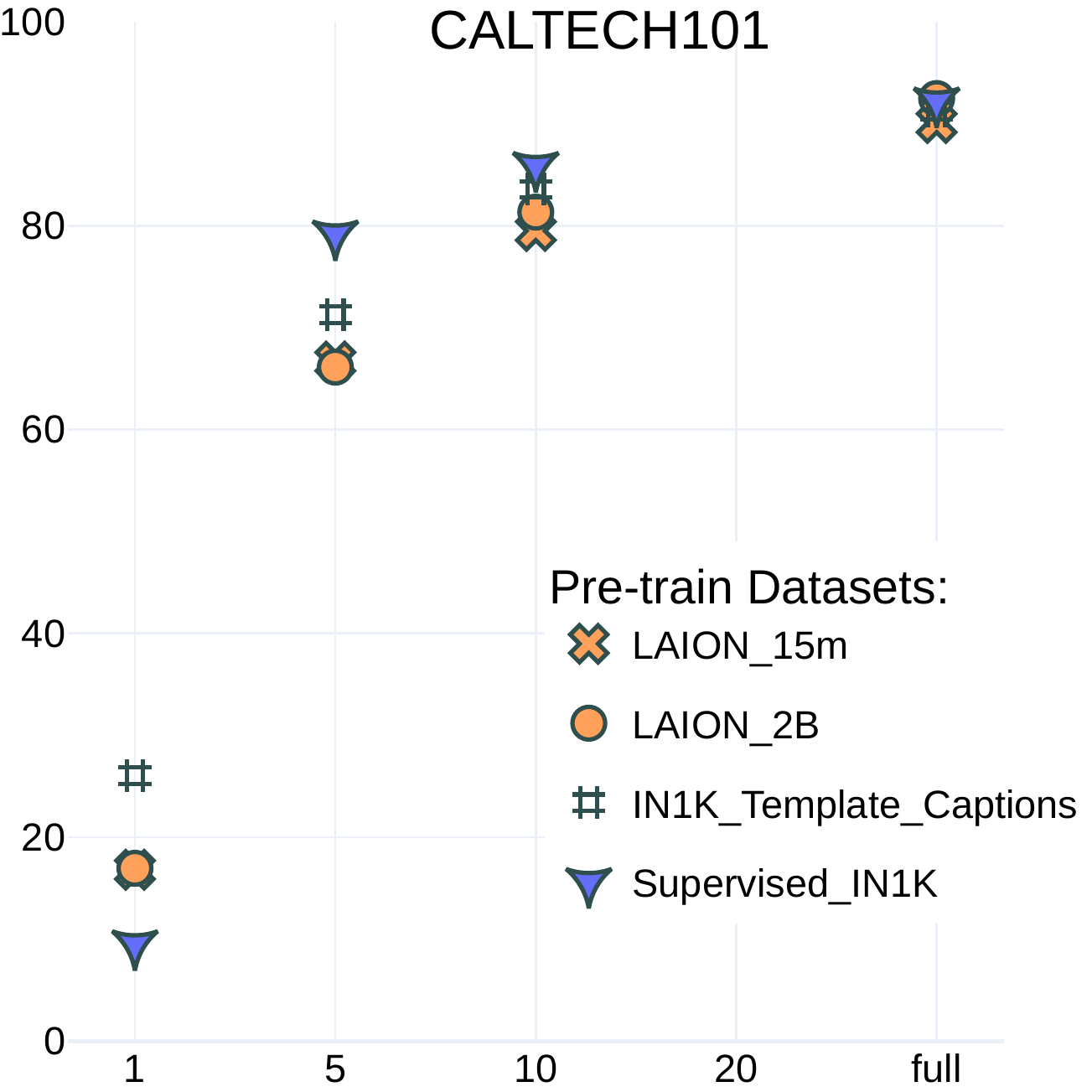}}
    \subfigure{\includegraphics[width=.32\textwidth]{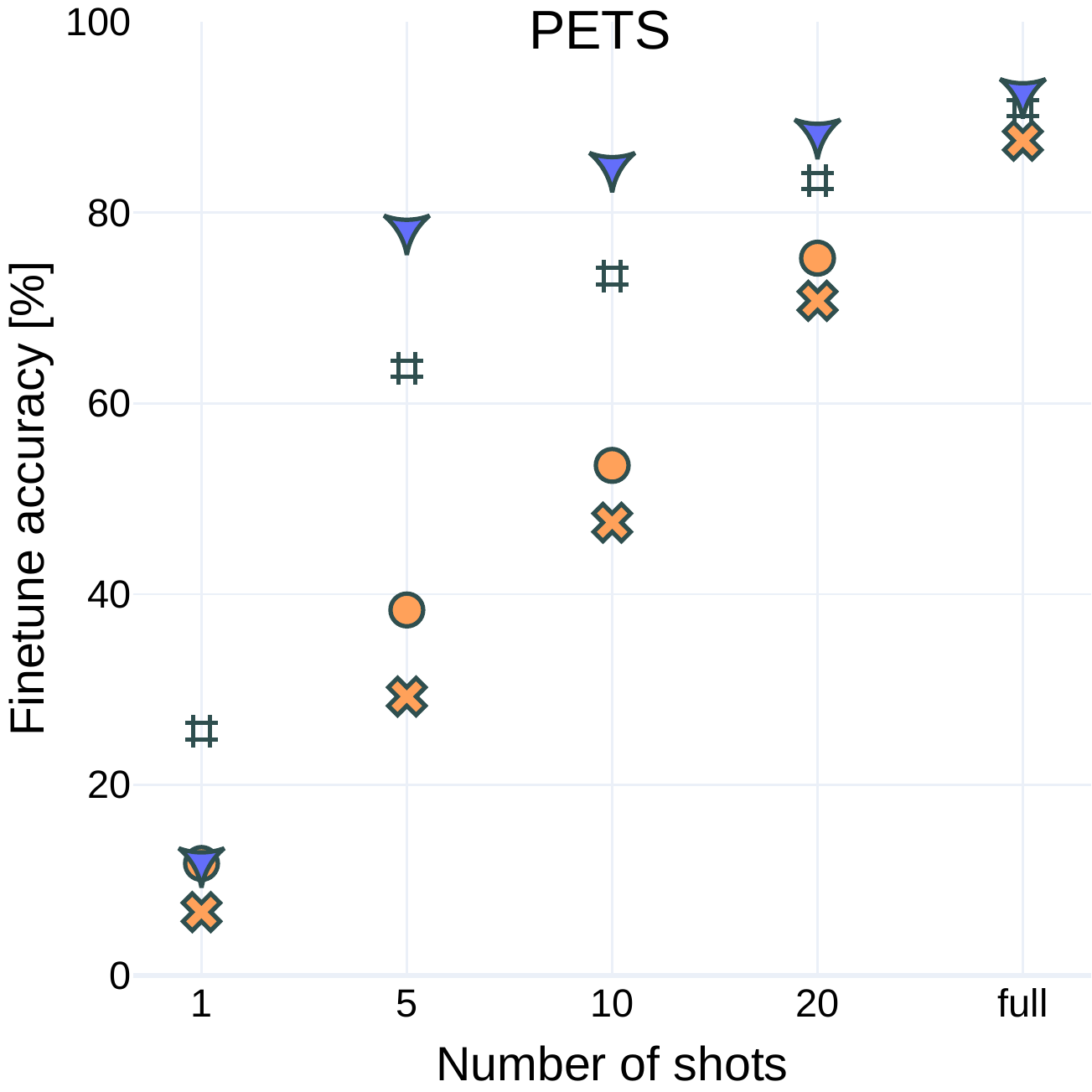}}
    \subfigure{\includegraphics[width=.32\textwidth]{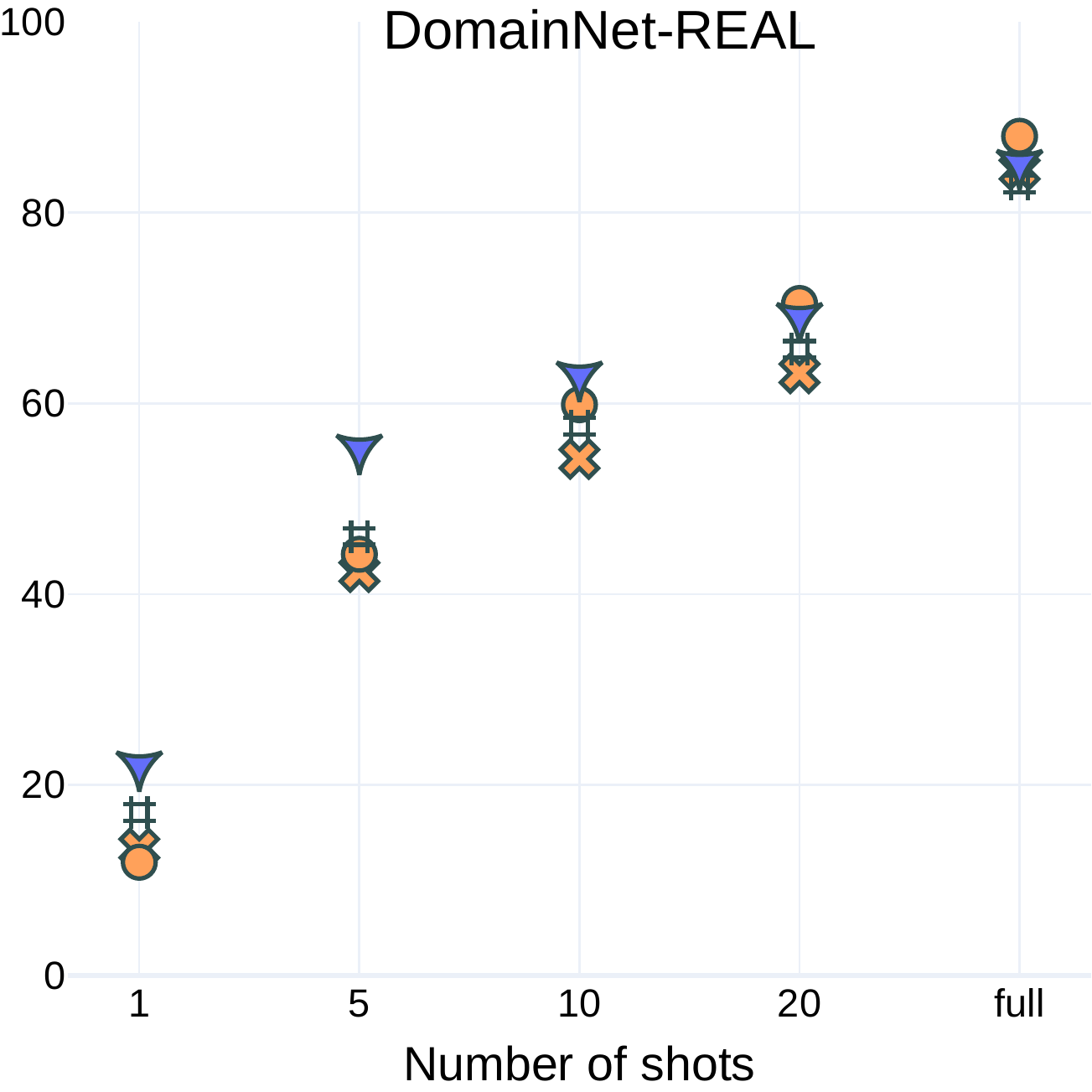}}
    \subfigure{\includegraphics[width=.32\textwidth]{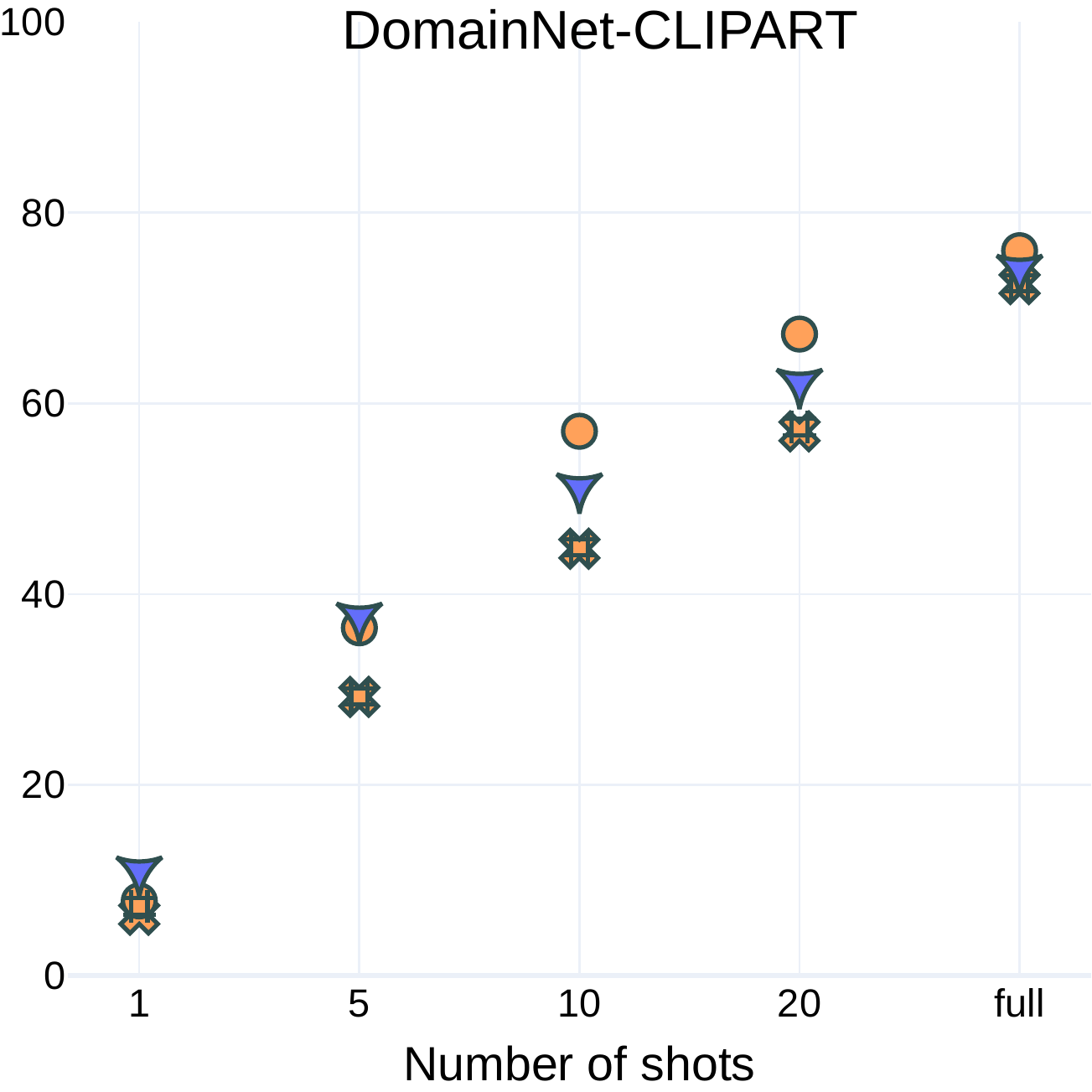}}
    \caption{\textbf{How much LAION data is worth of ImageNet pre-training?} While~\Figref{fig:shots_vs_acc_datacuration} shows the superiority of supervised pre-training over contrastive pre-training with the same size, here we increase the size of contrastive pre-training size to see if contrastive pre-training could perform better than supervised pre-training at scale. Including 15x more data from LAION outperform supervised ImageNet pre-training (and template captions) on CIFAR100. However, DTD, REAL, and CLIPART need 2000x more data from LAION to match or outperform ImageNet pre-training. Even including 2000x more data did not help CALTECH101 and PETS, where supervised ImageNet pre-training is still the best choice.}
    \label{fig:INworth}
\end{figure*}

\begin{figure*}[t]
    \centering
    \subfigure[CIFAR100]{\includegraphics[width=.32\textwidth]{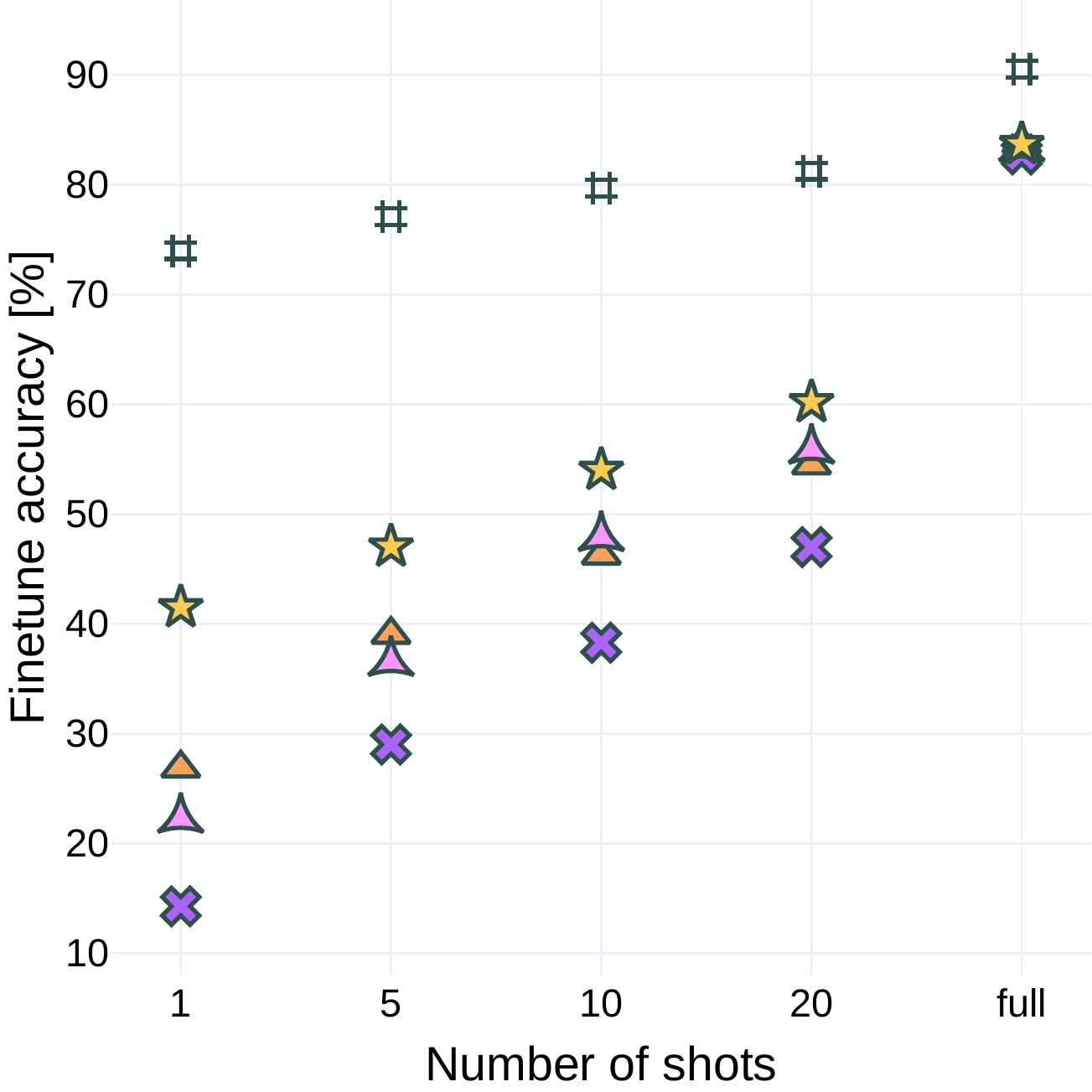}}
    \subfigure[DTD]{\includegraphics[width=.32\textwidth]{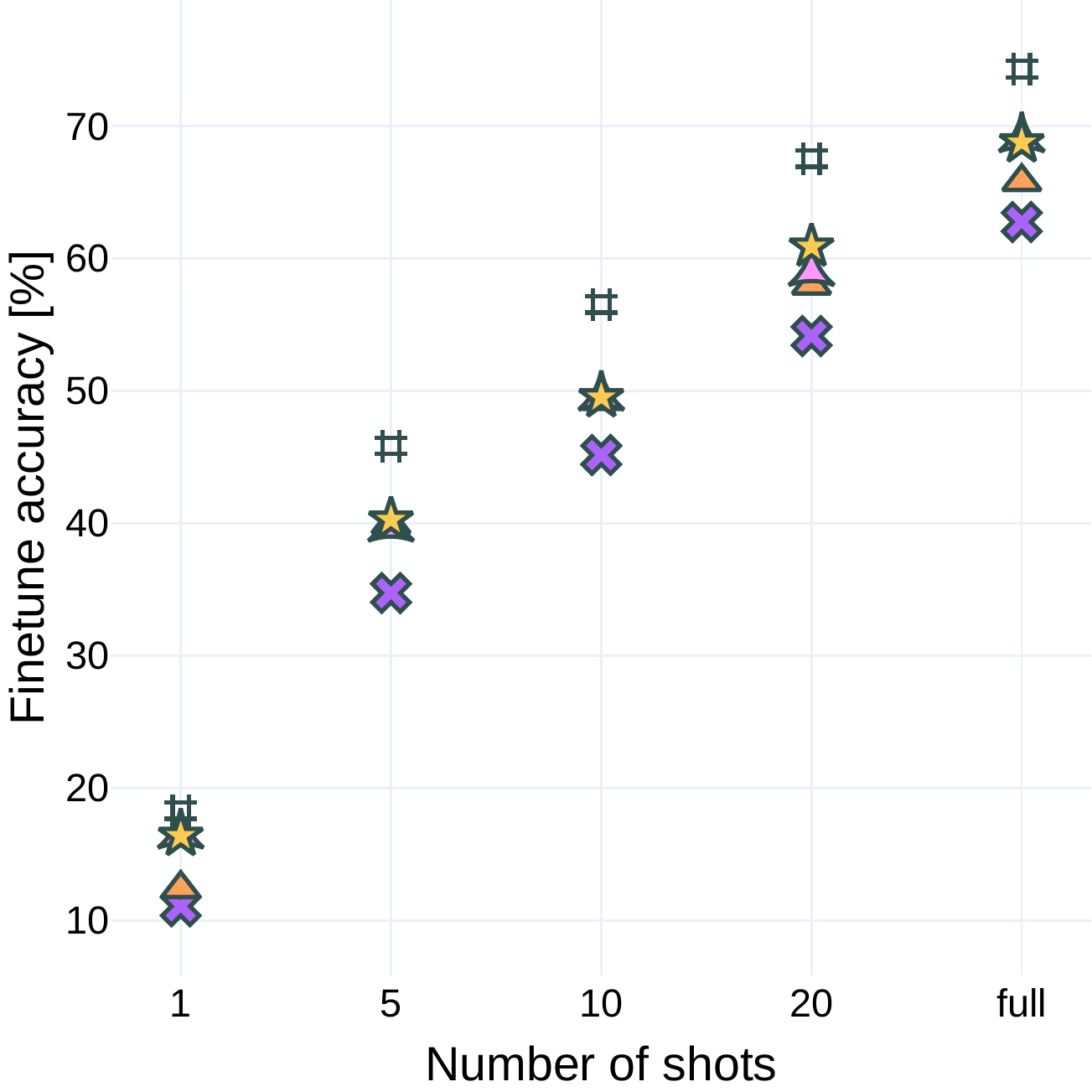}}
    \subfigure[CALTECH101]{\includegraphics[width=.32\textwidth]{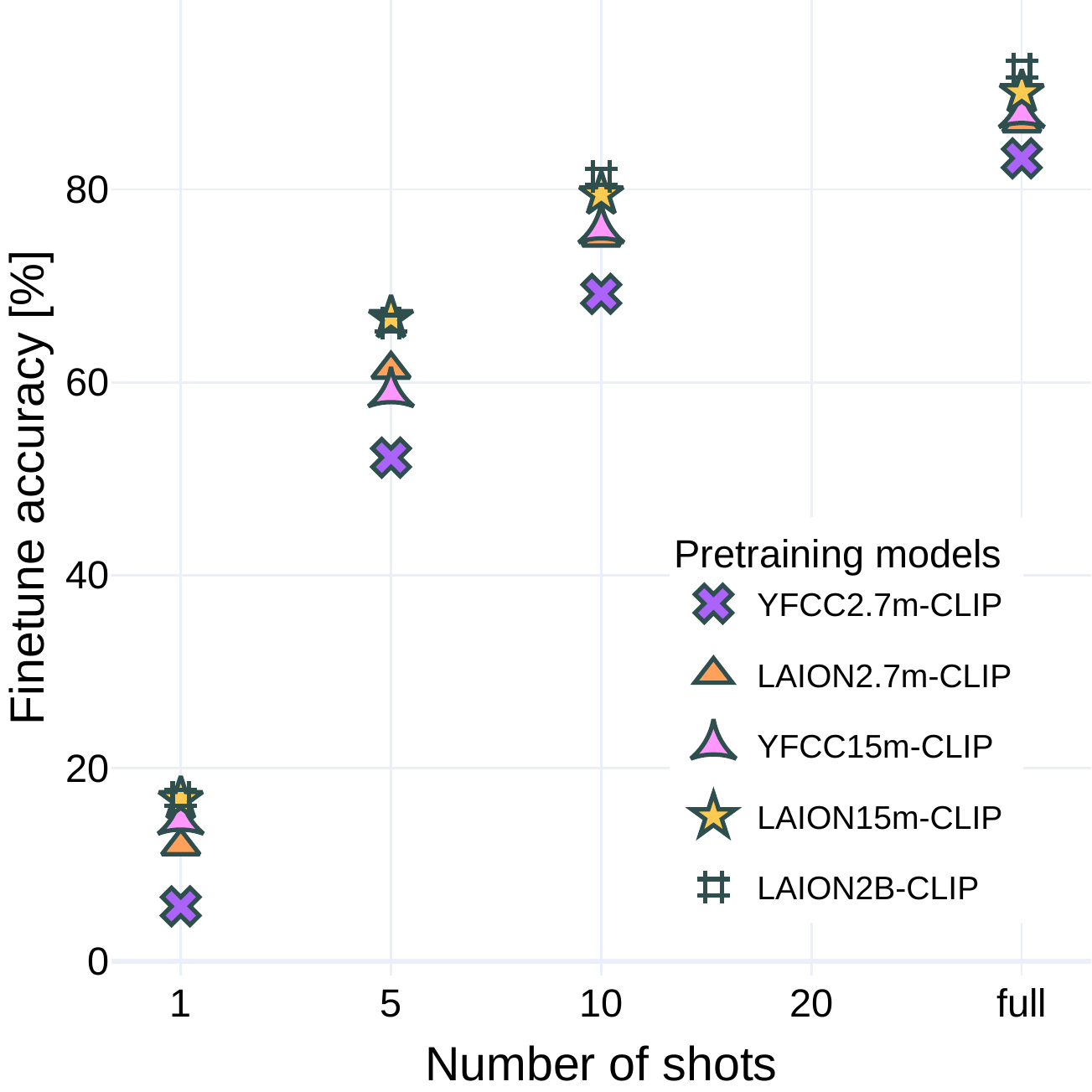}}
    \subfigure[PETS]{\includegraphics[width=.32\textwidth]{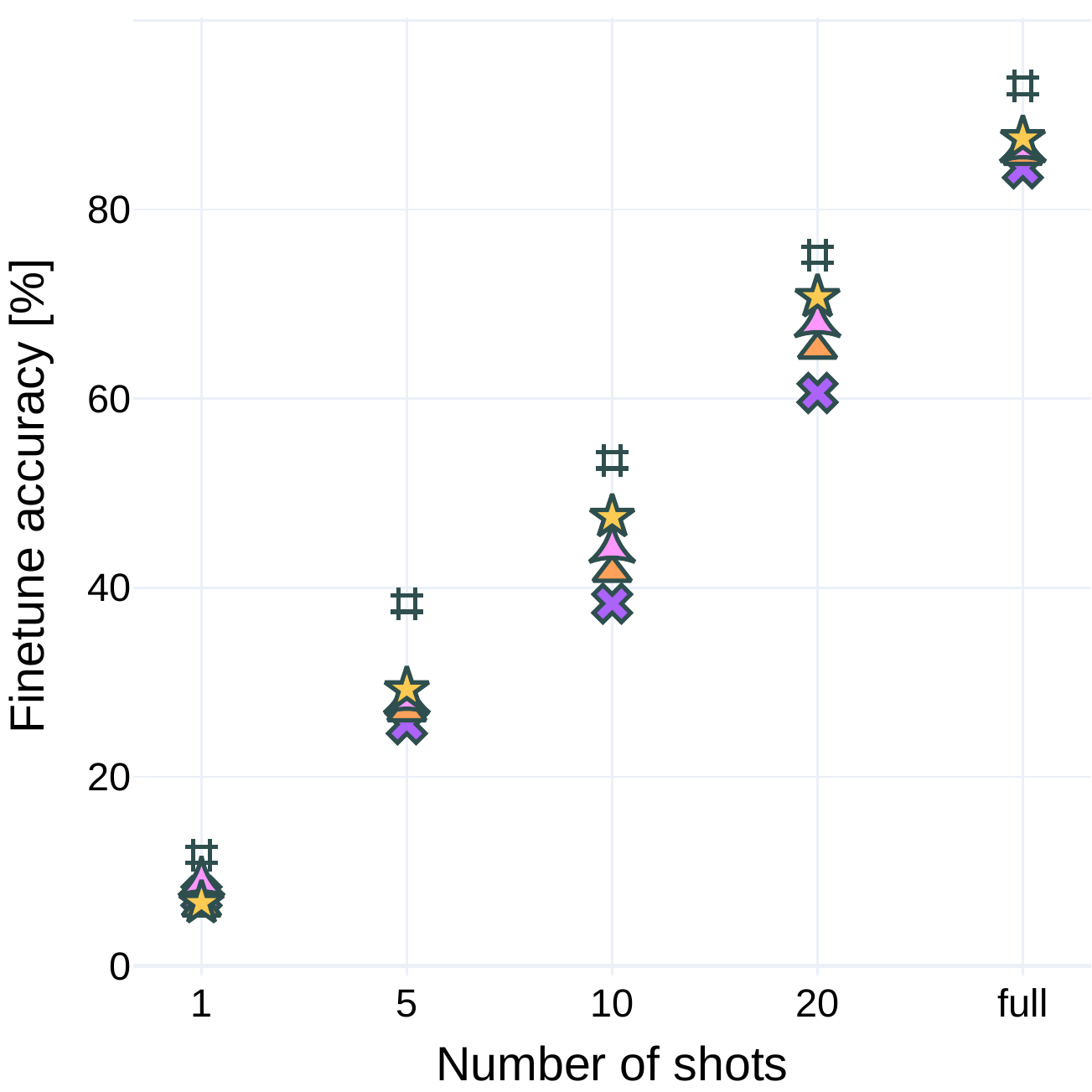}}
    \subfigure[DomainNet-REAL]{\includegraphics[width=.32\textwidth]{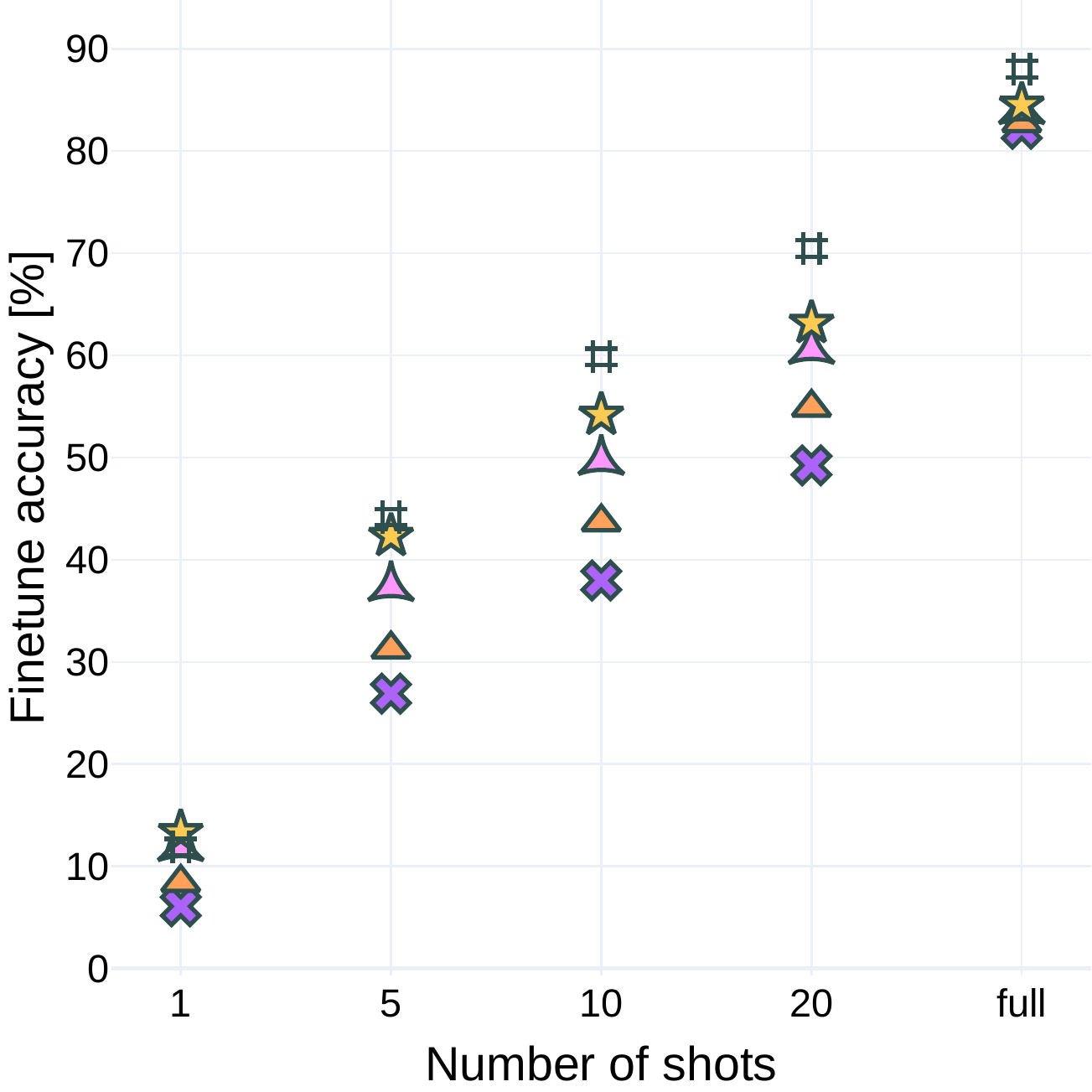}}
    \subfigure[DomainNet-CLIPART]{\includegraphics[width=.32\textwidth]{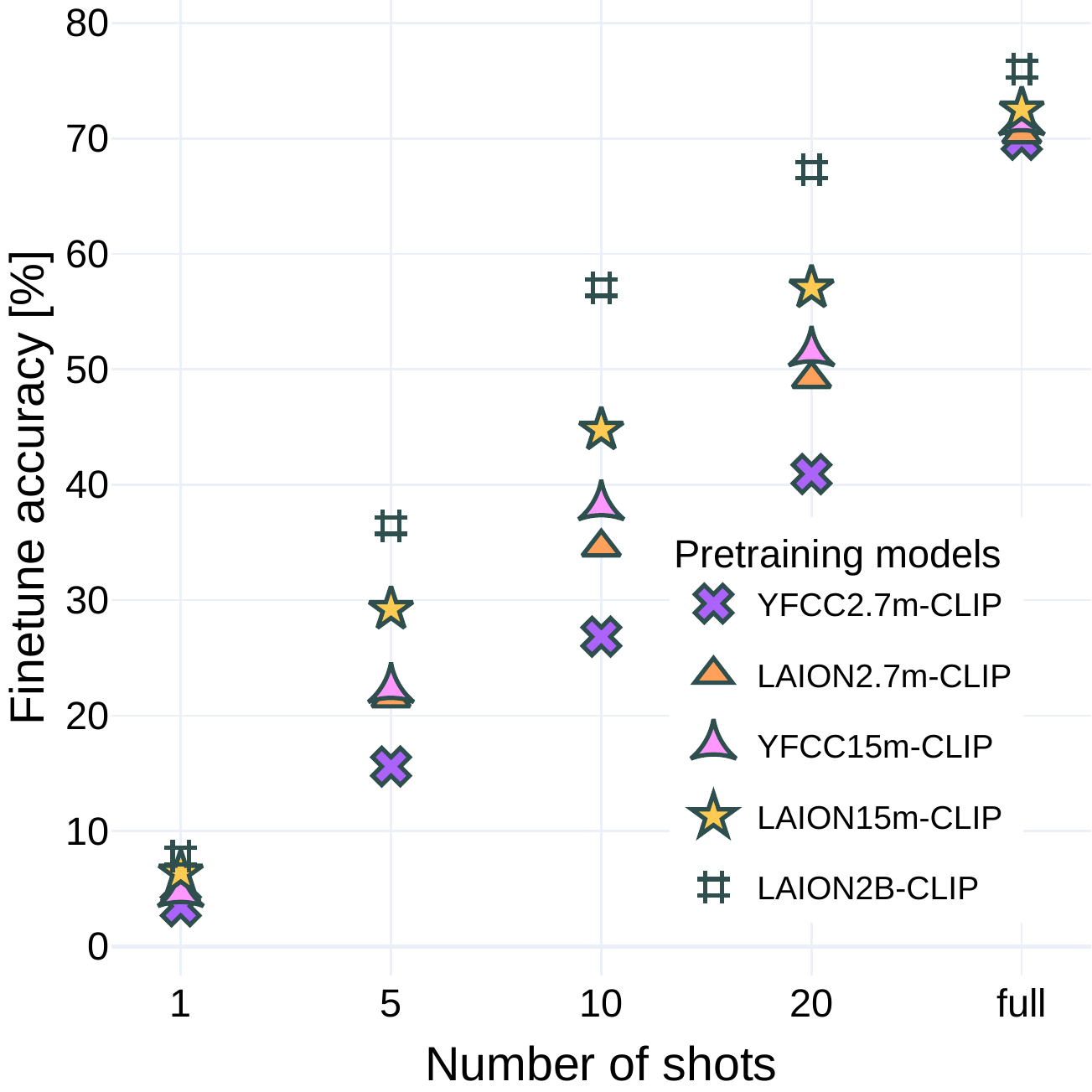}}
    \caption{\textbf{Effect of the pre-training dataset size.} Increasing the size of the dataset used for pre-training results in better transfer accuracy on downstream tasks. However, the absolute accuracy difference is smaller in the high-shot regime, even when pre-training consists of $100\times$ more data. The benefit of pre-training on LAION-2B is different on target tasks. While there is major gap between LAION-2B and LAION-15m for CIFAR100, the performance gain from scaling up the pre-training dataset on CALTECH101 gets saturated.}
    \label{fig:shots_vs_acc_datasetsize}
\end{figure*}

\begin{figure*}[t]
    \centering

    \subfigure[CIFAR100]{\includegraphics[width=.32\textwidth]{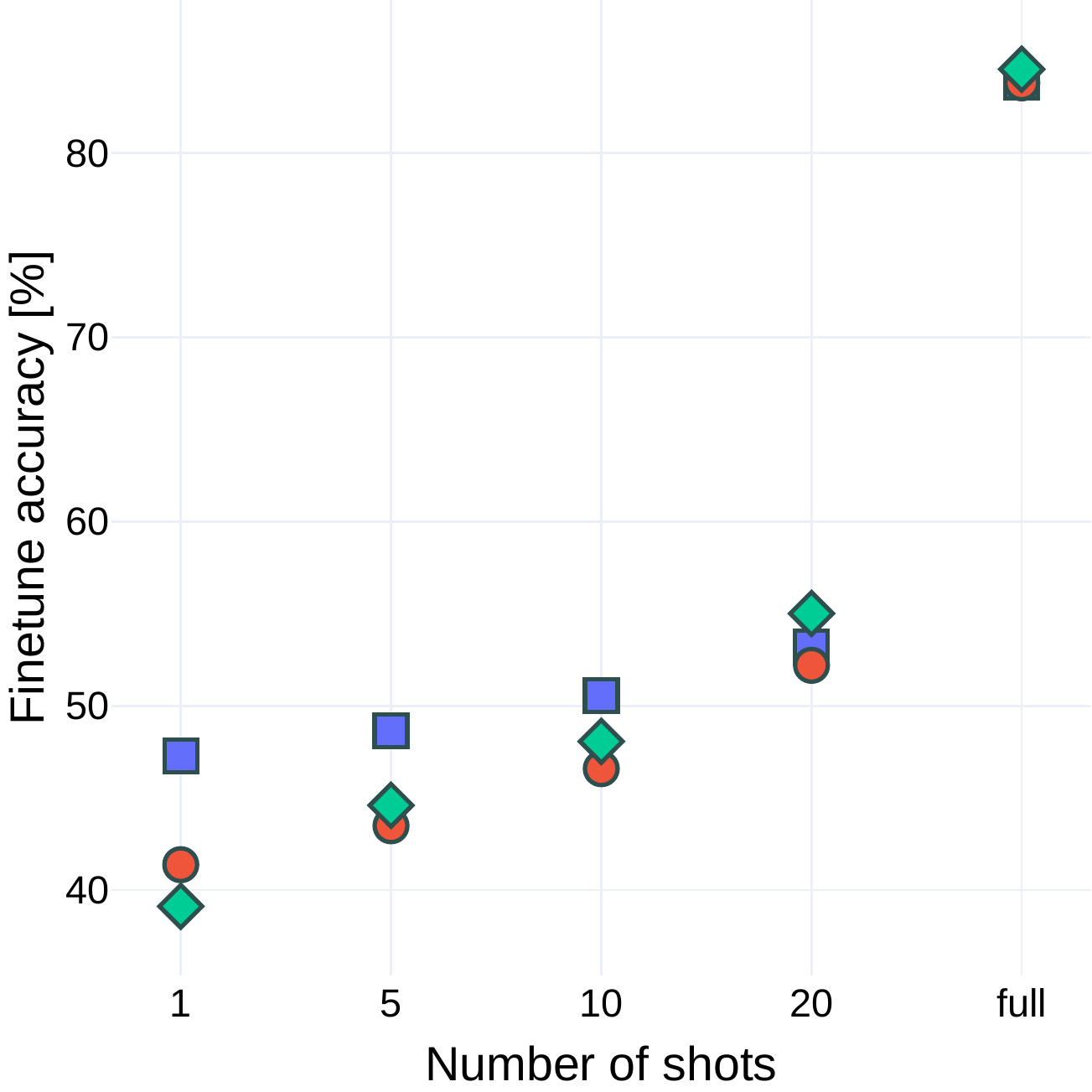}}
    \subfigure[DTD]{\includegraphics[width=.32\textwidth]{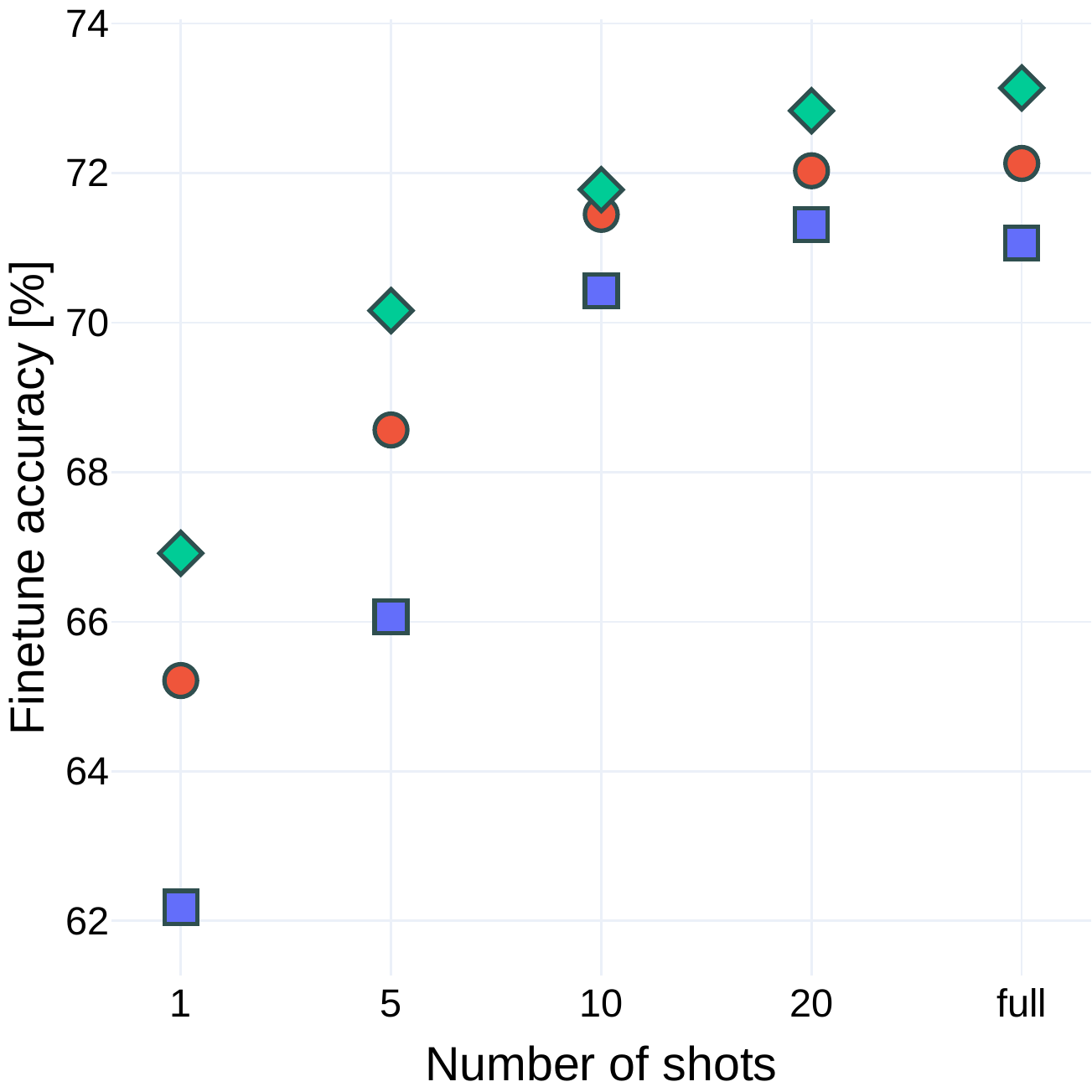}}
    \subfigure[CALTECH101]{\includegraphics[width=.32\textwidth]{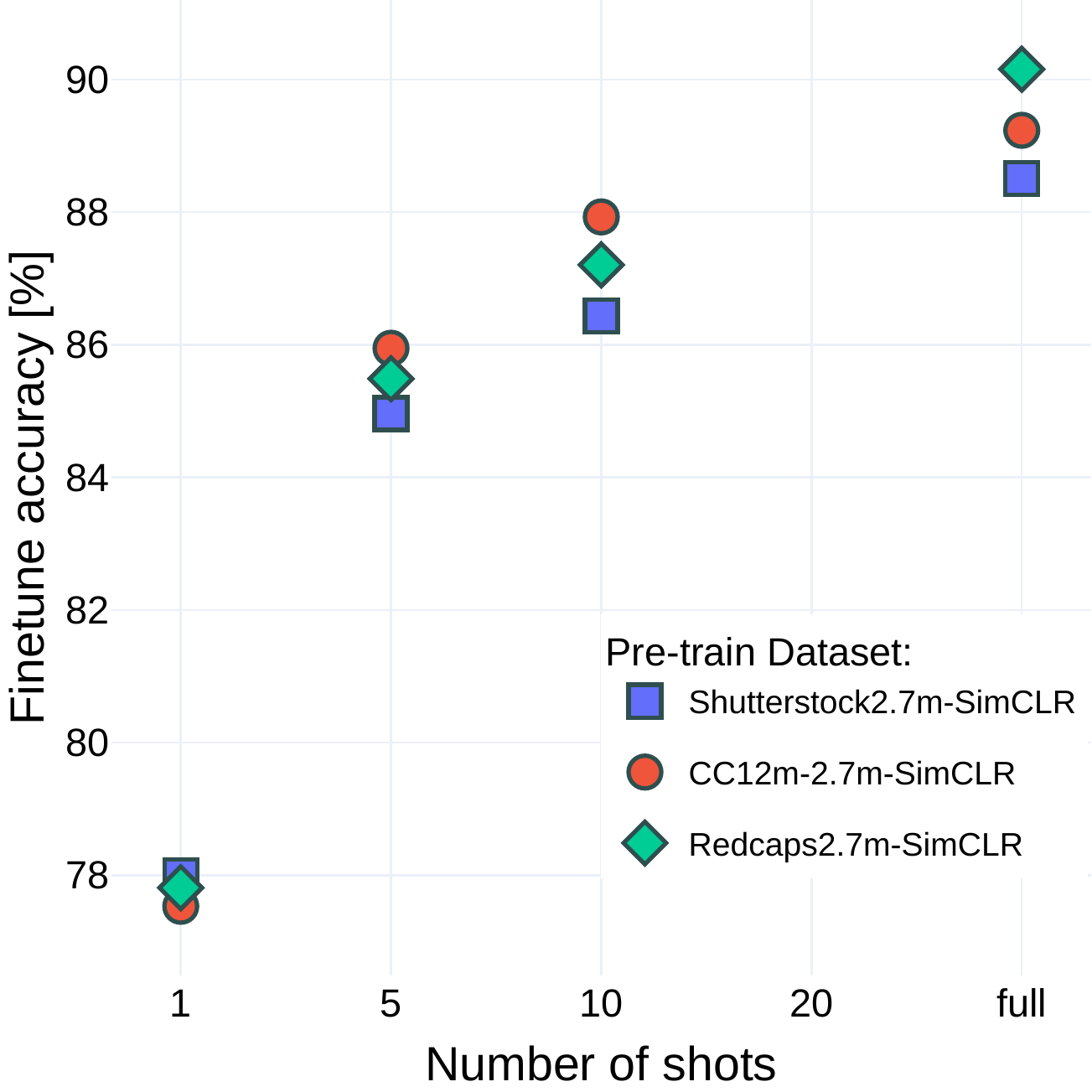}}
    \subfigure[PETS]{\includegraphics[width=.32\textwidth]{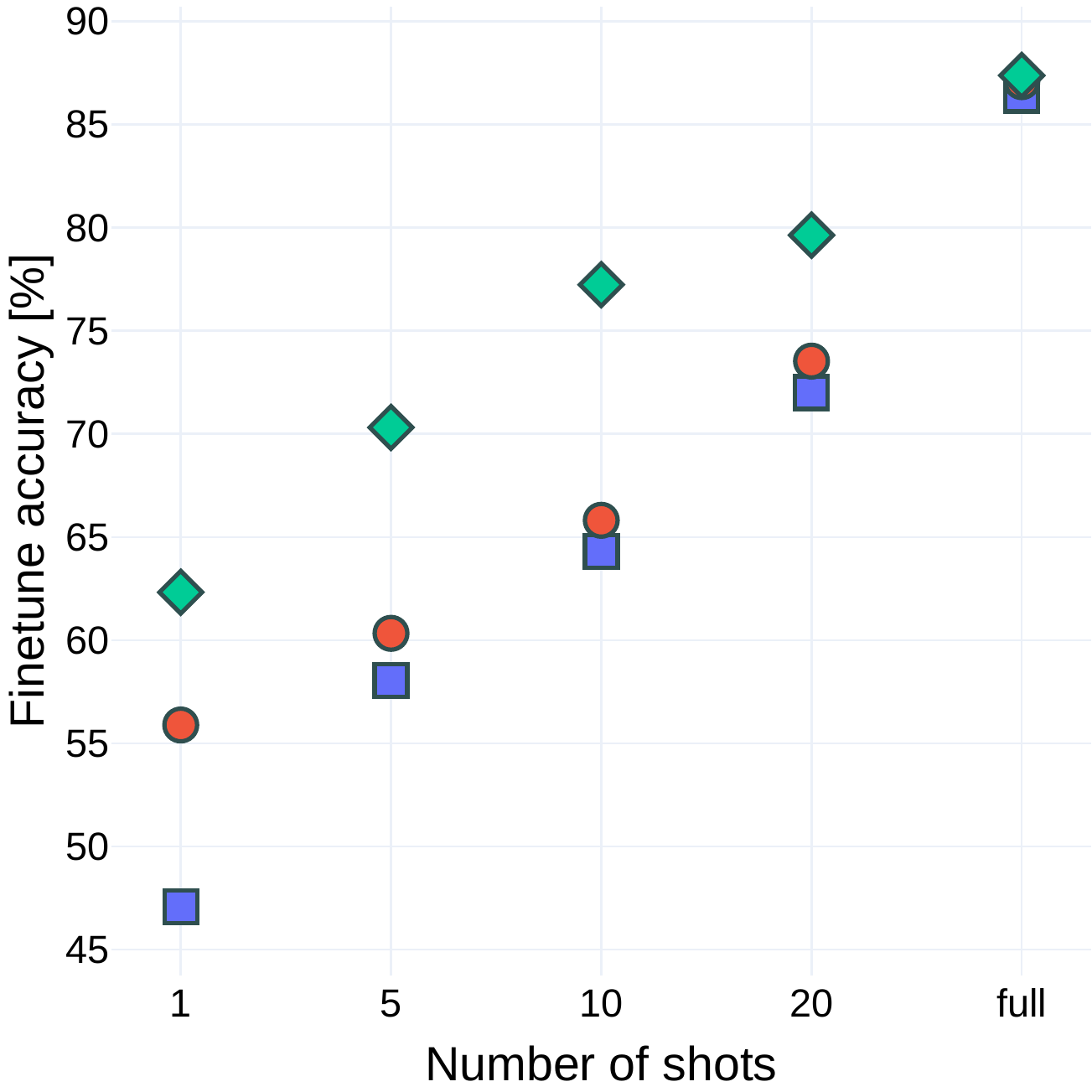}}
    \subfigure[REAL]{\includegraphics[width=.32\textwidth]{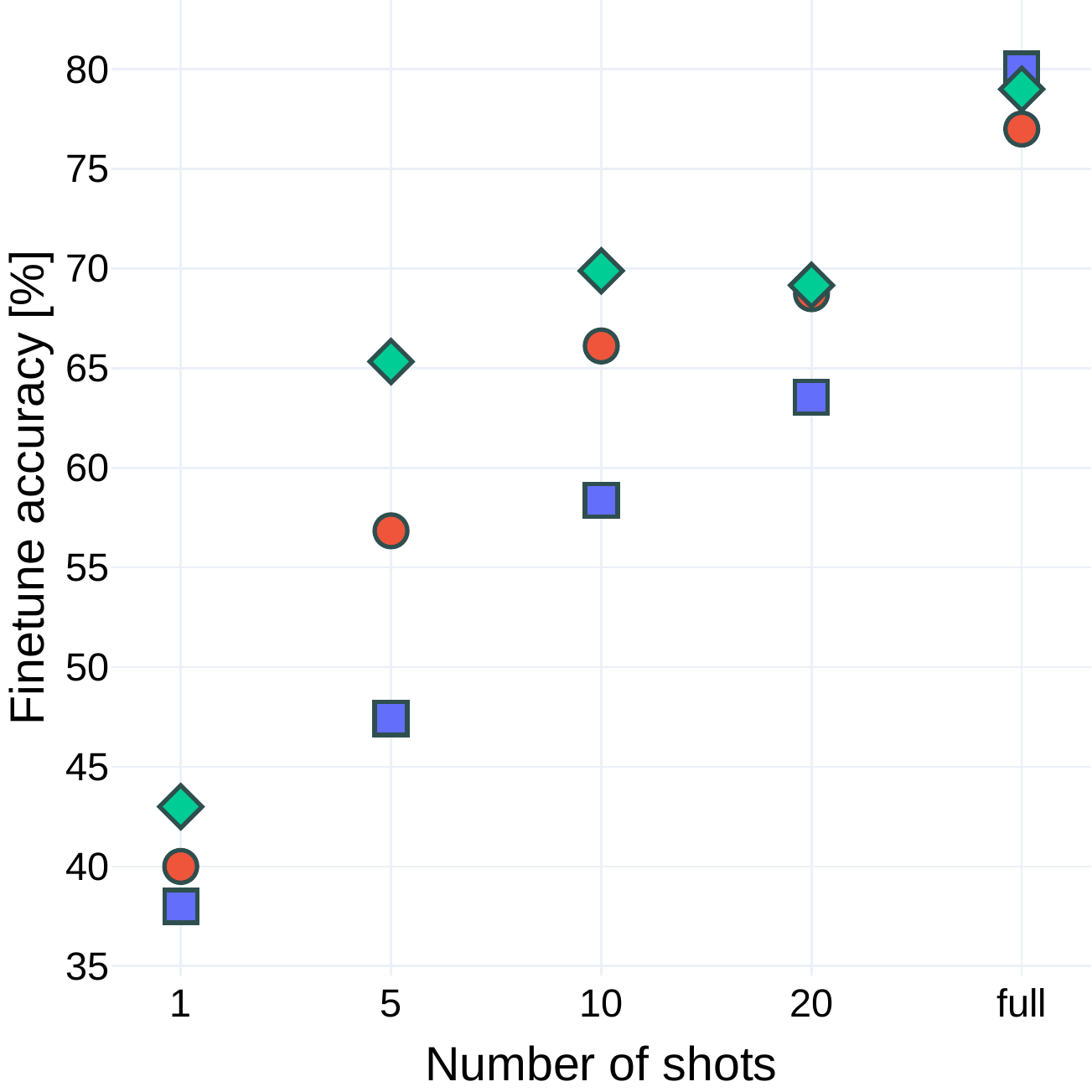}}
    \subfigure[CLIPART]{\includegraphics[width=.32\textwidth]{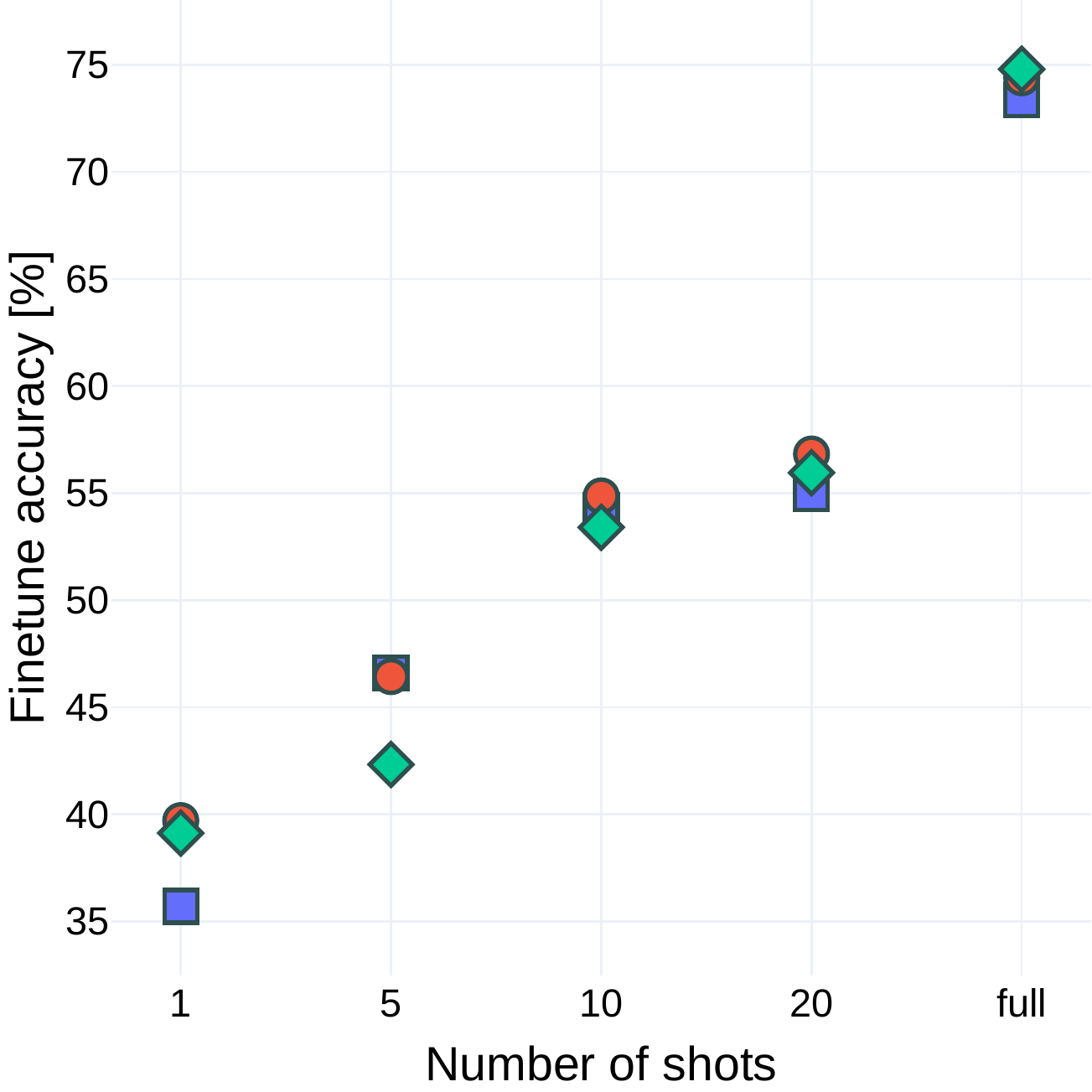}}

    \caption{\textbf{Effect of the pre-training data distribution when using SimCLR as the pre-training method.} Using different datasets for pre-training leads to a noticeable difference in downstream transfer accuracy. Similarly to the previous results for CLIP pre-training, the absolute difference in downstream transfer accuracy between different pre-training datasets is smaller when many images are available for fine-tuning.}
    \label{fig:shots_vs_acc_simclr}
\end{figure*}

\begin{figure*}[t]
    \centering
    \subfigure[CIFAR100]{\includegraphics[width=.32\textwidth]{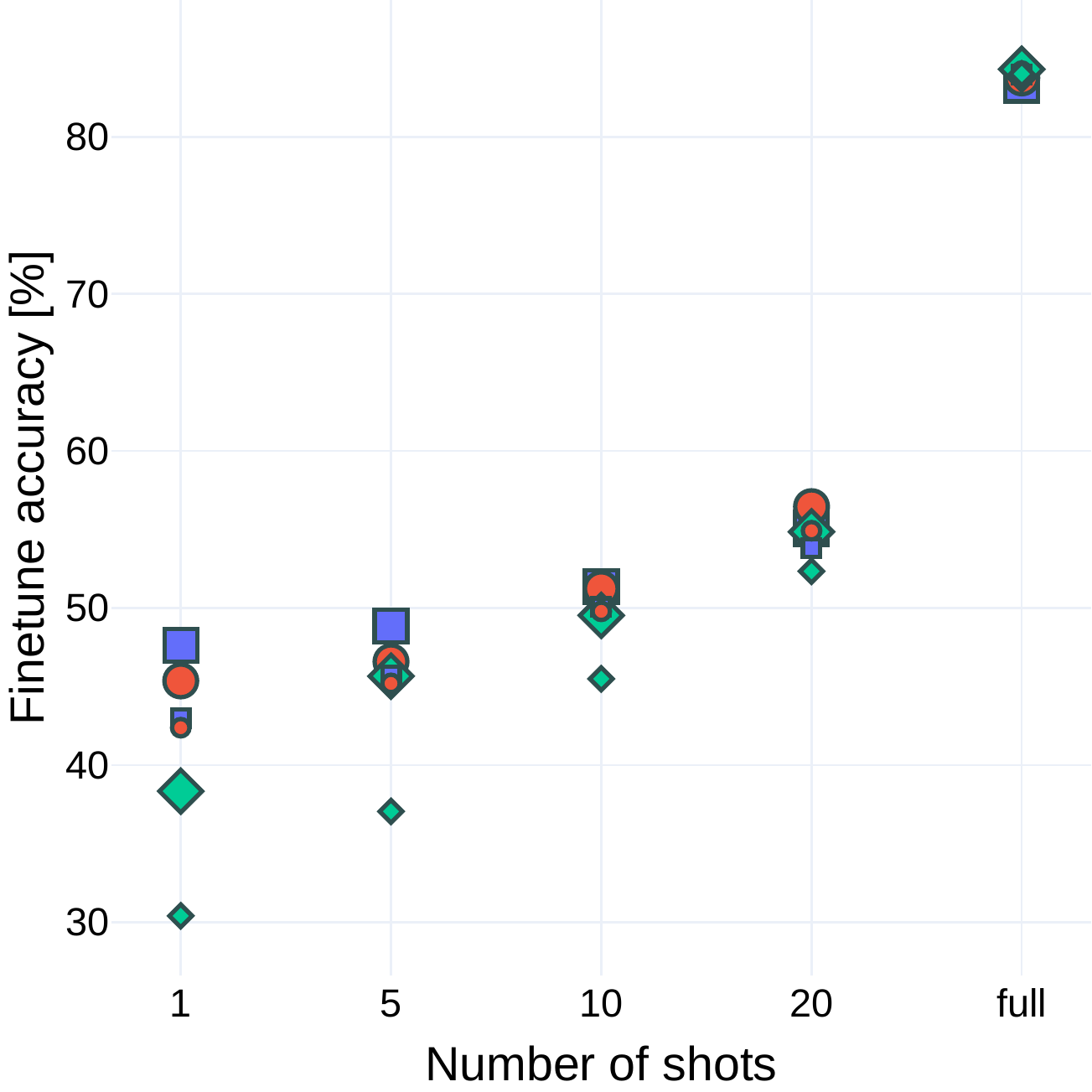}}
    \subfigure[DTD]{\includegraphics[width=.32\textwidth]{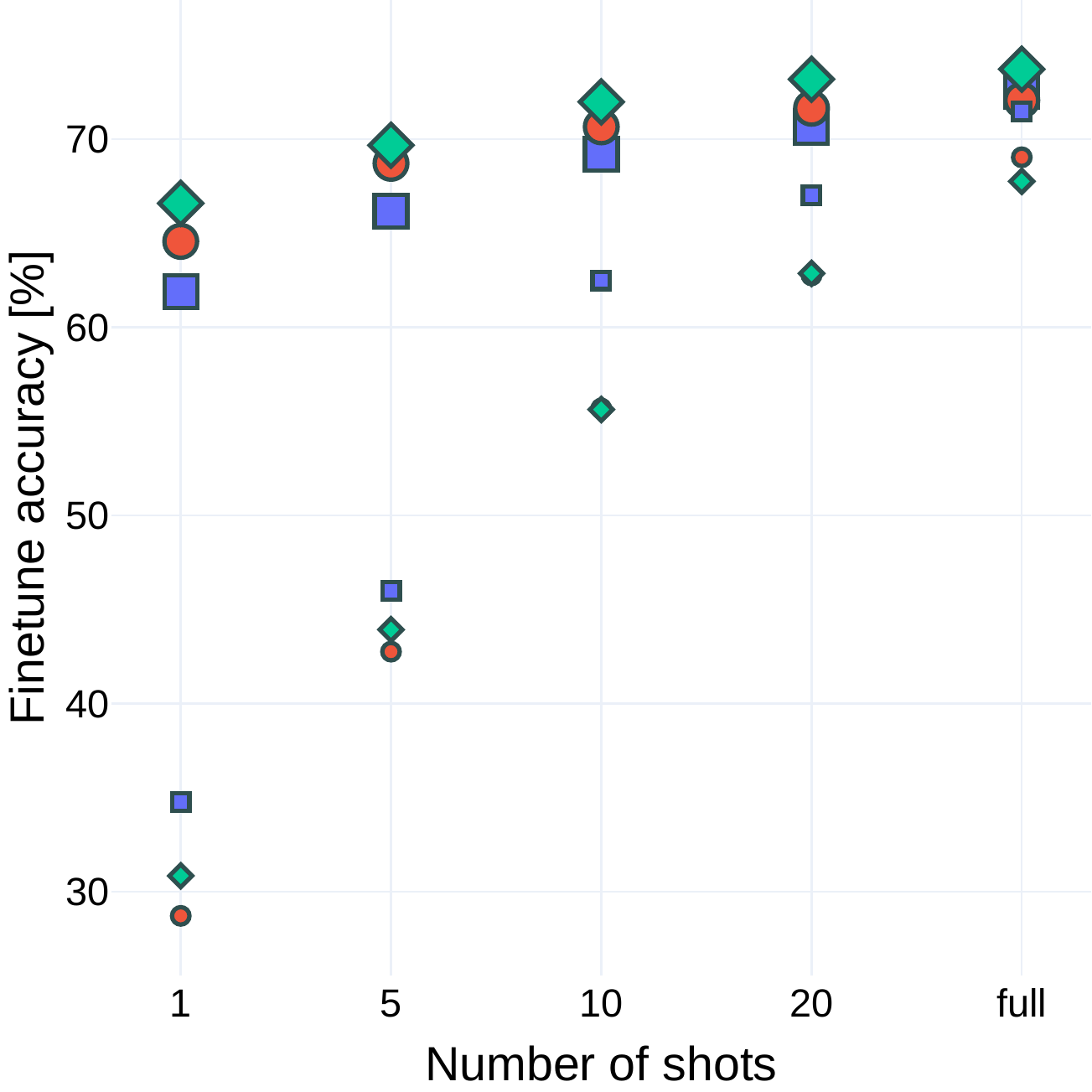}}
    \subfigure[CALTECH101]{\includegraphics[width=.32\textwidth]{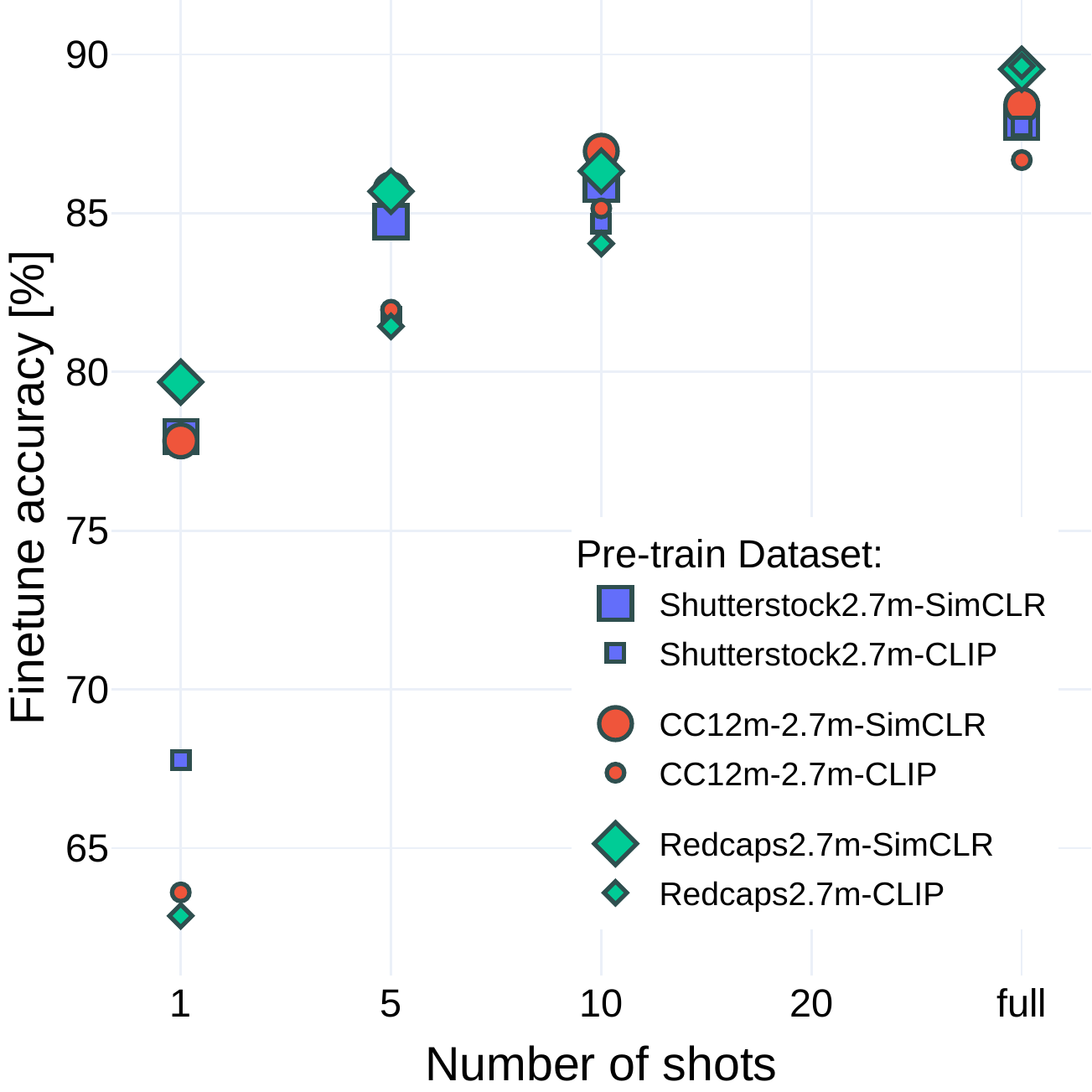}}
    \caption{\textbf{CLIP vs. SimCLR for pre-training.} 
    Overall we observe that SimCLR pre-training leads to a better downstream transfer accuracy than using CLIP pre-training on the same dataset. These differences are more pronounced in the few-shot setting.}
    \label{fig:shots_vs_acc_simclrvsclip}
\end{figure*}


\subsection{Which data distribution is better for transfer learning?}
The results presented in~\Figref{fig:shots_vs_acc_datadist_extend} demonstrate that pre-training on the Shutterstock and LAION datasets results in superior transfer performance across a range of downstream tasks. A closer look shows the superior performance of Redcaps for PETS. We investigate this further and inspect many pets by looking at random samples from Redcaps at~\Figref{fig:samples_redcaps}. We also look into the most common words in the captions of these pre-training datasets, summarized in Table \ref{tab:top20_words}. We observe that "cats" and "dogs" are among the most common words in the Redcaps dataset. Table \ref{tab:top20_words} also shows that "background", "design", "pattern", and "texture" are among the most common words in the captions of Shutterstock, supporting a high correlation to DTD (Describable Textures Dataset).
WiT yields the worst performance in most cases because both captions and images (\Figref{fig:samples_wit}) are related to topics about people and geography that are far from the studied downstream tasks.
\subsection{How much pre-training contributes to downstream performance as opposed to training from scratch?}
While transfer learning from a large pre-training dataset outperforms training from scratch for all downstream tasks, the magnitude of the improvement varies for different datasets in \Figref{fig:shots_vs_acc_datadist_extend}. We observe a large improvement for PETS, CALTECH-101, and CLIPART. PETS for example has a small number of samples per class for training (30), which makes it hard to train from scratch. It is also scraped from the web (Google search~\cite {fei2004learning}), similar to our web-scraped pre-training sources. We also hypothesize that a pre-training shows the best improvement when increases both diversity(how hard pre-train data is to fit) and affinity(how pre-training shifts the decision boundary of the scratch model)~\citep{gontijo2020tradeoffs}, meaning it should be semantically close to the classes of target task while enriching the distribution over the samples.


\subsection{Do well-curated pre-training datasets lead to better transfer?}\label{sec:ImageNet_sup}
There has been a significant effort to create computer vision datasets with high-quality labels. 
On the other hand, many recent datasets are large but noisy. 
In this section, we are going to investigate: \textit{How much is laborious ImageNet labeling worth?} 

To answer this question, we first start by pre-training ResNet-50 on Large Scale Visual Recognition Challenge (ILSVRC) 2012~\citep{russakovsky2015imagenet}, known as ImageNet-1K, using supervised cross-entropy loss and fine-tune on our downstream datasets in \Figref{fig:shots_vs_acc_datacuration}.
To investigate the role of supervision, we then discard ImageNet labels and use CLIP to pre-train on ImageNet. Because the ImageNet dataset has no captions, we include original Flickr captions, which reduces the size of the image and captions to 0.5M samples (Appendix~\ref{sec:app_curation} describes the required steps to create ImageNet-Flickr). \Figref{fig:shots_vs_acc_datacuration} shows that supervised pre-training on ImageNet outperforms CLIP pre-training on ImageNet with Flickr captions by a large margin in all downstream tasks.

However, such a gap could be attributed to two differences between mentioned pre-trainings: (1) supervised vs. contrastive image-language loss, and (2) the size of training samples for supervised-ImageNet (1.2m) is two times larger than CLIP with ImageNet-Flickr captions (0.5m). To remove the second effect we then use all the images from ImageNet, paired with templated clean captions, e.g., “a photo of a \textit{class name}”. This allows us to have a fair comparison between supervised and CLIP pre-training on ImageNet, given the same size. \Figref{fig:shots_vs_acc_datacuration} shows that pre-training with clean captions improves the performance of CLIP pre-training by a large margin and outperforms supervised pre-training on CIFAR100. However, supervised pre-training on ImageNet still performs best for the rest of the other datasets.

\subsection{How much LAION data is the ImageNet pre-training worth?}\label{sec:in_worth}
\Figref{fig:shots_vs_acc_datacuration} compares the ImageNet distribution with LAION. Pre-training CLIP on the ImageNet distribution (with template captions) outperforms LAION-1m by a large margin. Findings from~\Figref{fig:shots_vs_acc_violon} with the same pre-training loss are now extended to different losses in ~\Figref{fig:shots_vs_acc_datacuration},~\ie the gap between the supervised ImageNet (with template captions) pre-training and the contrastive LAION-1m pre-training shrinks as more data for the downstream task are available. Interestingly, pre-training CLIP on LAION-1m is only as good as ImageNet with Flickr captions with half of the data.
We also scale LAION pre-training size in~ \Figref{fig:INworth} to see if LAION can outperform ImageNet pre-training and downstream performance. \Figref{fig:INworth} shows that including 15$\times$ more data from LAION outperforms ImageNet pre-training with template captions only on CIFAR100. However, DTD, REAL, and CLIPART need 2000$\times$ more data from LAION to match or outperform ImageNet pre-training. Even including 2000$\times$ more data did not help CALTECH101. ImageNet pre-training also outperforms LAION-2B on PETS by a large margin. This is probably because PETS and ImageNet both share many samples of pets like dog breeds.  
\subsection{How does the downstream performance improve as more data is available for pre-training?}
\label{sec:datasetsize}
Can we expect that more pre-training data implies a better performance, or can the pre-training effectiveness saturate at some point? We fix the pre-training distribution to YFCC and LAION and compare pre-training on 2.7m samples with 15m samples. We also extended our experiments to see the effect of extreme sample sizes and include ViT-B/32 CLIP model trained on 2b samples from LAION. 
\Figref{fig:shots_vs_acc_datasetsize} shows that increasing the size of the dataset used for pre-training results in higher downstream transfer accuracy. However, the magnitude of the improvement varies across different downstream datasets. While increasing the pre-training size of YFCC and LAION improves the CIFAR100 performance by a large margin, this improvement is more modest for the rest of the downstream datasets. Specifically including 2 billion samples from LAION does not help much for CALTECH101 and PETS. 
Similarly to the findings in \Figref{fig:shots_vs_acc_violon}, in larger sizes, we observe more noticeable differences in downstream performance in the few-shot regime. The difference in the absolute accuracy when more data is available for fine-tuning is usually smaller. 
However, in contrast to the findings by \citet{abnar2021exploring}, pre-training on the extremely large LAION-2B still manages to boost the downstream performance in the full fine-tuning mode.
\Figref{fig:shots_vs_acc_datasetsize} shows that LAION pre-training outperforms YFCC pre-training on downstream tasks for both 2.7m and 15m subsets of the datasets. Interestingly, pre-training on LAION-2.7m performs similarly to a much larger size of YFCC-15m pre-training, highlighting the efficiency of LAION distributions.
Using an extremely large dataset of LAION-2B improves the performance by a significant margin in the few-shot regime for CIFAR100. While differences in absolute accuracy are smaller if more data is available for fine-tuning, LAION-2B pre-training still performs consistently better.

\subsection{Effect of pertaining loss} \label{sec:pretrainloss}
In this section, we replace the pre-training loss from language-image contrastive in CLIP with image-only contrastive loss in SimCLR~\citep{chen2020simple}. 
\Figref{fig:shots_vs_acc_simclr} highlights that our observations from~\Figref{fig:shots_vs_acc_violon} are now extended to image-only pre-training,~\ie changing the pre-train dataset leads to differences only in the few-shot downstream performance. 
Next, we evaluate the distinction between pre-training using CLIP and SimCLR. The results are presented in Figure~\ref{fig:shots_vs_acc_simclrvsclip} 
(Appendix~\Secref {sec:simclr_vs_clip} describes more details on training SimCLR for a fair comparison to CLIP).
Overall we find that models pre-trained with SimCLR have better downstream transfer accuracy than models pre-trained with CLIP in the few-shot regime.

Similarly to our observations regarding the effect of the pre-training data distribution (Figures \ref{fig:shots_vs_acc_violon}, \ref{fig:shots_vs_acc_datacuration}, and \ref{fig:shots_vs_acc_simclr}), the absolute accuracy difference is smaller when more data is used for fine-tuning. We note that this is different from what 
~\citep{santurkar2022caption} observed. However, we suspect this difference is because we are fine-tuning all model parameters while they only consider a linear classifier.

The difference in the downstream transfer accuracy for CLIP and SimCLR pre-training varies across different datasets. While SimCLR is only marginally better than CLIP for CIFAR100, the difference is significantly larger for DTD and CALTECH101, especially in the few-shot setting.

\section{Discussion, Limitations, and Future Work}
\label{sec:discussion}

\fakeparagraph{Discussion}
As better pre-trained models become available, and more workloads shift from training from scratch to fine-tuning, understanding the transfer learning paradigm becomes increasingly important. Presumably, in the future, a sea of pre-trained models will be available for download from the Internet.
Therefore, researchers and practitioners will be faced with the question of where to begin.
It will be important to make this choice well, but also to understand to what extent this choice matters.
Overall we have observed that different pre-training distributions and methods can lead to differences in downstream transfer accuracy.
However, these differences are  the largest in the few-shot transfer regime. If many images are used for fine-tuning these differences are mostly diminished.
Moreover, while different pre-training decisions lead to similar accuracy in the high-shot regime, they still outperform training from scratch in the setting we consider.
We also observed that the pre-training method affects performance on downstream transfer (supervised vs. CLIP vs. SimCLR) and including more pre-training data may compensate for the performance gap between training methods.

\fakeparagraph{Limitations and Future work}
There are a number of limitations in our study.
For one, we consider only end-to-end fine-tuning, because this method produces the highest accuracy.
However, if compute is limited, one may choose to instead use linear probing or other lightweight fine-tuning methods.
So far this is not addressed in our study.
Another limitation is that we did not do an exhaustive hyperparameter sweep for pre-training.
While fine-tuning is cheaper and we are therefore able to do a grid search, for pre-training we are mostly limited to using existing checkpoints.
While we think that this reflects a realistic setting, in the future we wish to also better understand the role of hyperparameters. 

In addition to the mentioned limitations, future works might include extending experiments to include different samples of ImageNet.
One example may include subsets of ImageNet-21K (2.7m in \Figref{fig:shots_vs_acc_datadist_extend} and 15m in \Figref{fig:INworth}) and respective comparison to Shutterstock and LAION distributions. Given our observation of the role of data curation, we also hope that our findings stimulate further direction toward creative methods for dataset curation.

\section{Acknowledgements}
This work is in part supported by the Austrian Marshall Plan Foundation, Google Cloud Research Credit, Rudolf Chaudoire Programm from the Faculty of Electrical and Information Engineering at Graz University of Technology, Graz Center for Machine Learning at Graz University of Technology (GraML), and NSF IIS 1652052, IIS 17303166, DARPA N66001-19-2-4031, DARPA W911NF-15-1-0543 and gifts from Allen Institute for Artificial Intelligence.
\bibliography{references}
\bibliographystyle{arxiv.bst}
\cleardoublepage
\appendix
\onecolumn
\section*{Appendix}
\begin{figure}[H]
    \centering
    \subfigure[CIFAR100]{\includegraphics[width=.32\textwidth]{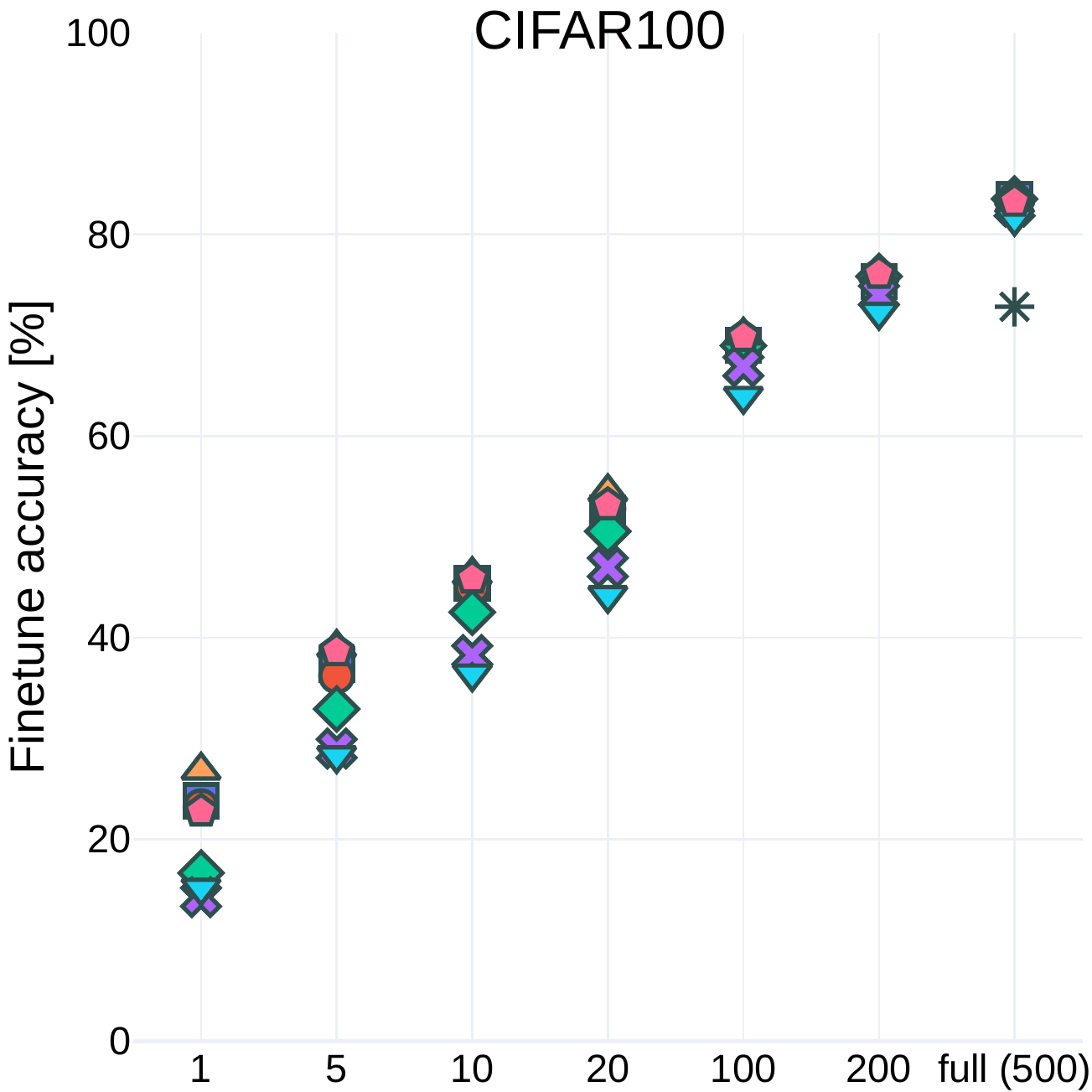}}
    \subfigure[DTD]{\includegraphics[width=.32\textwidth]{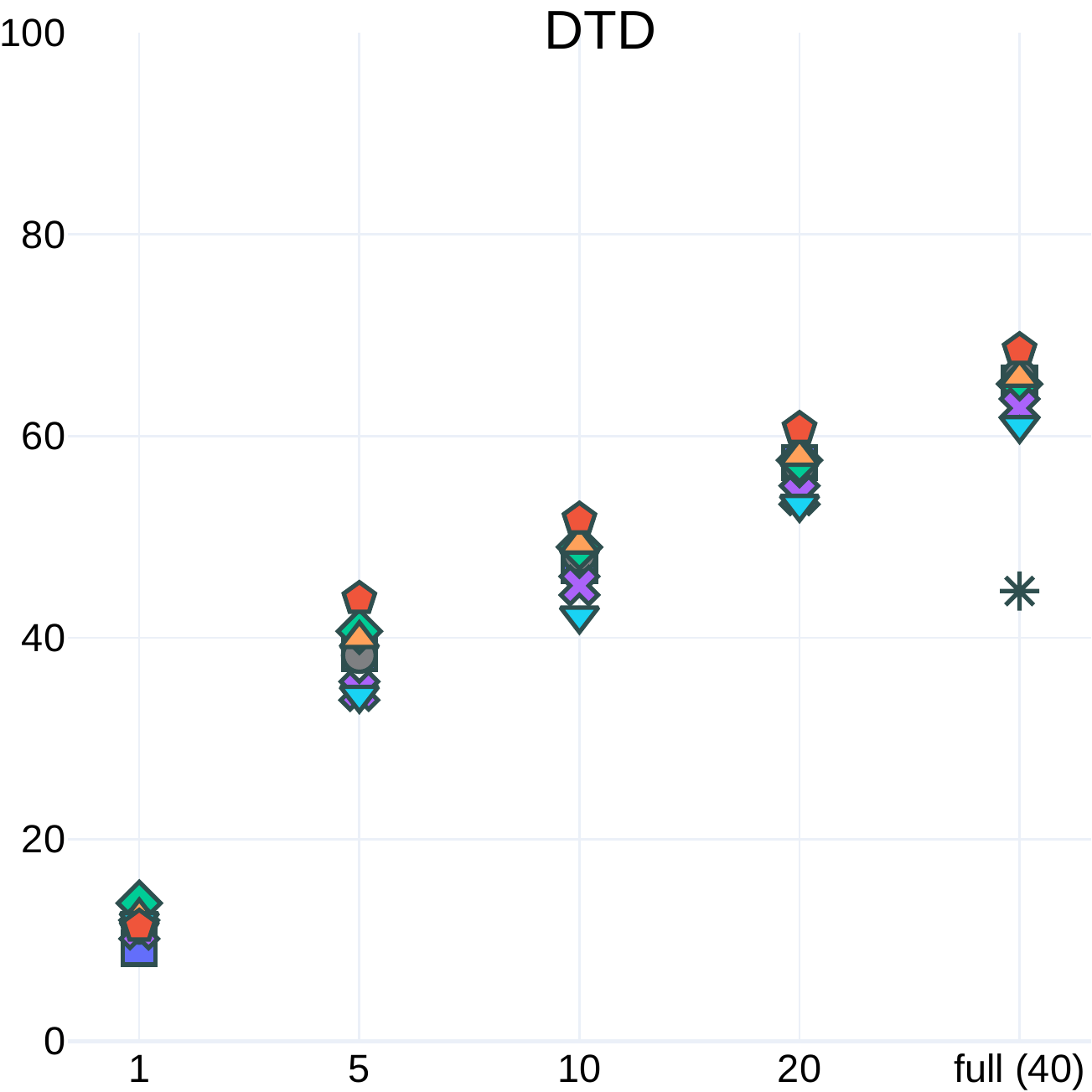}}
    \subfigure[CALTECH101]{\includegraphics[width=.32\textwidth]{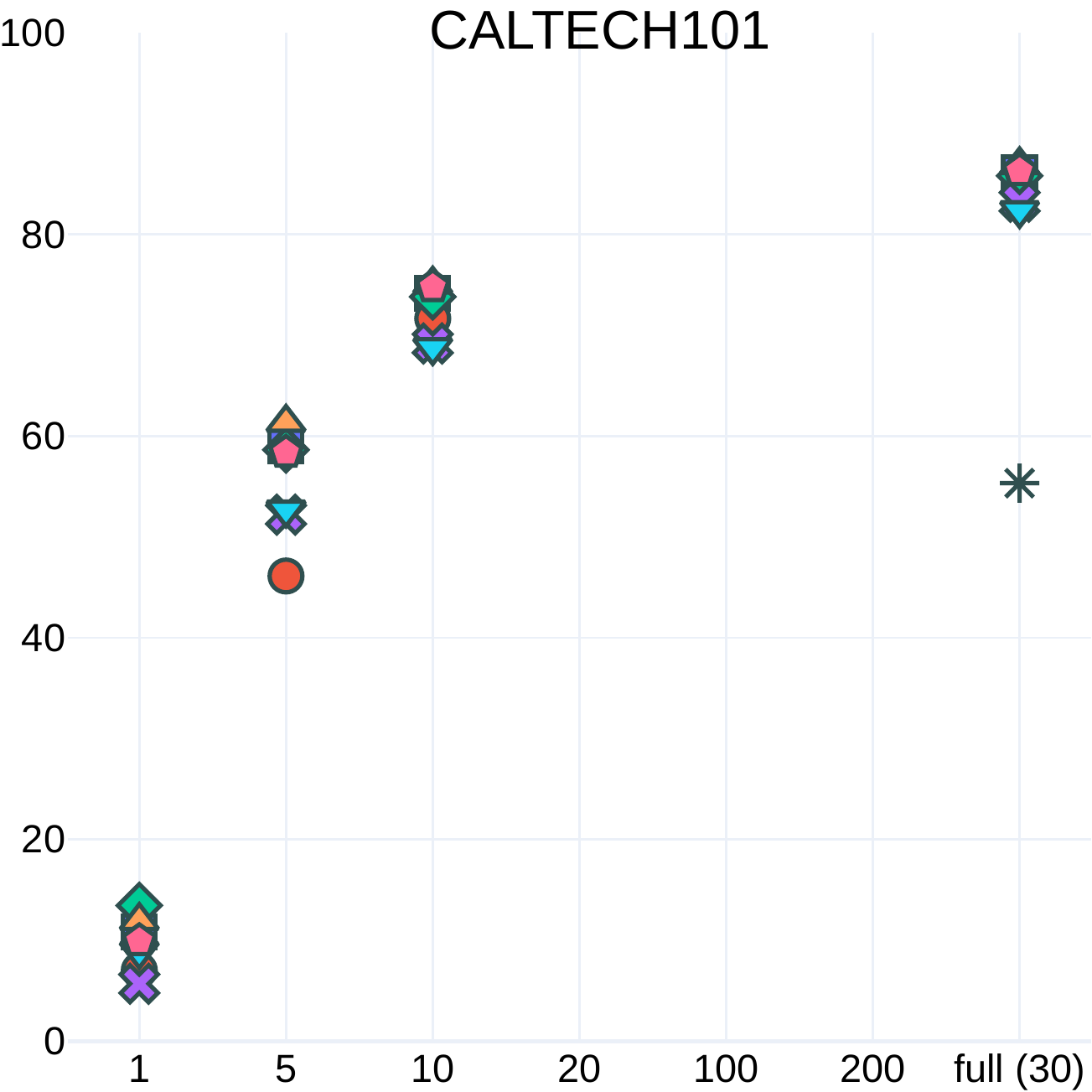}}
    
    \subfigure[PETS]{\includegraphics[width=.32\textwidth]{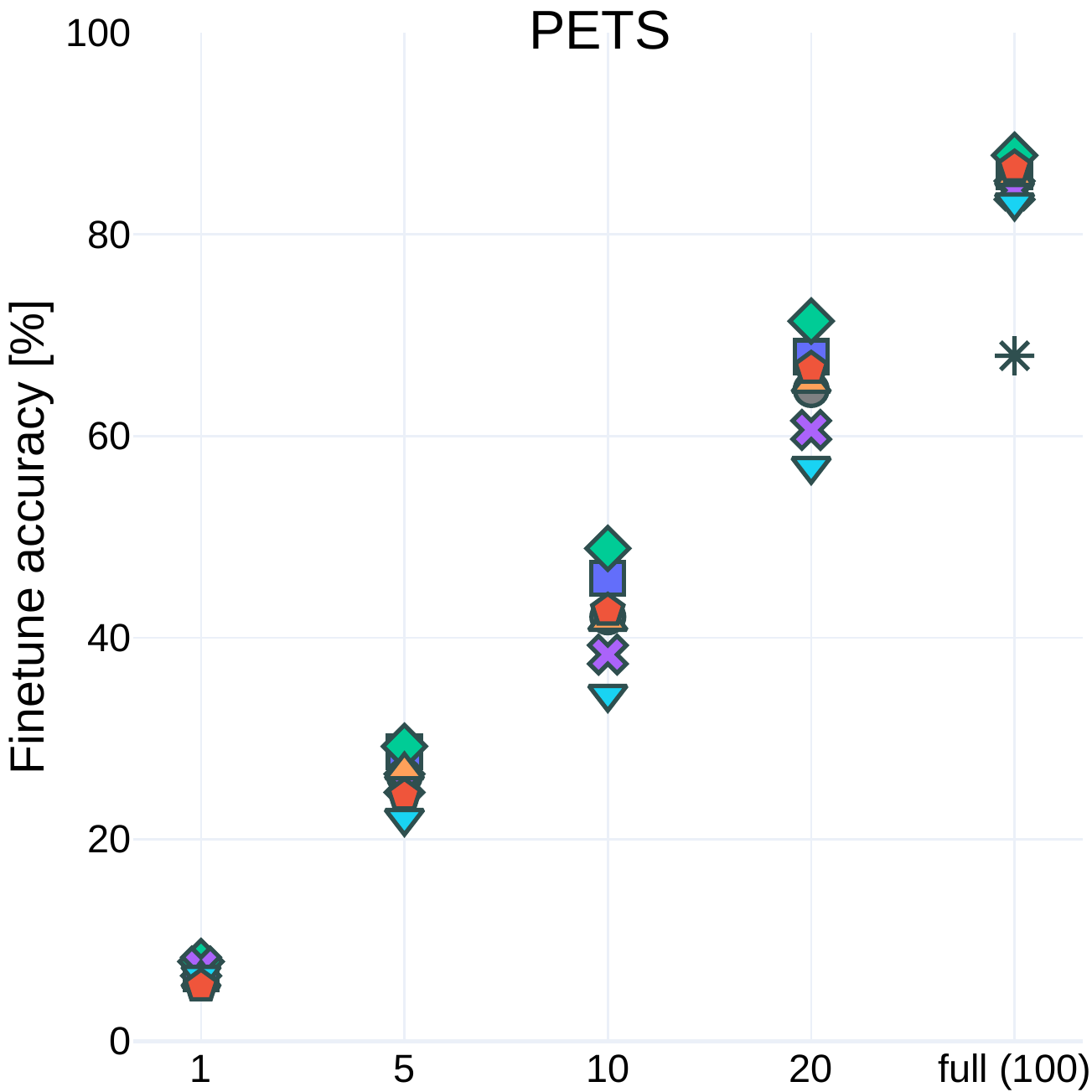}}
    \subfigure[DomainNet REAL]{\includegraphics[width=.32\textwidth]{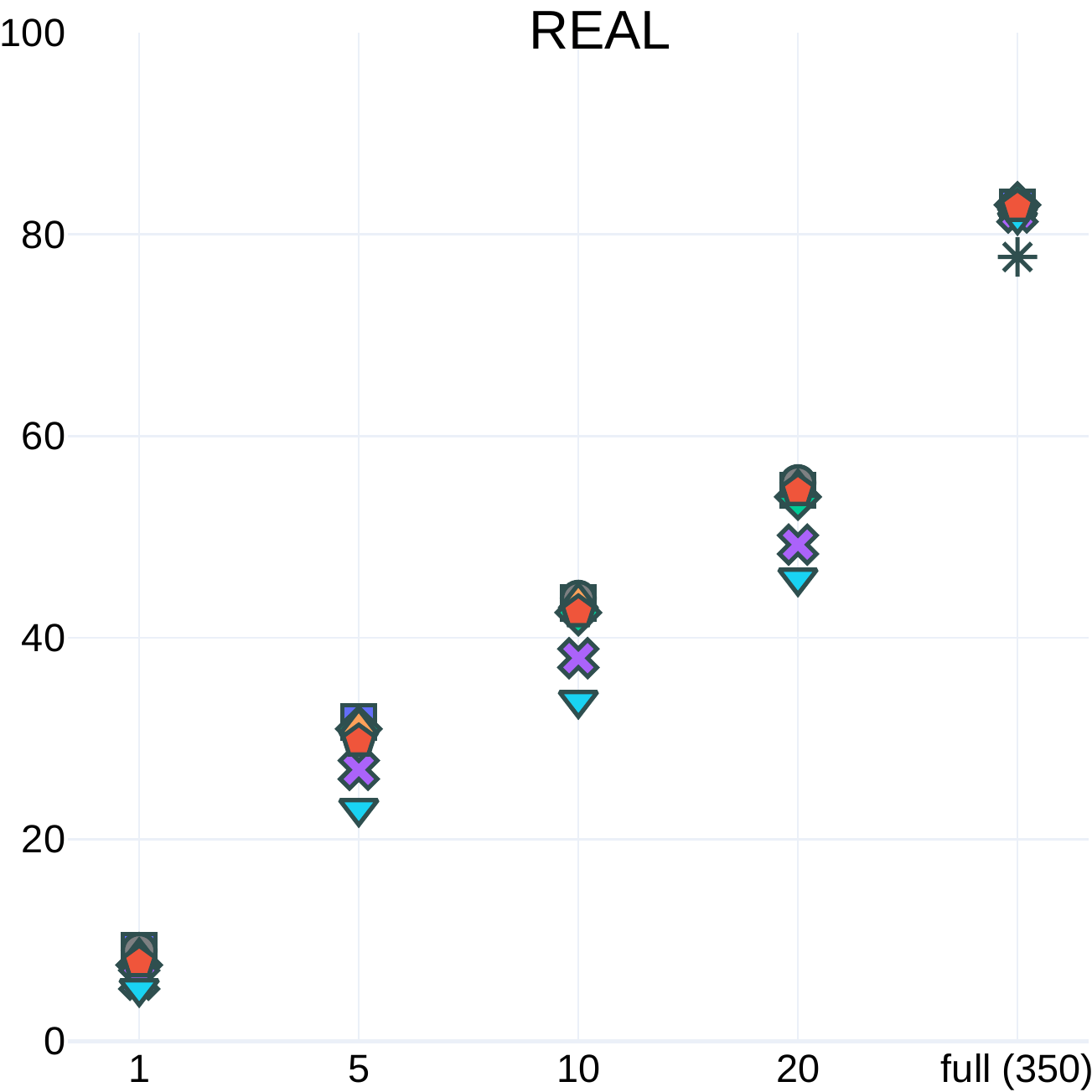}}
    \subfigure[DomainNet CLIPART]{\includegraphics[width=.32\textwidth]{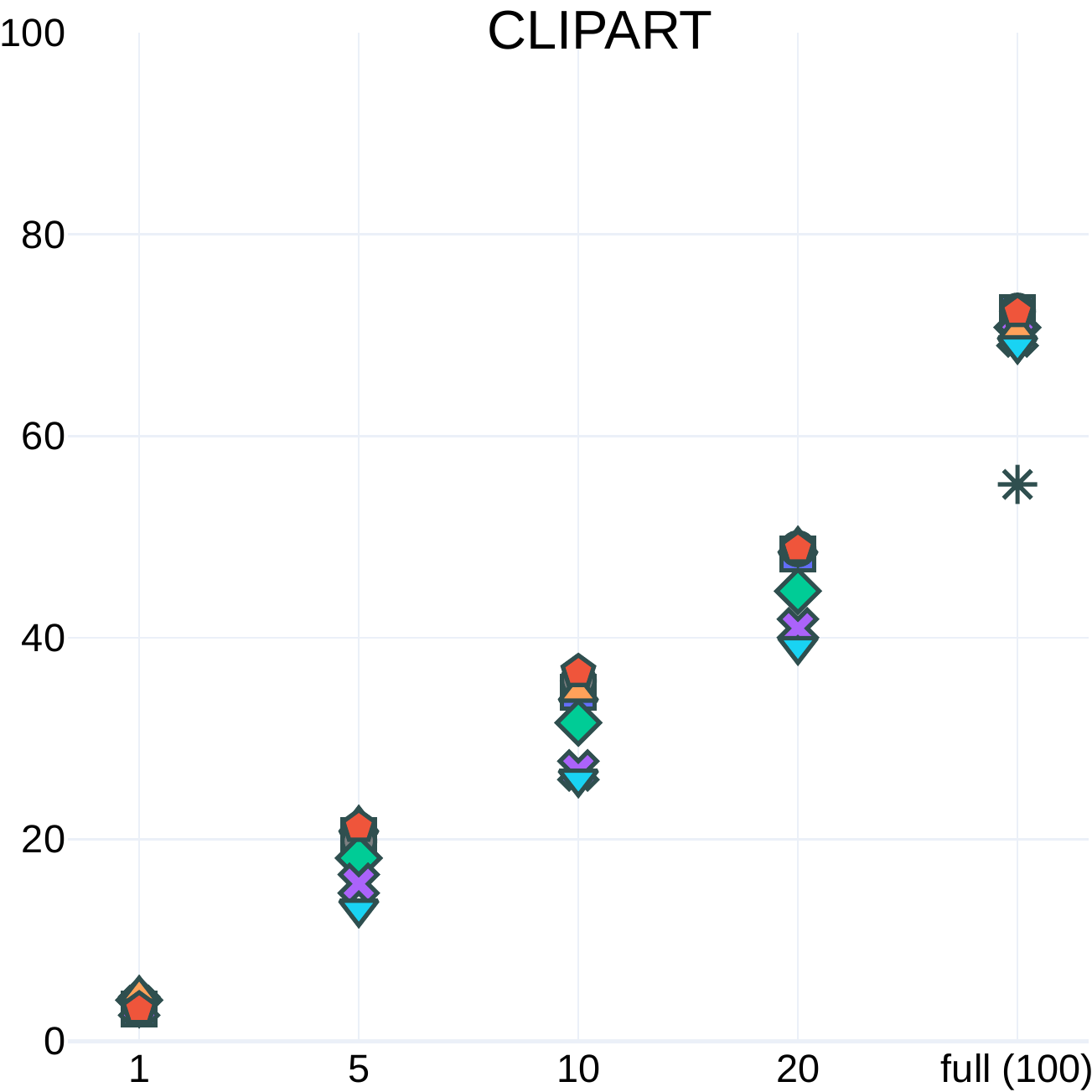}}
    \subfigure[Camera Traps]{\includegraphics[width=.32\textwidth]{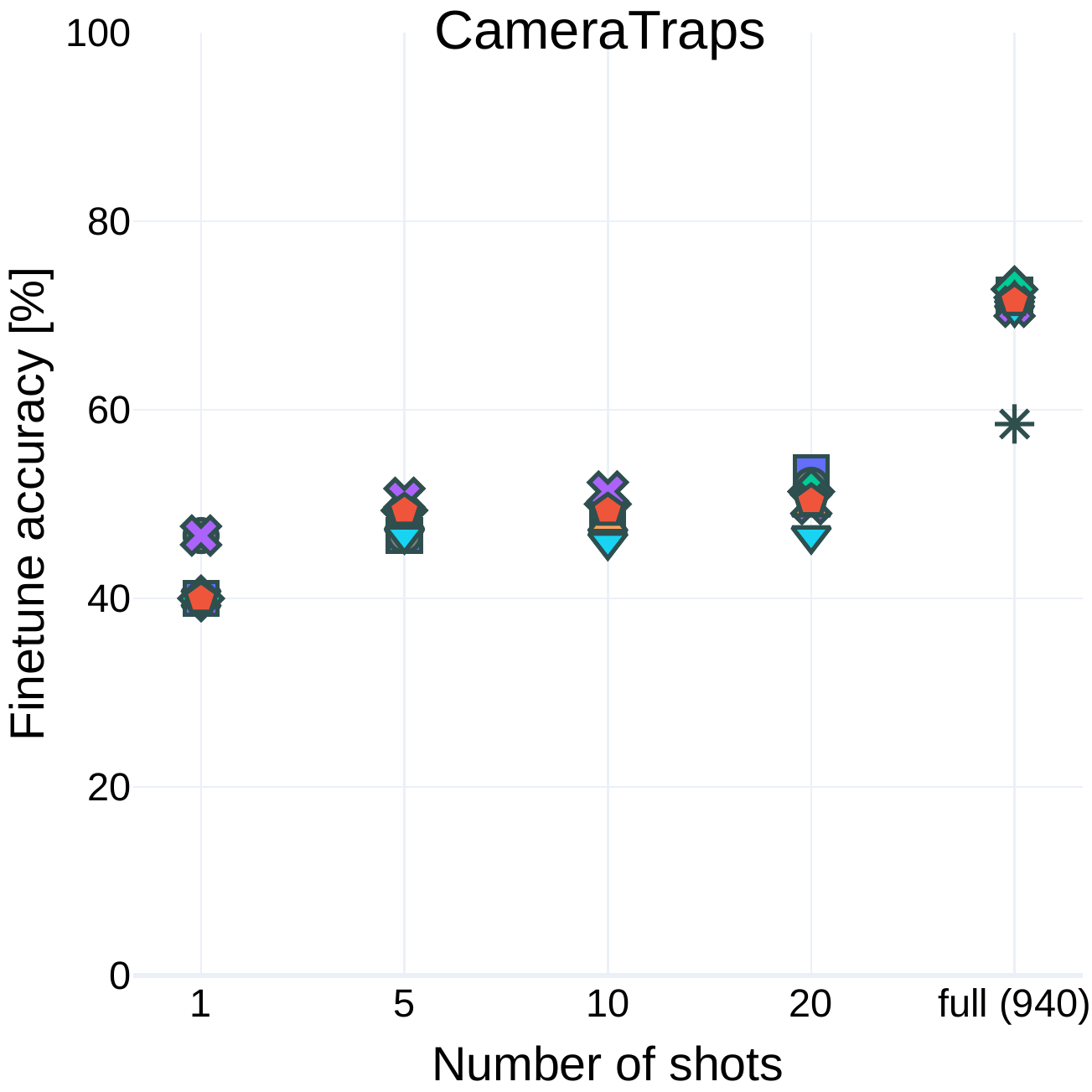}}
    \subfigure[Cassava Leaf]{\includegraphics[width=.32\textwidth]{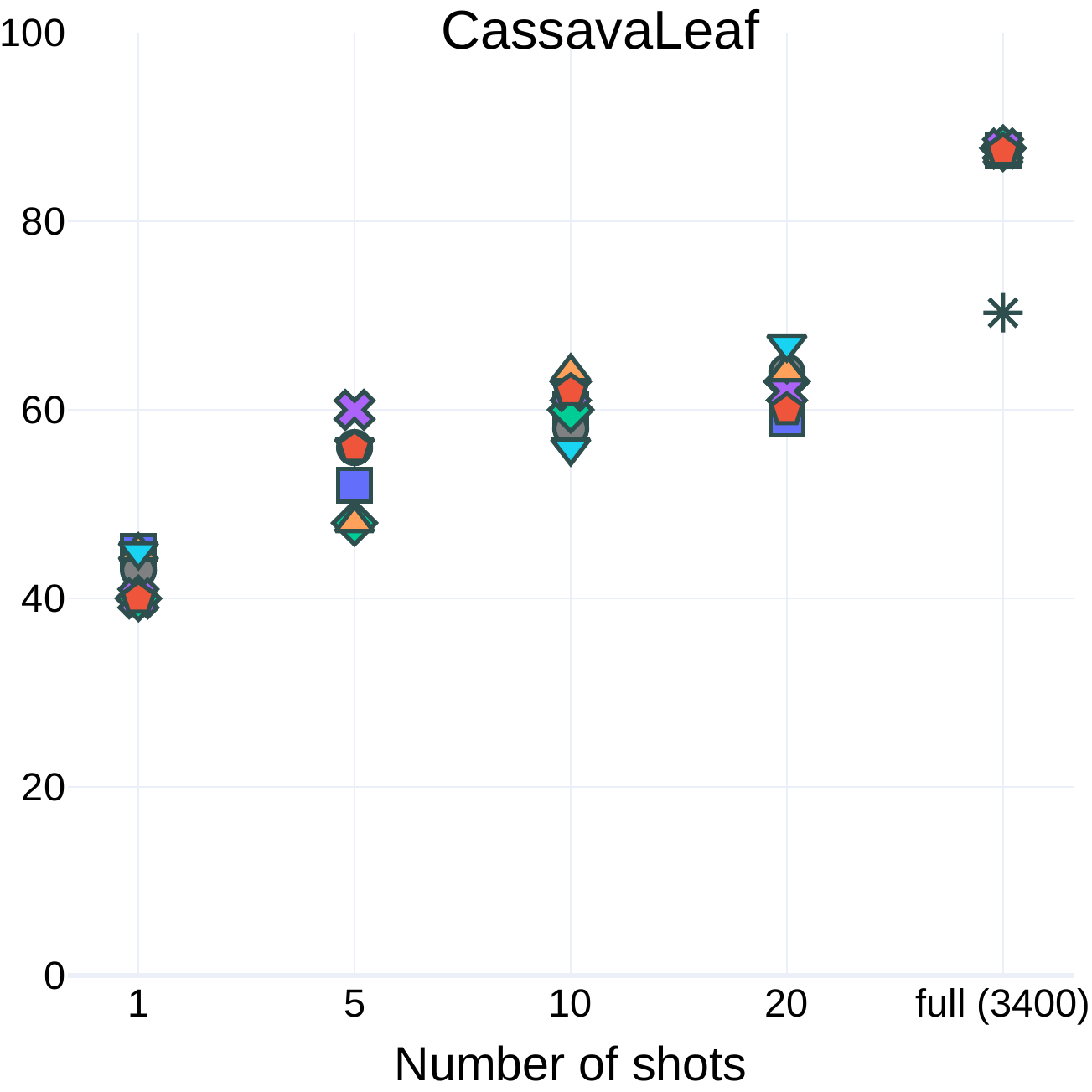}}
    \subfigure[EuroSAT]{\includegraphics[width=.32\textwidth]{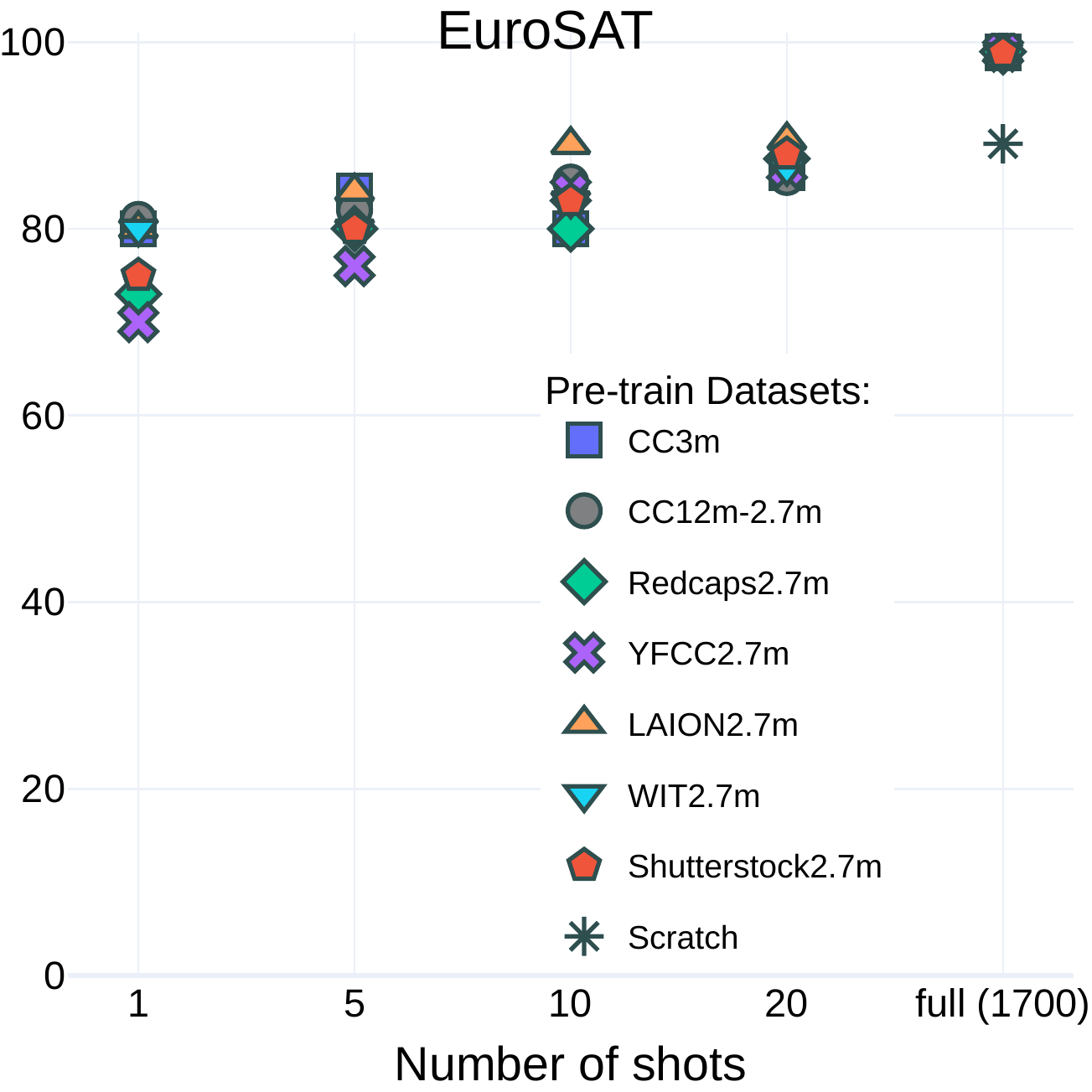}}
    
    \caption{\textbf{Effect of the pre-training data distribution.} While~\Figref{fig:shots_vs_acc_violon} shows the aggregated results all downstream datasets,  here we include the performance for each pair of (pretraining, downstream) datasets in detail. In the low-shot setting, different pre-training datasets lead to noticeable differences in downstream performance. If many samples are available for fine-tuning, the difference in absolute accuracy between models pre-trained on different sources largely evaporates.}
    \label{fig:shots_vs_acc_datadist_extend}
\end{figure}
\section{Effect of the pre-training data distribution}\label{sec:ds_extend}

\Figref{fig:shots_vs_acc_datadist_extend} shows a detailed for aggregated results shown in~\Figref{fig:shots_vs_acc_violon}. In the low-shot setting, different pre-training datasets lead to noticeable differences in downstream performance. If many samples are available for fine-tuning, the difference in absolute accuracy between models pre-trained on different sources largely evaporates.

\Figref{fig:shots_vs_acc_datadist_extend} compares different data sources for pre-training.
While Shutterstock shows superior performance on the first six datasets (except for PETS), the best pre-training distribution changes between Camera Traps, Cassava Leaf, and EuroSAT. Changing the pre-training dataset leads to noticeable differences in the downstream low-shot performance of nine datasets. 


\section{Training Details}
\label{sec:trainingdetails}
\subsection{CLIP training}
Our CLIP models are trained from scratch on each of the pre-training datasets unless otherwise mentioned and follow the training code from the OpenCLIP GitHub repository\citep{ilharco_gabriel_2021_5143773}. CLIP models are trained using AdamW optimizer~\citep{loshchilov2017decoupled} with default PyTorch parameters $\beta_1= 0.9$, $\beta_2 = 0.999$, $\epsilon = 10^{-8}$, batch size 1024, and weight decay of 0.1. For learning rate, we start with a learning rate of $10^{-3}$ and apply a cosine-annealing learning rate schedule~\citep{loshchilov2016sgdr} with 5,000 steps warm-up. We use the same data augmentations as in\citep{radford2021learning}. 
\subsection{SimCLR training}
Our SimCLR implementation closely follows the training code from the SLIP\citep{mu2021slip}. 
SimCLR models are also trained for 16 epochs from scratch using AdamW optimizer~\citep{loshchilov2017decoupled} with $\beta_1= 0.9$, $\beta_2 = 0.98$, $\epsilon = 10^{-8}$, batch size 1024, and weight decay of 0.1. we start with a learning rate of $10^{-3}$ and apply a cosine-annealing learning rate schedule~\citep{loshchilov2016sgdr} with 2 epochs of warm-up. The hidden dimension of SimCLR MLP projection head is set to 4,094 and the output embedding dimension of MLP projection head is set to 256.

\subsection{Finetuning details}\label{sec:finetunedetails}
Each pretrained model is finetuned on the specific downstream task for 128 epochs while the learning rate is mostly from {0.0001, 0.0003, 0.001, 0.003} as starting and applying a cosine-annealing learning rate schedule~\citep{loshchilov2016sgdr} with 500 steps warm-up and batch size of 128. For each fine-tuning, we choose the best-performing result on the test set among the performed grid search. We use the implementation from the WiSE-FT GitHub repository for fine-tuning, where we have only one model and $\alpha=1$~\citep{wortsman2021robust}. For a list of all \textbf{4000 experiments}, including their hyperparameters and performance see \url{https://github.com/rahimentezari/DataDistributionTransferLearning/blob/main/Hyperparameters_results.csv}

\begin{figure}[t]
    \centering
    {\includegraphics[width=.99\textwidth]{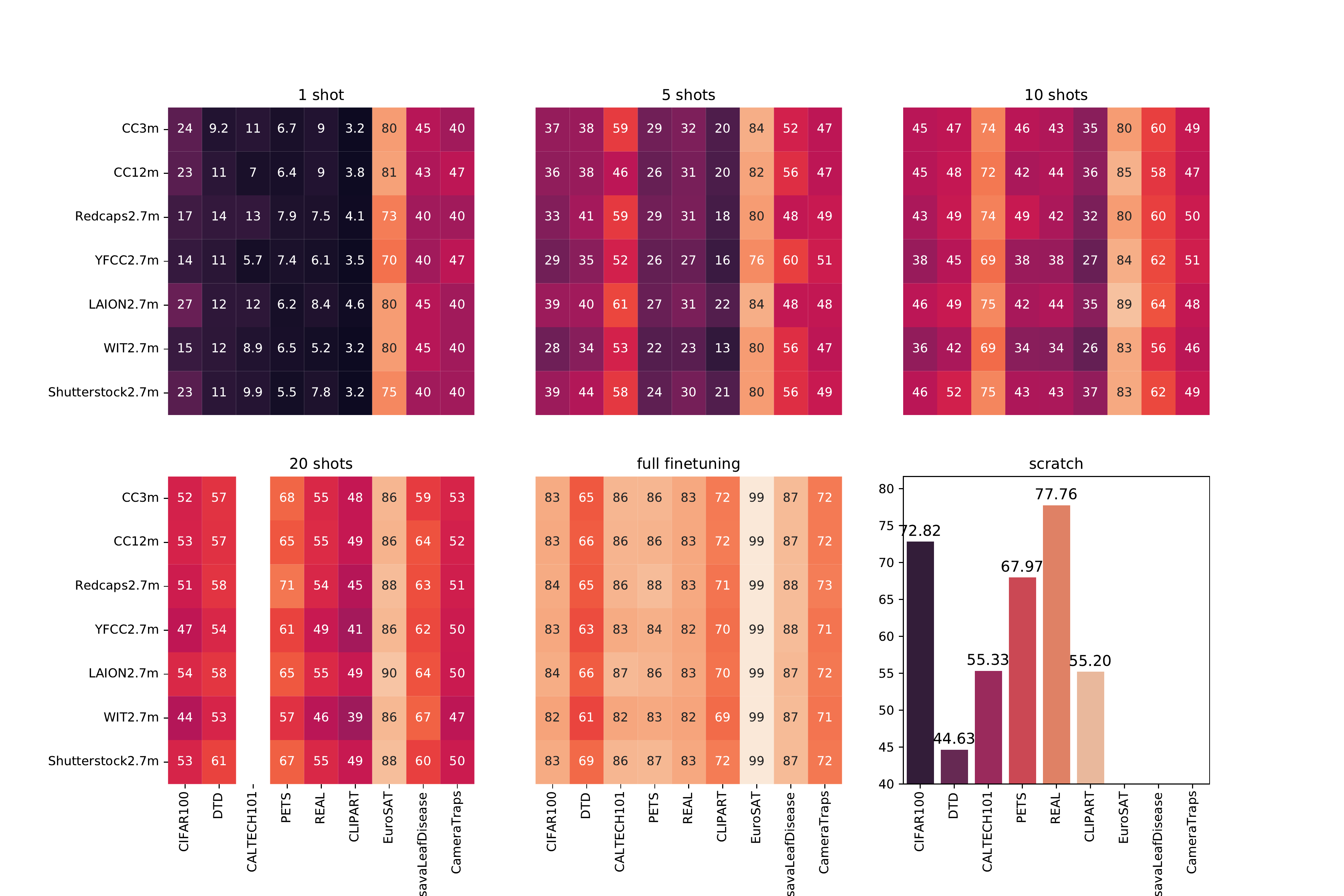}}
    \caption{\textbf{Effect of pre-training data distribution: a better view.} We change the presentation of~\Figref{fig:shots_vs_acc_datadist_extend} for a better view of exact performance numbers on different data distributions and datasets. }
    \label{fig:samples_betterview}
\end{figure}

\section{Effect of data curation: ImageNet captioning}\label{sec:app_curation}
We compare CLIP models pre-trained on LAION with CLIP models pre-trained on the following two versions of the curated ImageNet dataset:
\begin{itemize}
    \item IN1K-Flickr-Captions: This is a subset of the ImageNet Large Scale Visual Recognition Challenge (ILSVRC) 2012 training set, paired with the original image title, description, and tags from Flickr.
    Therefore, we can use it for CLIP pre-training. To construct this dataset, \citet{fang2022data} start from 14,197,122 image URLs in the ImageNet fall 2011 release, and filter to only include images from Flickr. Next, they restrict the images to the 1,000 classes included in the 2012 ImageNet competition, run the image deduplication routine, and remove text containing profanity. As a result, the dataset of 463,622 images is left along with the newly obtained corresponding text data.
    \item IN1K-Template-Captions: This dataset includes all data in the ImageNet dataset, paired with templated captions, e.g., ``a photo of a {classname}''. This allows us to use CLIP pre-training but on clean images and text. In terms of ImageNet accuracy, this training scheme is very similar to standard supervised training.
    However, this is now a controlled experiment as we are always using CLIP pre-training.
\end{itemize}

\section{Other architectures}\label{sec:extend_archs}
\begin{figure}[ht]
    \centering

    \subfigure[CIFAR100]{\includegraphics[width=.32\textwidth]{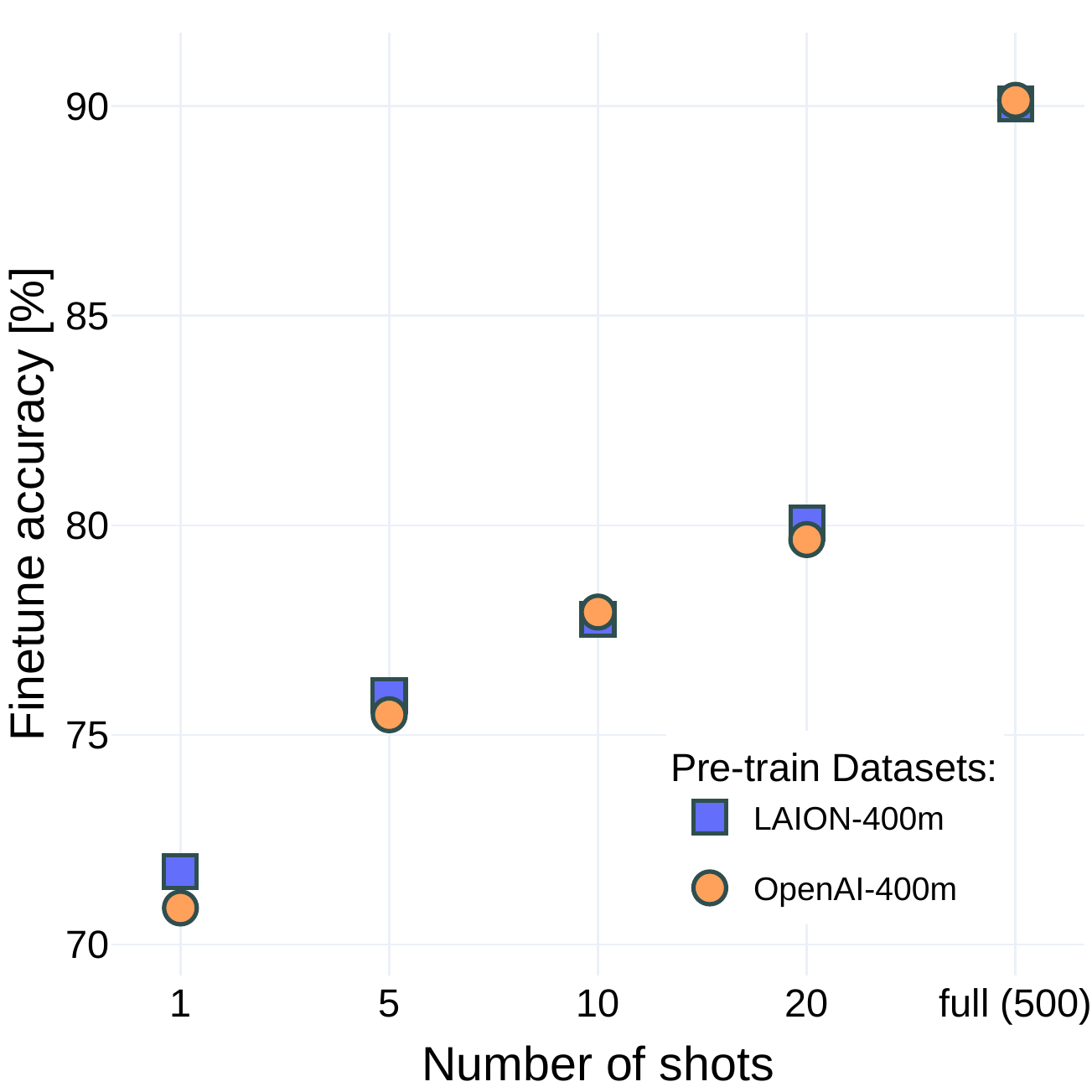}}
    \subfigure[DTD]{\includegraphics[width=.32\textwidth]{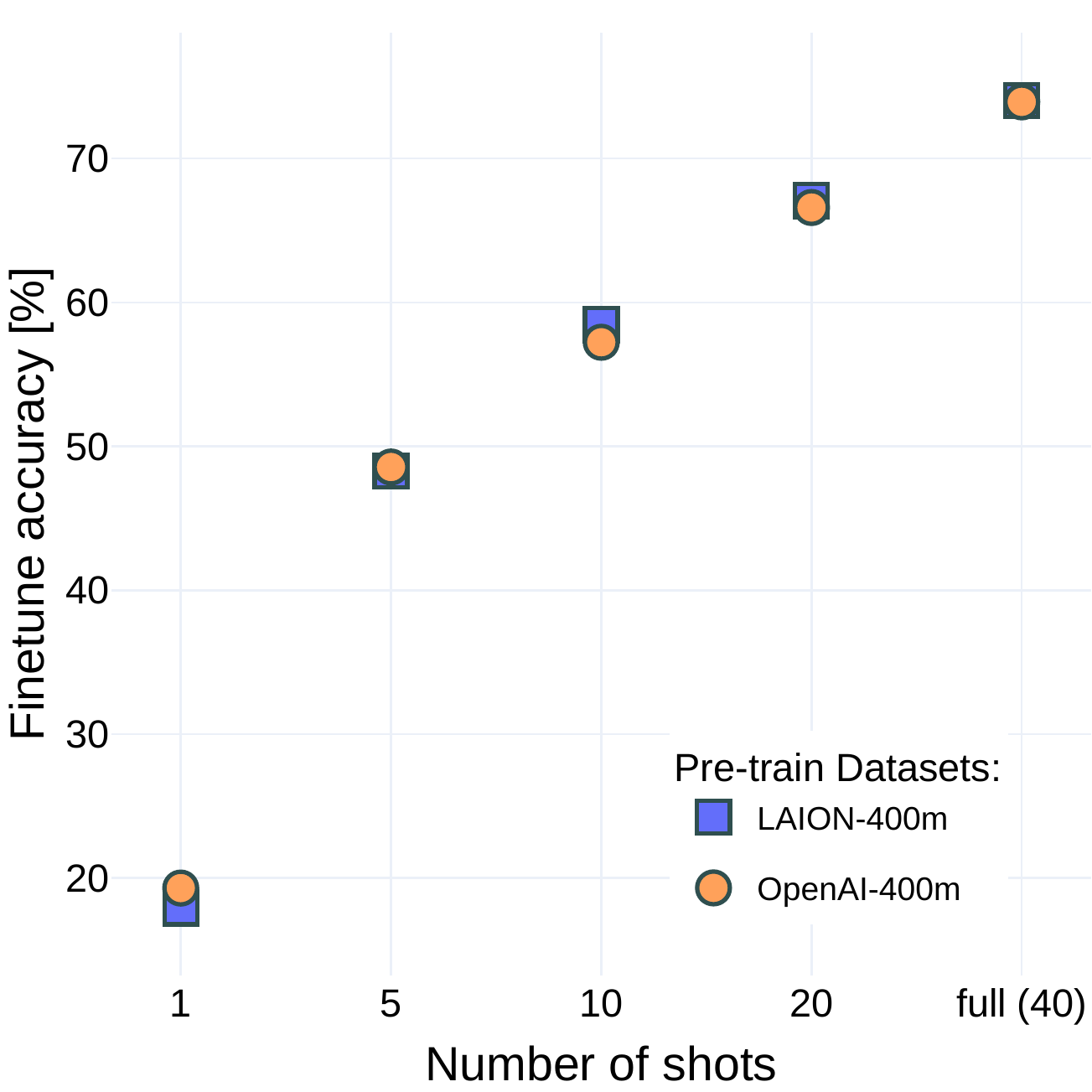}}
    \subfigure[CALTECH101]{\includegraphics[width=.32\textwidth]{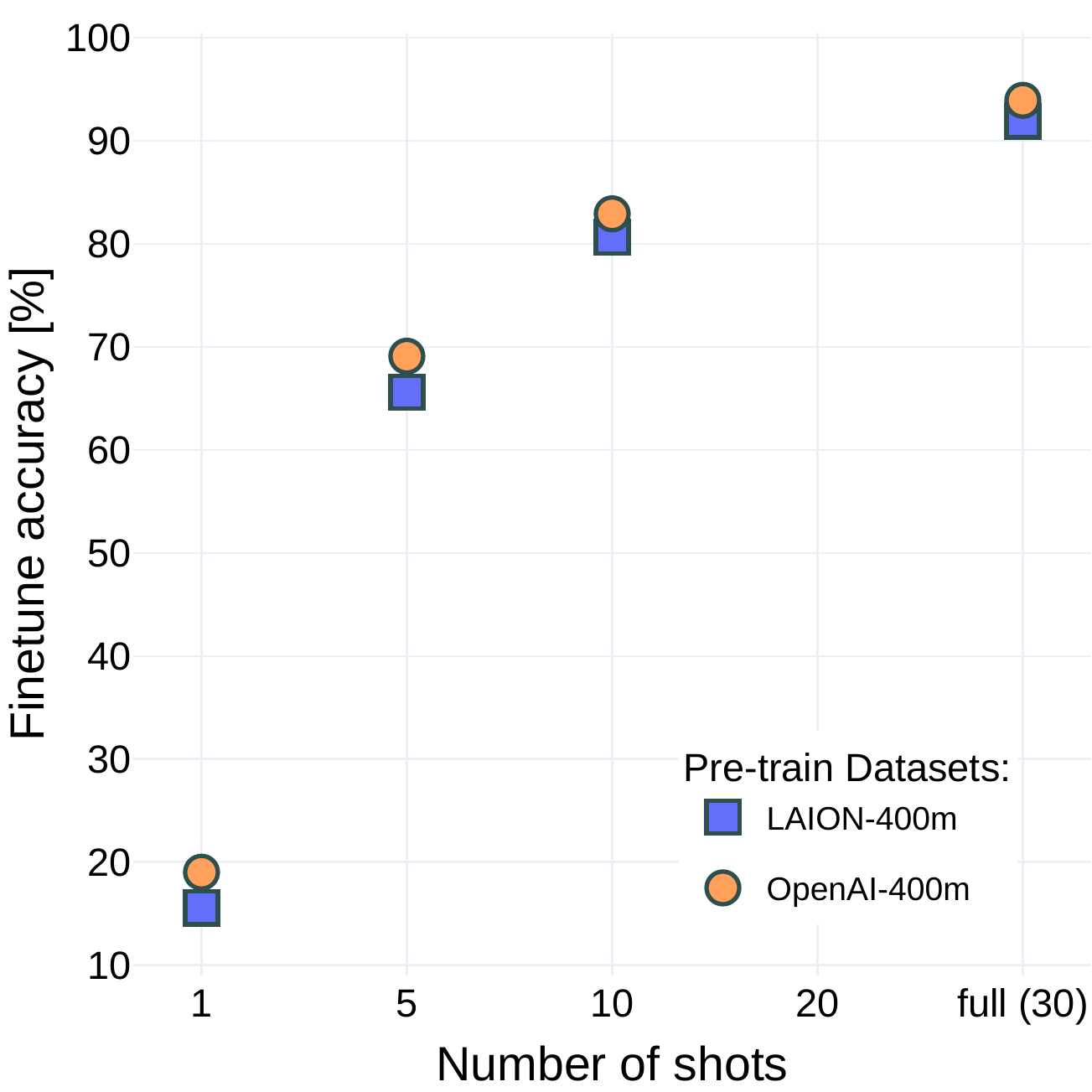}}
    
    \caption{\textbf{Effect of the pre-training data distribution: ViT instead of ResNet-50} While similar to~\Figref{fig:shots_vs_acc_datadist_extend} the difference between the fine-tune performance is minimal, we observe that both models perform also very similarly in the few-shot setting. We hypothesize that this observation could be attributed to the similarity between LAION and OpenAI distributions rather than employing transformer instead of ResNet-50. }
    \label{fig:shots_vs_acc_datadist_vit}
\end{figure}

In order to see the effect of architecture on the observed trends, we extend the results to the effect of pre-training distribution in ~\Figref{fig:shots_vs_acc_datadist_extend} to include Vison Transformers. To do so, we used ViT-B/32 released checkpoints trained on LAION-400m and OpenAI-400m, .~\Figref{fig:shots_vs_acc_datadist_vit} shows the effect of data distribution on finetune transfer to CIFAR100, DTD, and CALTECH101 when using ViT instead of ResNet-50. While similar to~\Figref{fig:shots_vs_acc_datadist_extend} the difference between the fine-tune performance is minimal, we observe that both models perform also very similarly in the few-shot setting. We hypothesize that this observation could be attributed to the similarity between LAION and OpenAI distributions rather than employing a transformer instead of ResNet-50. A controlled study may include to replicate~\Figref{fig:shots_vs_acc_datadist_extend} but with ViT, and we leave that for future work.

\section{Effect of pre-training data distribution: SimCLR instead of CLIP}\label{sec:simclr_vs_clip}
In contrast to previous experiments with CLIP where we fine-tuned end-to-end from the zero-shot pre-trained model, in SimCLR finetuning we fine-tune using LP-FT~\citep{kumar2022fine} because we are no longer able to start with a zero-shot pre-trained model. When we compare to CLIP, we fine-tune both models with LP-FT to facilitate a fair comparison.
LP-FT is the following two-step procedure: for each number of shots $k$ we first freeze the encoder and train a classification head from random initialization using $k$ examples per-class from the downstream task.
In the second step, we initialize the classification head with this linear probe (LP) then unfreeze all weights and finetune (FT) the whole model.



\section{Extended related works}\label{sec:extend_related}
Transfer learning is widely used in deep learning research and practice and has become a cornerstone in both computer vision and natural language processing.
Through the years, there have been many questions on why transfer helps and how to choose a good pre-trained model to transfer from.  
\citet{neyshabur2020being} separated the effect of feature reuse from that of learning low-level pre-training data statistics.
\citet{raghu2019transfusion} investigate the similarity of the pre-training and downstream datasets by looking into medical datasets and found that transfer learning from ImageNet pre-trained models shows little benefit in performance. 
\citet{ericsson2021well} studied the downstream performance of self-supervised models and found that the best self-supervised models of that time could outperform supervised pre-training as an upstream source of knowledge transfer and that the performance of self-supervised models on ImageNet is indicative of downstream performance on natural image classification tasks.
Similarly, \citet{islam2021broad} found that contrastively trained models consistently outperform standard cross-entropy models in transfer learning.
\citet{goyal2021self} showed that self-supervised models outperform supervised models on ImageNet, even when trained on random and uncurated images from the web.
Moreover, they showed that these models are also good at few shot learning by achieving 77.9\,\% top-1 accuracy using only 10\,\% on ImageNet.

Building on contrastive techniques, \citet{radford2021learning} introduced CLIP which learns a joint embedding space for both images and their descriptive captions, making it possible to effectively leverage a large-scale dataset from the Internet. Flamingo~\citep{alayrac2022flamingo}, a visual language model, is another successful example in the line of multimodal models and enables visual question answering and image captioning.
CLIP and similar models like ALIGN~\citep{jia2021scaling}, BASIC~\citep{pham2021combined}, and LiT~\citep{zhai2022lit} demonstrated unprecedented robustness to challenging data distribution shifts.
This accomplishment raised questions on the probable sources of such robustness---whether this robustness is caused by language supervision, the pre-training data distribution, size, or contrastive loss functions.

\citet{fang2022data} investigated this question and found that the diverse training distribution is the main cause of the robustness properties of CLIP. \citet{nguyen2022quality} explored the role of the pre-training dataset for CLIP with a testbed of six pre-training sources, finding that no single pre-training dataset consistently performs best. In recent work, \citet{santurkar2022caption} carefully investigated the effect of language supervision in CLIP-like models, finding it an important factor if the pre-training dataset is large and the captions are descriptive enough.
Unlike their work, we consider end-to-end fine-tuning which result in higher accuracy.
\citet{cherti2021effect} study the effect of scaling the pre-training model and data for both in-domain and out-of-domain transfer. They conduct supervised pre-training while varying pre-training model size and data source(ImageNet-1k/21k or large medical chest XRay datasets), and transfer pre-trained models to different natural or medical targets. They find that, when performing transfer to large X-Ray targets, pre-training on natural ImageNet-21k is as good or better than pre-trained medical X-Ray data.
\citet{djolonga2021robustness} also investigate the impact of the pre-training data size and model scale, finding that increasing both the training set and model sizes significantly improve the distributional shift robustness.

\section{Datasets}\label{sec:datasets}
\subsection{Downstream tasks}\label{sec:downstreamdatasets}
We have used 9 different downstream datasets. ~\tabref{tab:downstreamdatasets} describes the first six datasets in \Figref{fig:shots_vs_acc_datadist_extend}. While these six datasets are internet-crawled datasets and are more common in transfer learning in computer vision benchmarks, we include three new downstream datasets that are domain-specific, \ie the dataset is created after a specific challenge is defined in a specific domain. 
\begin{itemize}
    \item EuroSAT~\citep{helber2019eurosat}: The task is to classify land use and land cover based on Sentinel-2 satellite images. The dataset covers 13 spectral bands and consists of 10 classes within a total of 27,019 labeled and geo-referenced images. we create an $80\%$-$20\%$ random class-balanced split with the provided dataset.
    \item Cassava Leaf Disease Classification~\citep{Cassavaleafdisease2021}: The dataset contains 21,397 images from the Kaggle competition, to give farmers access to methods for diagnosing plant
    diseases. The images are labeled as healthy or as one of four different diseases. we split the dataset with $80\%$-$20\%$ random class-balanced ratio for train and test, respectively.
    \item  Caltech Camera Traps-20~\citep{beery2018recognition}: CCT-20 contains 57,864 images in 15 classes, taken from camera traps deployed to monitor animal populations. Classes are either single species \eg ”Coyote”) or groups
    of species, \eg ”Bird”). CCT-20 is a subset of the iWildCam Challenge 2018, whose yearly editions have been hosted on Kaggle. Here we study the subset of CCT-20 that come from the same locations~\footnote{“Cis” in the main dataset refers to images from locations seen during training, and “trans” refers to new locations not seen during training}, including 14,071 and 16,395 images for train and test respectively.
\end{itemize}

\begin{table}[ht]
  \caption{Details on the downstream datasets used in the experiments.}
  \begin{adjustbox}{width=1\textwidth}
  \begin{tabular}{cp{11cm}}
    \toprule
    Downstream Task                & Description\\
    \midrule
    CIFAR100                             & The task consists in classifying natural images (100 classes, with 500 training images each). Some examples include apples, bottles, dinosaurs, and bicycles. The image size is 32x32. \\
    DTD                               & The task consists in classifying images of textural patterns (47 classes, with 120 training images each). Some of the textures are banded, bubbly, meshed, lined, or porous. The image size ranges between 300x300 and 640x640 pixels.\\
    CALTECH-101                       & The task consists in classifying images of objects (9144  images in 101 classes plus a background clutter class), including animals, airplanes, chairs, or scissors. The image size varies, but it typically ranges from 200-300 pixels per edge.\\
    PETS                                    & The task consists in classifying images of cat and dog breeds (7000 images in 37 classes). Images dimensions are typically 200 pixels or larger\\
    REAL                         &  The task is a subset of larger DomainNet from six distinct domains, including photos (real), painting, clipart, quickdraw, infograph, and sketch. Total size of 172,000              \\
    CLIPART                                   & The task is a subset of larger DomainNet from six distinct domains,  including photos (real), painting, clipart, quickdraw, infograph, and sketch. Total size of 172,000 \\
    \bottomrule
\end{tabular}
\end{adjustbox}
\label{tab:downstreamdatasets}
\end{table}

\subsection{Pre-training datasets}\label{sec:pretraindatasets}
Our study covers 7 pre-training datasets as follow:
\begin{itemize}
    \item YFCC: Our experiments mostly include YFCC-2.7M, a random subset of YFCC-15M. The 15M subset of the YFCC-100M dataset~\citep{thomee2016yfcc100m} was filtered to only include images with English titles or descriptions. The dataset contains 14,829,396 images with natural language captions associated with each image. The images and captions are collected from Flickr. 
    \item LAION~\citep{schuhmann2021laion}: The images and corresponding alt-texts come from web pages collected by Common Crawl~\citep{CommonCrawl} between 2014 and 2021. We randomly select a subset of 2.7M and 15M samples for our experiments.
    \item Redcaps~\citep{desai2021redcaps}: Redcaps contains 11,882,403 examples from 350 manually curated subreddit collected between 2008 and 2020. The subreddits are selected to contain a large number of image posts that are mostly photographs and not images of people.
    \item Shutterstock: 11,800,000 images and captions from the Shutterstock website.
    \item Conceptual Captions-3m~\citep{sharma2018conceptual}: The raw descriptions in Conceptual Captions are harvested from the alt-text HTML attribute associated with web images. This dataset contains 2,799,553 samples, denoted as CC\_2.7m in the plots.
    \item Conceptual Captions-12m~\citep{changpinyo2021conceptual}: A dataset with ~12 million image-text pairs. It is larger than CC\_2.7m and covers a much more diverse set of visual concepts. We randomly select 2.7M samples from this dataset, denoted as CC\_12\_2.7m.
    \item WIT~\citep{srinivasan2021wit}: Image-text pairs come from Wikipedia pages. We use reference description as the source of text data and obtain 5,038,295  examples in total after filtering to include only the English language.
   
\end{itemize}

\begin{figure}[t]
    \centering
    {\includegraphics[width=.99\textwidth]{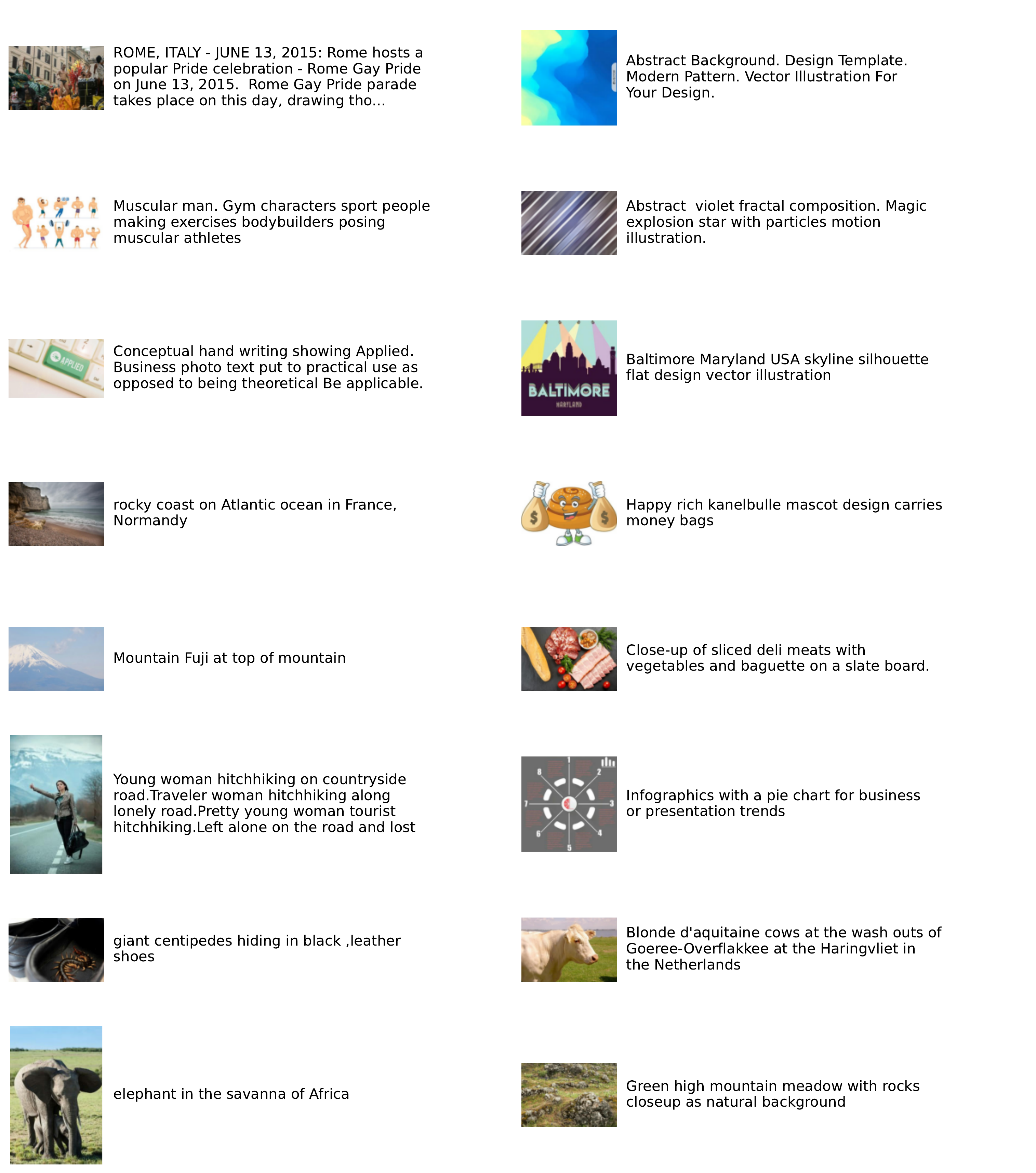}}
    \caption{\textbf{Random training samples from Shutterstock}}
    \label{fig:samples_shutterstock}
\end{figure}

\begin{figure}[t]
    \centering
    {\includegraphics[width=.99\textwidth]{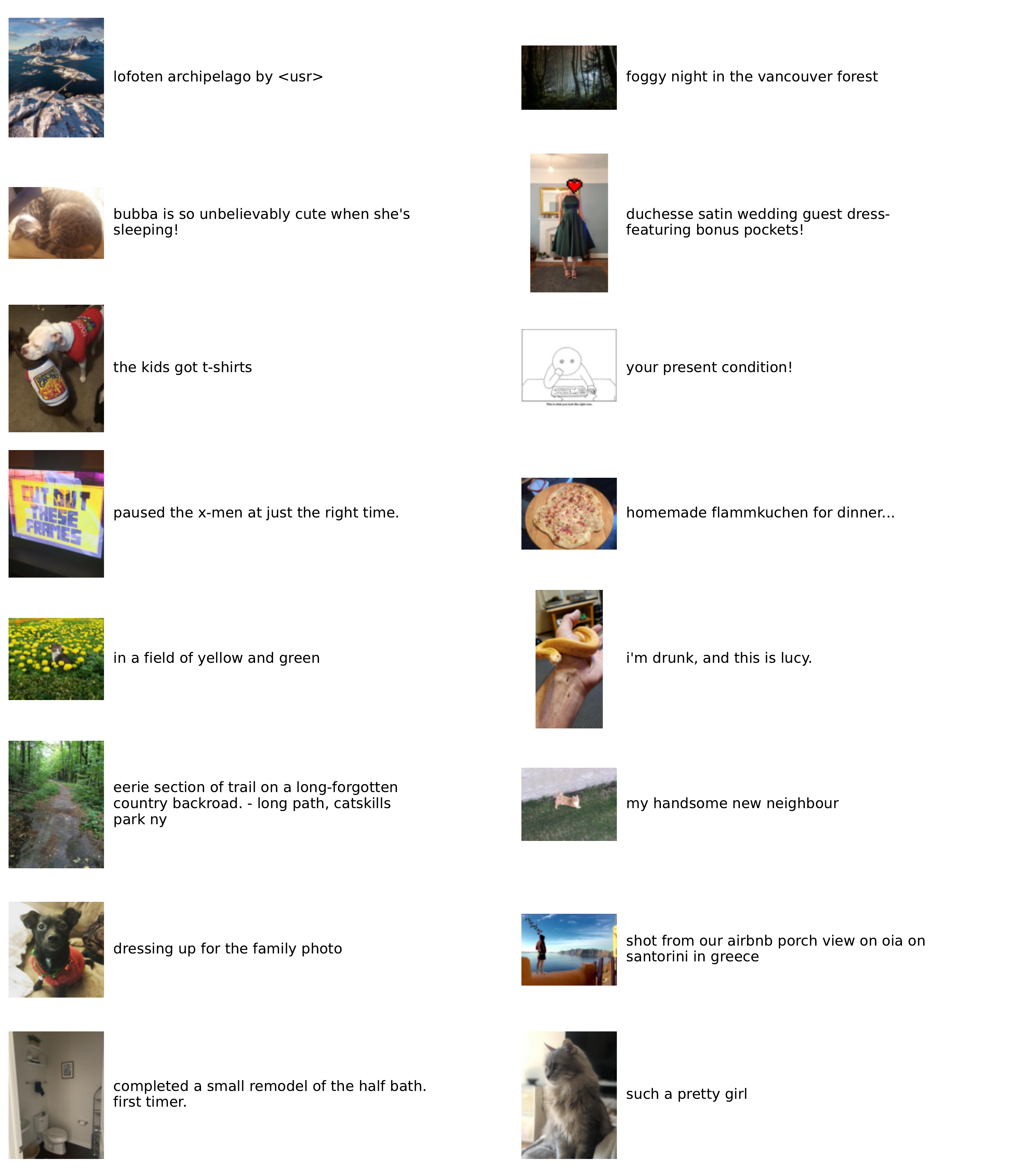}}
    \caption{\textbf{Random training samples from Redcaps }}
    \label{fig:samples_redcaps}
\end{figure}

\begin{figure}[t]
    \centering
    {\includegraphics[width=.99\textwidth]{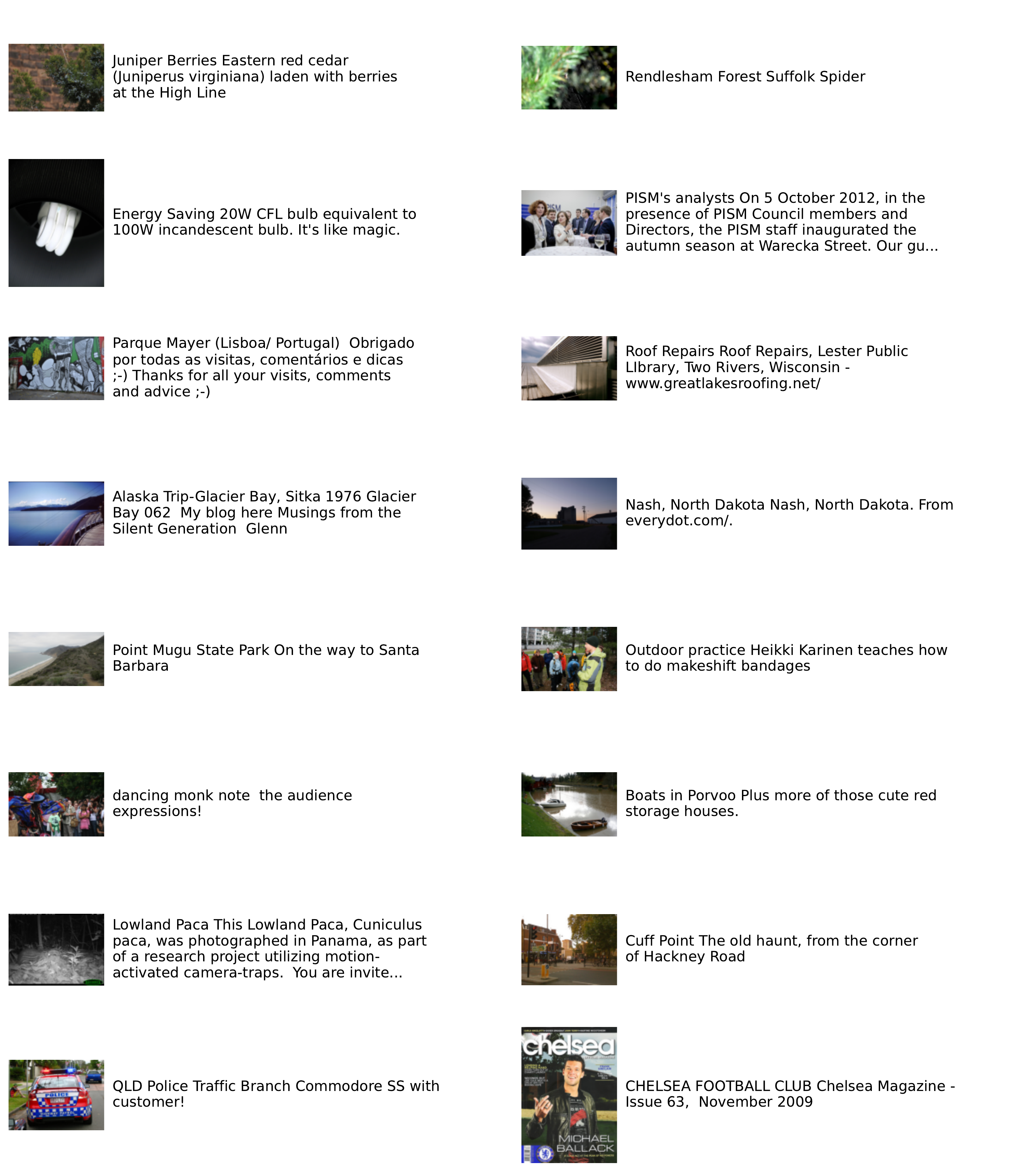}}
    \caption{\textbf{Random training samples from YFCC}}
    \label{fig:samples_yfcc15m}
\end{figure}

\begin{figure}[t]
    \centering
    {\includegraphics[width=.99\textwidth]{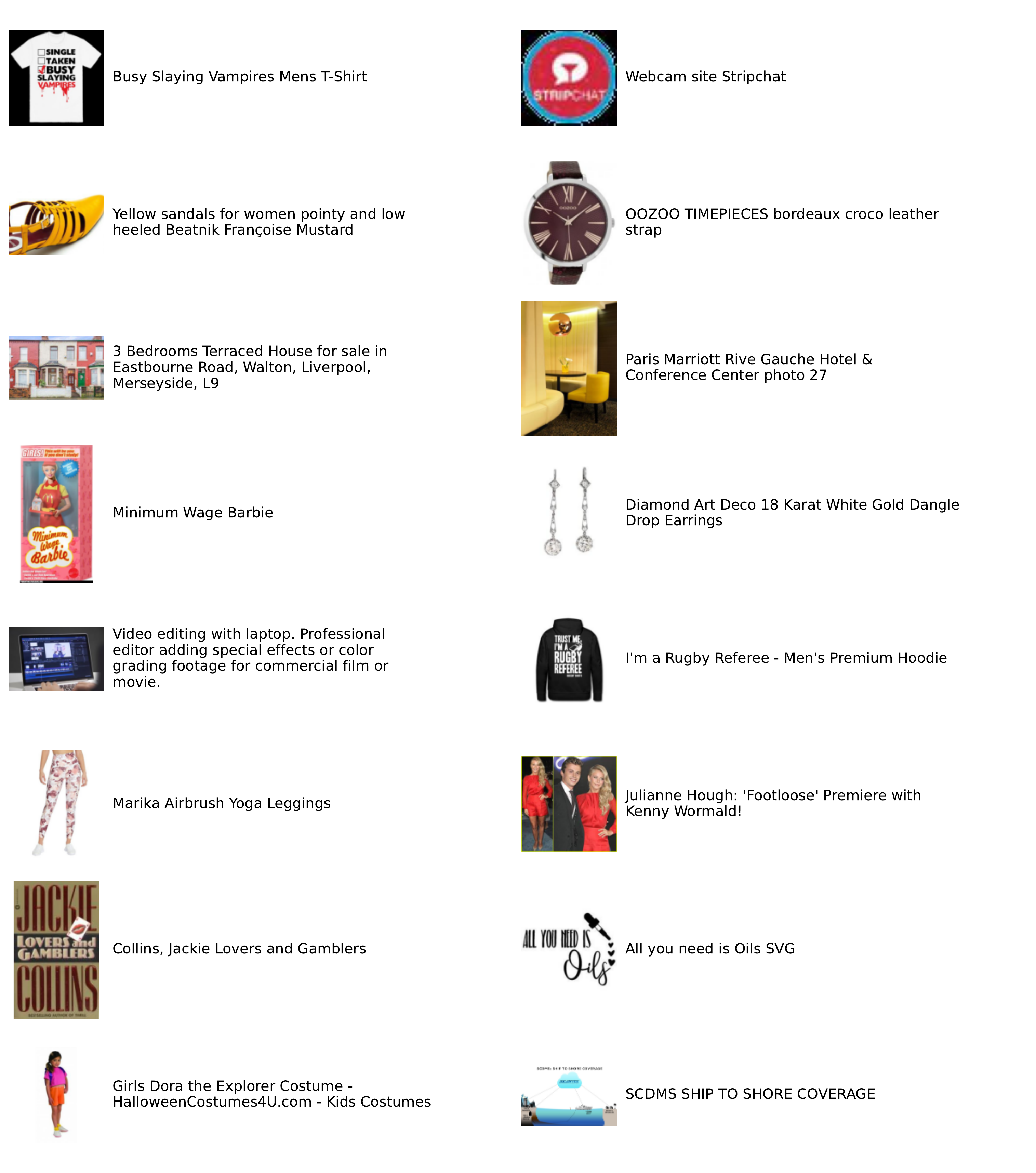}}
    \caption{\textbf{Random training samples from LAION}}
    \label{fig:samples_laion15m}
\end{figure}

\begin{figure}[t]
    \centering
    {\includegraphics[width=.99\textwidth]{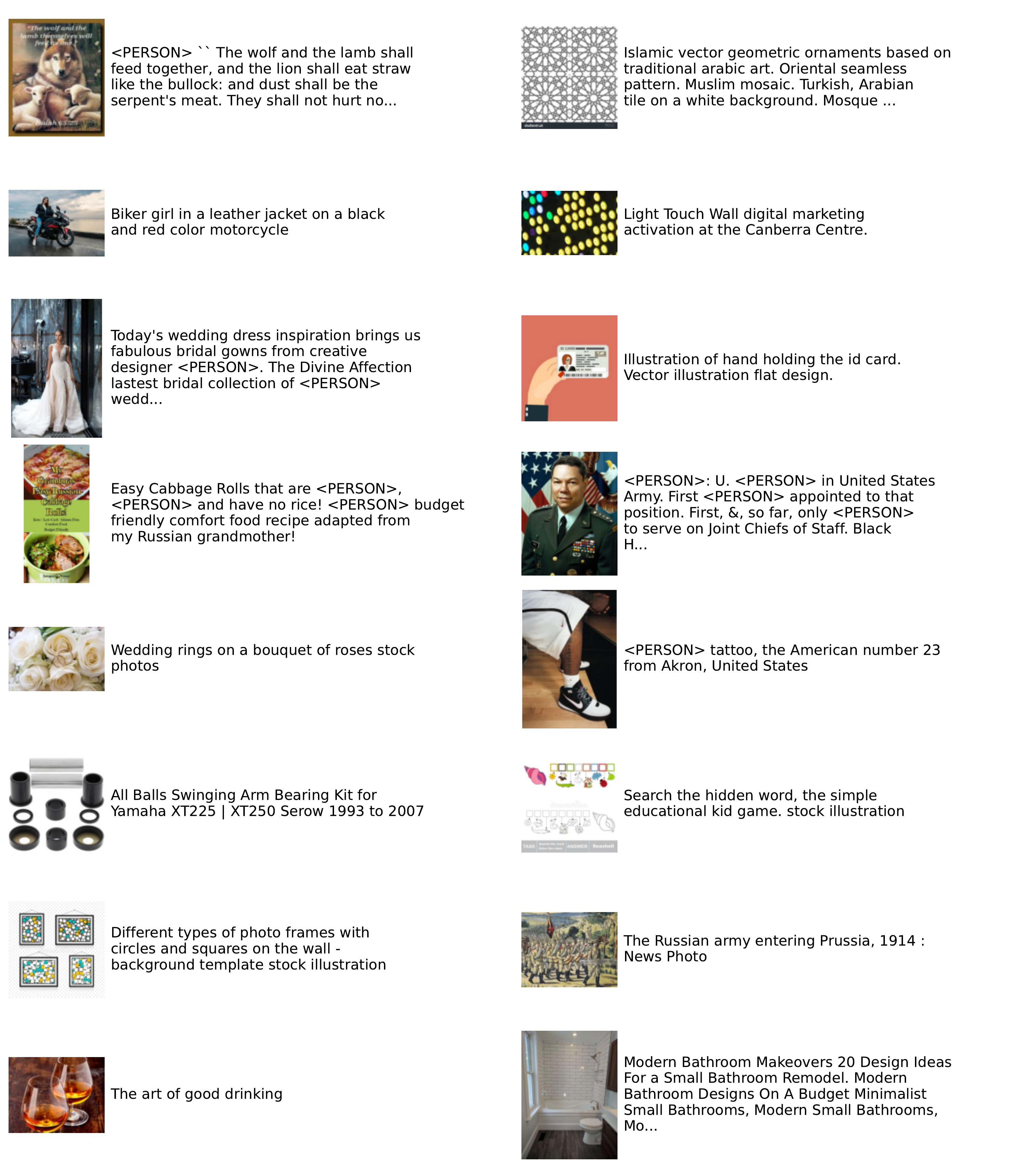}}
    \caption{\textbf{Random training samples from Conceptual Captions }}
    \label{fig:samples_cc12m}
\end{figure}

\begin{figure}[t]
    \centering
    {\includegraphics[width=.99\textwidth]{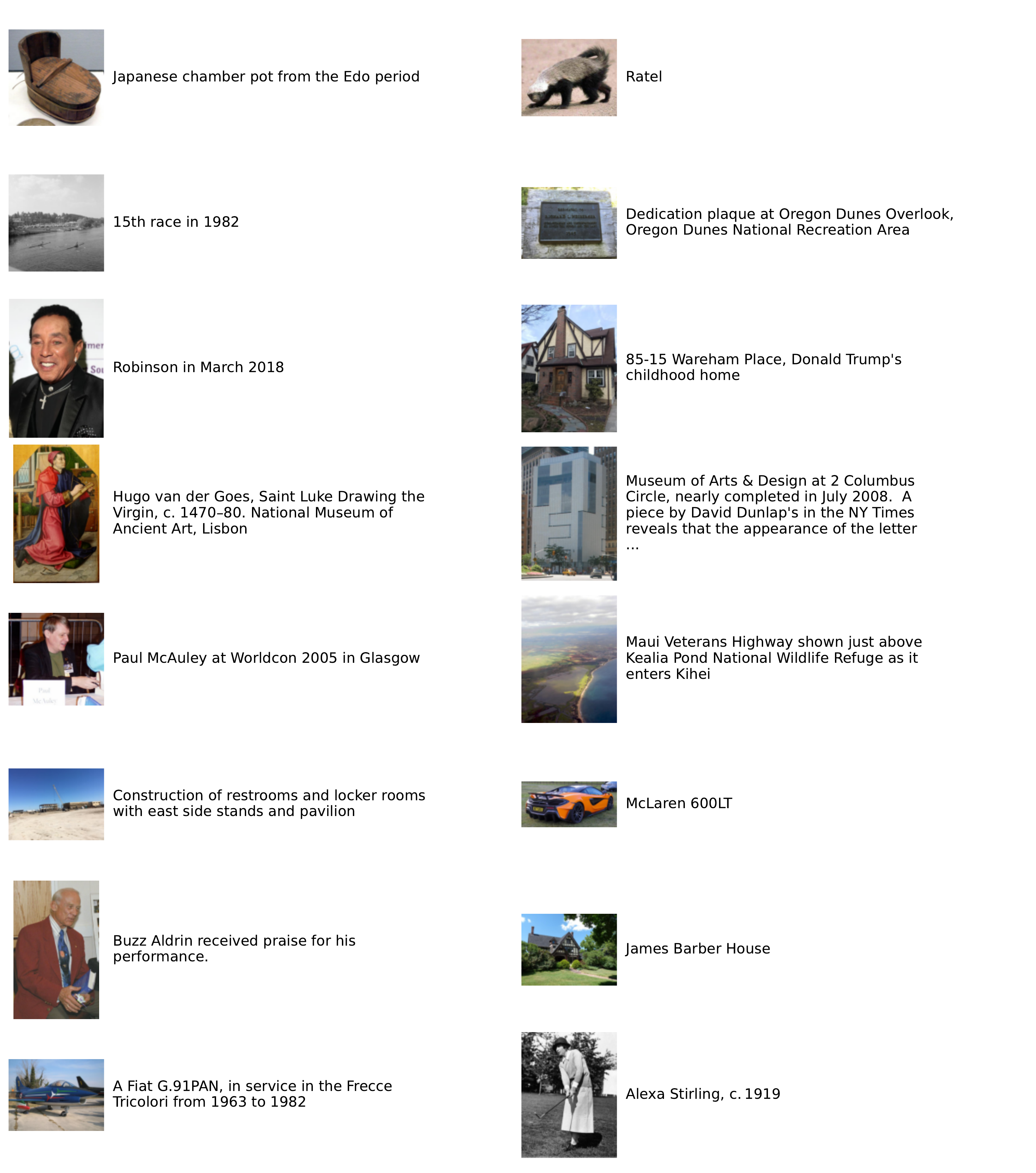}}
    \caption{\textbf{Random training samples from WIT}}
    \label{fig:samples_wit}
\end{figure}

\tabref{tab:pretraindatasets} shows their main source and total size.  We also show some examples of image-caption pairs randomly selected from Shutterstock in \Figref{fig:samples_shutterstock}, Redcaps in~\Figref{fig:samples_redcaps}, YFCC-15m in ~\Figref{fig:samples_yfcc15m}, LAION-15m in \Figref{fig:samples_laion15m}, Conceptual Captions in ~\Figref{fig:samples_cc12m}, and WIT in \Figref{fig:samples_wit}. \tabref{tab:top20_words} also shows the most common words in captions of these pre-training datasets. 

Looking at Redcaps samples in~\Figref{fig:samples_redcaps} and also the top 20 captions shows many samples of animals. This is showing why Redcaps perform better on PETS. Samples from WIT in~\Figref{fig:samples_wit} and also its top 20 words mostly featuring geographical locations, which is rare in our downstream task, hence performing worst compared to other pre-training distributions. Shutterstock top 20 words also include words like "pattern", "texture", "and design" which are close to DTD classes, hence showing superior performance in this downstream task.

\begin{table}[ht]
  \caption{Details on pre-training datasets}
  \centering

  \begin{tabular}{ccccc}
    \toprule
    Dataset             & Source                & Total size\\
    \midrule
    YFCC                & Flickr                & 14,826,000\\
    LAION               & Common Crawl          & 15,504,742 \\
    CC-12M              & Unspecified web pages & 9,594,338  \\
    RedCaps             & Reddit                & 11,882,403 \\
    WIT                 & Wikipedia             & 5,038,295 \\
    ShutterStock        & ShutterStock          & 11,800,000     \\
    IN1K-Captions       & ImageNet              & 463,622     \\
    \bottomrule
\end{tabular}

\label{tab:pretraindatasets}
\end{table}

\begin{table}[ht]
  \caption{Most common words in captions of pre-training distributions}
  \begin{adjustbox}{width=1\textwidth}
  \begin{tabular}{cp{11cm}}
    \toprule
    Pre-training dataset                & Top 20 words in 1M sample of captions\\
    \midrule
    Shutterstock                             & \textbf{background}, vector, illustration, \textbf{design}, icon, \textbf{pattern}, \textbf{texture}, style, woman, concept, hand, color, flower, view, template, line, business, logo, card, symbol\\
    Redcaps                               & day, today, year, time, cat, plant, friend, anyone, picture, baby, guy, week, dog, home, morning, night, month, way, boy, work\\
    YFCC-15m                       & photo, day, park, street, city, picture, view, time, world, year, house, state, center, part, garden, shot, image, building, road, museum\\
    LAION-15m                                    & photo, stock, image, black, woman, design, set, vector, white, print, home, men, blue, dress, art, card, sale, gold, bag, cover\\
    CC-12m                         &  illustration, stock, art, design, photo, image, background, room, vector, house, home, woman, wedding, style, photography, royalty, car, fashion, girl, world\\
    CC-3m                                   & background, actor, artist, player, illustration, view, woman, man, football, team, tree, premiere, city, vector, day, girl, beach, game, hand, people\\
    WIT                                   & view, church, station, map, house, building, hall, museum, city, location, street, park, river, state, john, county, town, center, bridge, world\\
    \bottomrule
\end{tabular}
\end{adjustbox}
\label{tab:top20_words}
\end{table}

\end{document}